%% file: main.tex
\documentclass{article}

\usepackage{PRIMEarxiv}

\usepackage{amsmath,amssymb,amsthm,mathtools,bm}
\usepackage{graphicx}
\usepackage{booktabs}
\usepackage{multirow}
\usepackage{makecell}
\usepackage{xcolor}
\usepackage{hyperref}
\usepackage[authoryear,round]{natbib}
\usepackage{subcaption}
\usepackage{enumitem}
\usepackage{algorithm}
\usepackage{algpseudocode}
\usepackage{wrapfig}
\usepackage{tabularx}
\usepackage[capitalise,noabbrev]{cleveref}   
\crefname{assumption}{Assumption}{Assumptions}
\Crefname{assumption}{Assumption}{Assumptions}
\graphicspath{{figures/}{figures/qualitative/}}

\input{macros}

\input{theory_compat}

\title{A Verifier from the Past: Test-Time Compute for Time Series Foundation Models}

\author{
\begin{tabular}{c}
Arjun Choudhry$^{1,2}$\thanks{Work done during an internship at Synthefy}\quad
Aditya Narayanan$^2$ \quad
Raghav Mallampalli$^2$ \quad
Po-han Li$^2$ \\[4pt]
Sai Shankar Narasimhan$^{2,3*}$ \quad
Raimi Shah$^2$ \quad
B. Aditya Prakash$^1$ \quad
Shubhankar Agarwal$^2$ \\[8pt]
$^1$Georgia Institute of Technology \qquad
$^2$Synthefy \qquad
$^3$The University of Texas at Austin \\[3pt]
\end{tabular}
}

\begin{document}
\maketitle

\input{sections/00_abstract}
\input{sections/01_introduction}
\input{sections/09_related}
\input{sections/03_method}
\input{sections/05_setup}
\input{sections/06_main_results}

\input{sections/07_mechanism}
\input{sections/08_limits}
\input{sections/10_conclusion}

\bibliographystyle{plainnat}
\bibliography{references}

\clearpage
\appendix
\input{appendix/A_proofs}
\newpage
\input{appendix/B_views}
\newpage
\input{appendix/B2_frontier_controls}
\newpage
\input{appendix/C_full_tables}
\newpage
\input{appendix/D_mechanism_details}
\newpage
\newpage

\end{document}

%% file: macros.tex
\newtheorem{theorem}{Theorem}
\newtheorem{proposition}{Proposition}

\makeatletter
\let\c@proposition\c@theorem
\let\c@lemma\c@theorem
\let\c@corollary\c@theorem
\makeatother
\theoremstyle{definition}
\newtheorem{assumption}{Assumption}

\newcommand{\draftmode}{1}
\newcommand{\todo}[1]{\ifnum\draftmode=1\textcolor{red}{\textbf{[TODO: }#1\textbf{]}}\fi}
\newcommand{\needsrun}[1]{\ifnum\draftmode=1\textcolor{orange}{\textbf{[NEEDS RUN: }#1\textbf{]}}\fi}
\newcommand{\note}[1]{\ifnum\draftmode=1\textcolor{blue}{\small[#1]}\fi}

\AtBeginDocument{%
  \crefname{theorem}{Theorem}{Theorems}%
  \Crefname{theorem}{Theorem}{Theorems}%
  \crefname{proposition}{Proposition}{Propositions}%
  \Crefname{proposition}{Proposition}{Propositions}%
  \crefname{lemma}{Lemma}{Lemmas}%
  \Crefname{lemma}{Lemma}{Lemmas}%
  \crefname{corollary}{Corollary}{Corollaries}%
  \Crefname{corollary}{Corollary}{Corollaries}%
  \crefname{assumption}{Assumption}{Assumptions}%
  \Crefname{assumption}{Assumption}{Assumptions}%
  \crefname{remark}{Remark}{Remarks}%
  \Crefname{remark}{Remark}{Remarks}%
  \crefname{definition}{Definition}{Definitions}%
  \Crefname{definition}{Definition}{Definitions}%
}

\newcommand{\bgaa}{\textsc{Bgaa}}

\newcommand{\gift}{GIFT-Eval}
\newcommand{\boom}{BOOM}
\newcommand{\fev}{fev-bench}
\newcommand{\ttc}{test-time compute}

\newcommand{\mtirex}{\textsc{TiRex-2}}
\newcommand{\mchronos}{\textsc{Chronos-2}}
\newcommand{\mrone}{\textsc{PatchTST-FM-R1}}
\newcommand{\mtoto}{\textsc{Toto-2-313M}}

\newcommand{\qfan}{\widehat{Q}}      
\newcommand{\admit}{\widehat{\mathcal{A}}}  
\newcommand{\bt}{\widehat{L}}        
\newcommand{\crps}{\mathrm{CRPS}}

%% file: theory_compat.tex
\DeclareMathOperator{\CRPS}{CRPS}

\newcommand{\dg}{\,\mathrm{d}}

%% file: sections/00_abstract.tex
\begin{abstract}
Test-time compute (TTC) improves language models by spending additional inference compute to generate, verify, and select among candidate outputs, but this recipe has not transferred directly to time series forecasting.
A language model can verify a proof or vote across
samples, but a forecaster must be causal; it cannot use future information to score a candidate. We propose \emph{Backtest-Gated Augmentation Averaging} (\bgaa), a TTC algorithm for time series forecasting that builds a causal verifier from the series's own history, using rolling-origin backtests to score structured views of the context, discard those that underperform the untransformed context, and average the survivors' predictive distributions. 
\bgaa\ improves every tested TSFM–benchmark combination, reducing MASE by 0.5–1.0\% on GIFT-Eval, 1.3–5.0\% on fev-bench, and 1.7–5.9\% on BOOM, where a 313M model with TTC outperforms the 1B and 2.5B models in its family.
We further show that useful views have structured, model-dependent effects consistent with a prior-transport interpretation.
On the 53 cross-fittable \gift\ configs, the deployed policy gains only $0.4$--$1.3\%$ CRPS, versus $1.4$--$4.1\%$ when choosing the best view for each dataset configuration and $17$--$19\%$ when choosing the best view separately for each test window. This gap suggests that a major remaining challenge is window-level view selection.

\end{abstract}

%% file: sections/01_introduction.tex
\section{Introduction}
\label{sec:intro}

A time series foundation model (TSFM) is pretrained once but deployed across many diverse series. Improving performance usually requires larger models or more training. \emph{Test-time compute} offers another route: keep the model frozen and spend additional inference-time compute to improve forecast accuracy. Language models do this by thinking longer \citep{nye2021scratchpad,wei2022cot} or wider, sampling and selecting among multiple candidates \citep{snell2025scaling}. Thinking wider requires a \emph{verifier}, a signal for which candidate is better. Language modeling has several: program synthesizers run tests \citep{li2022alphacode}, provers check steps, reward models score answers \citep{cobbe2021verifiers}, and majority vote exploits agreement across correct chains \citep{wang2023selfconsistency}.

These verifiers don't transfer to time series. A numeric forecast has no
syntax to violate, nor tests to fail. A reward model would need labeled
forecasts for the deployment series, which are unavailable at inference time. Without a causal verifier, TSFMs are deployed with a single forward pass.

A forecaster has something a language model does not: the series' \emph{own past}. A language prompt provides only a single instance of the task, whereas a forecasting context contains earlier prediction problems with observed outcomes. We rebuild the deployment task from the context alone and score candidates by rolling-origin backtest, reading
nothing past the last observed point.

We introduce \emph{Backtest-Gated Augmentation Averaging} (\bgaa), a test-time compute algorithm for time series forecasting. \bgaa\ generates several structured views of the same context, obtains a candidate forecast from each using the same frozen model, and scores them using a rolling-origin backtest on the observed history. Candidates that outperform the untransformed input on the historical backtest are retained
and aggregated. Because past performance is only a proxy for future accuracy, the verifier is a \textit{surrogate}, but it is causal and available for every series at test time.

Why should structured views give different, \emph{sometimes better}, forecasts from the same observations? Under a Prior Fitted Network idealization, a frozen TSFM predicts under a distribution induced by pretraining \citep{muller2022}. For fixed pointwise, monotone transforms, forecasting a transformed context and mapping quantiles back is equivalent to prediction under a transported trajectory distribution (\cref{prop:transport-main}). Our deployed views are more general, so we treat this as an illustrative interpretation and test for systematic, model-dependent effects in \cref{sec:mechanism}.

Test-time compute on TSFMs needs safeguards against degradation.
The base forecast remains a candidate, and the backtest acts as a
\emph{referee}: a view is admitted only if it beats the base on held-out windows (\cref{sec:gate-main}).
This makes additional inference compute selective, since views are used only when the historical backtest indicates improvement over the base. 

\begin{figure}[t]
  \centering
  \includegraphics[width=\textwidth]{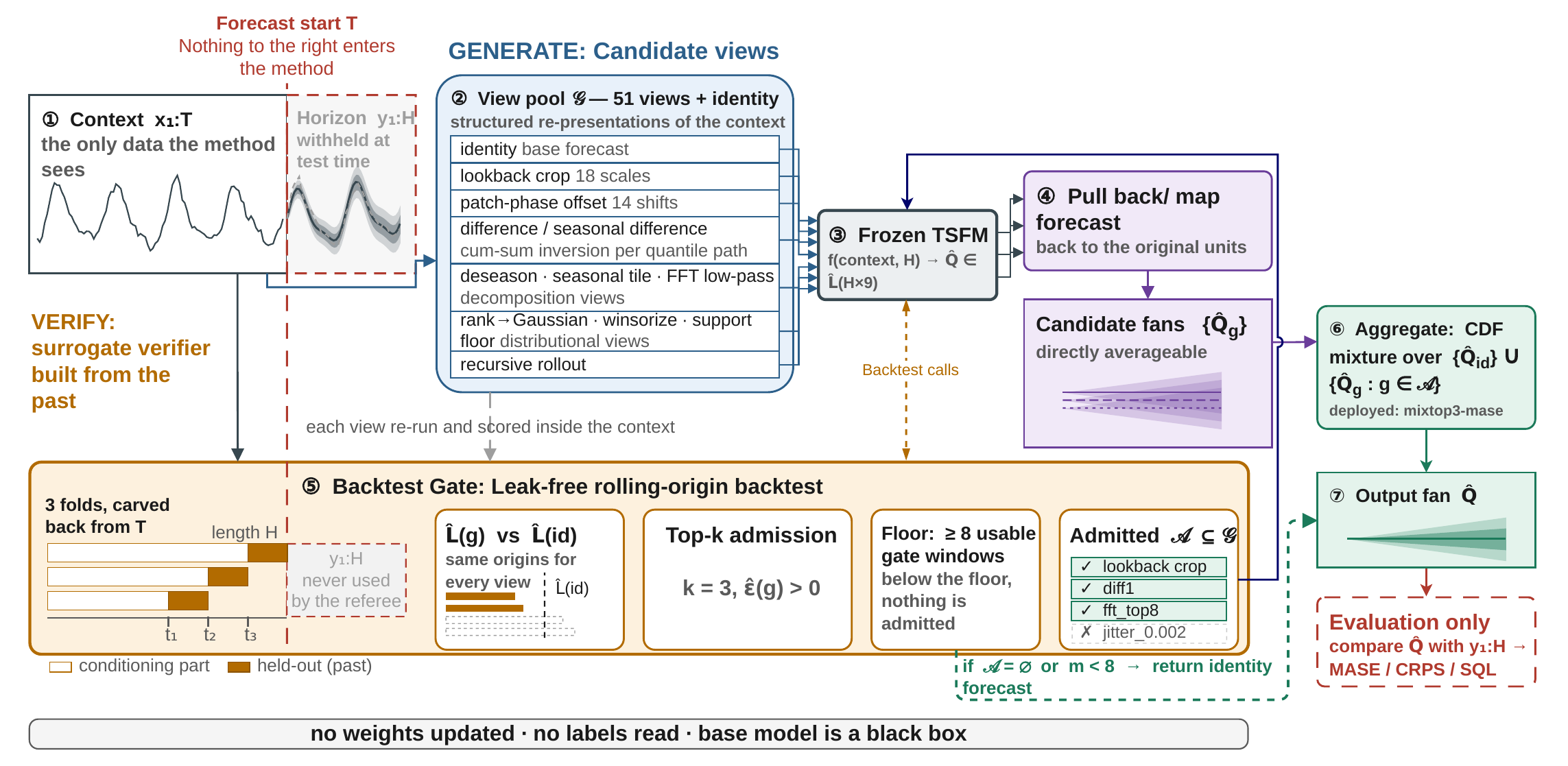}
  \caption{\textbf{\bgaa\ system overview.}
\textbf{(1)} The observed context is the only input available at test time.
\textbf{(2)} A fixed pool of transformed views is generated, and
\textbf{(3)} each is forecast with the same frozen TSFM.
\textbf{(4)} Forecasts are mapped back to the original observation space.
\textbf{(5)} A leak-free rolling-origin backtest scores the views against the
base and admits the best candidates.
\textbf{(6)} The admitted forecasts are aggregated with the base forecast to
produce \textbf{(7)} the final predictive fan.} 
  \vspace{-13pt}
  \label{fig:method}
\end{figure}

We test \mchronos, \mtirex, \mrone, and \mtoto\ on \gift, \boom, and \fev.
All $12$ model--benchmark combinations improve with no per-model re-tuning:
$0.5$ to $1.0\%$ MASE on \gift, $1.7$ to $5.9\%$ on \boom, $1.3$ to
$5.0\%$ on \fev. The gain often exceeds the spacing between model sizes: on
\boom, \ttc\ on \mtoto\ beats the $1$B and $2.5$B models of its own family. On \gift, the \texttt{high} \mtirex\ rung achieves lower MASE at lower inference FLOPs than \textsc{Toto-2-2.5B} (\cref{sec:results}). Simple view ensembling does not explain this gain: at matched compute on \mchronos, random selection from the same pool worsens MASE by 4.8\%, while BGAA improves it by 1.0\%.

We also measure the remaining headroom. On the $53$ \gift\ dataset configurations that support cross-fitting, \bgaa\ improves CRPS by $0.4$--$1.3\%$. Cross-fitted selection of the best single view per configuration raises this to 1.4–4.1\%, while choosing the best view separately for each forecast window using its realized outcome gives $17$--$19\%$. This gap motivates better window-level verification, while leaving open gains from better candidate views.

Our contributions are four-fold:
\begin{enumerate}[leftmargin=1.3em,itemsep=1pt,topsep=2pt]
  \item \textbf{Test-time compute for TSFMs} (\cref{sec:method}).
    \bgaa\ needs only a quantile-forecast API, updates no weights, reads no
    future labels, and improves every model we test on every benchmark.
  \item \textbf{Why augmented views help} (\cref{sec:mechanism}).
    Under the Prior Fitted Network idealization, fixed pointwise monotone views correspond to transported trajectory distributions. This motivates three predictions about structured, model-specific view effects, which \cref{sec:mechanism} shows them for all four models.
  \item \textbf{Evidence at scale} (\cref{sec:results,sec:mechanism}).
    Across $4$ models and $3$ benchmarks, we measure the accuracy--FLOPs frontier and show that the gains arise from structured, model-dependent view effects and effective backtest-based selection, rather than simply ensembling more forecasts.
  \item \textbf{Verification headroom} (\cref{sec:limits}).
    Using cross-fitted and label-based diagnostics, we quantify the remaining gap at the window-level, showing substantial window-specific headroom.
\end{enumerate}

%% file: sections/09_related.tex
\section{Related work}
\label{sec:related}

\textbf{Time series foundation models (TSFMs).}
TSFMs are models pretrained on large, heterogeneous corpora and deployed zero-shot. Given a context, they forecast an unseen series with no task-specific training. They differ in how series are tokenized and pretrained \citep{ansari2024,chronos2,wen2026patchtstfm,toto2024,woo2024moirai,das2024timesfm,kamarthi2023large, goswami2024moment}. We study a complementary question: how to improve forecasts from an already-trained TSFM by spending additional inference compute while keeping its weights fixed.

\textbf{Test-time compute in language modeling.}
Self-consistency, best-of-n with a reward model, verifier-guided search and
compute-optimal inference
\citep{wang2023selfconsistency,cobbe2021verifiers,snell2025scaling,lightman2024letsverify}
generate candidates and use different signals to select among them.
Forecasting complicates candidate selection:
the future observations needed to evaluate a forecast are unavailable at prediction
time, and agreement among sampled forecasts does not by itself establish accuracy.

\textbf{Test-time augmentation (TTA).}
TTA is standard in vision, where transformed inputs are evaluated and their
predictions aggregated, with the aggregation rule as important as the
augmentations \citep{chen2020,shanmugam2021,lyle2020}.
For forecasting, transformed contexts enable producing multiple candidate
forecasts; the question remains which views help for a time series.

\textbf{Forecast combination and label-free adaptation.}
Combining forecasts is a longstanding approach \citep{bates1969,clemen1989,timmermann2006}, including the choice between averaging quantiles and probabilities \citep{lichtendahl2013,gneiting2007}. Our use of forecast combination builds on this literature; the question here is which forecasts from transformed inputs to include in the combination. Conformal and recalibration methods use observed data to adapt predictive distributions or restore calibration \citep{gibbs2021acp,stankeviciute2021conformal}, whereas our problem is candidate selection before aggregation.

\textbf{Inference-time compute for frozen TSFMs.}
Two concurrent methods spend inference compute on a frozen forecaster.
\cite{hua2026diversified} widen a saturating sample distribution by perturbing
the input, but their main aggregator takes the median over all views, so harmful
views are not filtered. Their selective variant ranks views against
ground truth, and thus is limited to settings with labels.
\cite{pandey2026gitco} instead edit a single context, suppressing the most
disruptive patch when a meta-feature gate fires, with the gate fit by
cross-validation over a \emph{subset} of \gift.
\bgaa\ instead selects among transformed-input forecasts using rolling-origin
backtests on observed history, requiring neither future labels nor a separately
trained cross-dataset gate (\cref{sec:method}).

%% file: sections/03_method.tex
\section{Method: backtest-gated augmentation averaging}
\label{sec:method}

\textbf{Setup and notation.}
We consider univariate forecasting for this work. Multivariate and covariate-aware forecasting are out of scope.
Let $x_{1:T}=(x_1,\ldots,x_T)\in\mathbb{R}^{T}$ be the history observed up to
the \emph{forecast origin} $T$, the last observed time step (so $T$ is also the
history length); let $H$ be the forecast horizon and
$y_{1:H}=x_{T+1:T+H}\in\mathbb{R}^{H}$ the unobserved target. Such a
history--target pair is a \emph{forecast window}. Since time steps are counted
within each series, history lengths vary across series and windows; we index
windows by $w$, writing $x^{(w)}_{1:T_w}$, $T_w$, and $y^{(w)}_{1:H}$, and drop
$w$ when discussing a single window. A frozen forecaster $f$ returns
\begin{equation}
f(x_{1:T},H)=\widehat Q\in\mathbb{R}^{H\times|\mathcal L|},
\qquad
\mathcal L=\{0.1,\ldots,0.9\},
\end{equation}
a matrix with rows indexed by horizon steps $h\in\{1,\ldots,H\}$ and columns by
quantile levels $\tau\in\mathcal L$; its entry $\widehat Q_{h,\tau}$ is the
predicted $\tau$-quantile at step $h$. We call $f(x_{1:T},H)$ the \emph{base
forecast}.

A \emph{view} $g$ modifies how the history is presented to the
forecaster or how its forecast is returned to the original forecast space.
Formally, $g\colon f\mapsto f_g$ maps the frozen forecaster to a forecaster
$f_g$ with the same interface, $f_g(x_{1:T},H)\in\mathbb{R}^{H\times|\mathcal L|}$.
Let $\mathcal G$ be the fixed pool of candidate views, and
$\widehat Q^{g}=f_g(x_{1:T},H)$ the \emph{candidate forecast} of view $g$, with
entries $\widehat Q^{g}_{h,\tau}$ indexed as for $\widehat Q$. The
\emph{identity view} $\mathrm{id}\notin\mathcal G$ leaves the history and
forecast unchanged, $f_{\mathrm{id}}=f$, so $\widehat Q^{\mathrm{id}}=\widehat Q$
is the base forecast.

A configuration, or \emph{config}, $c$ defines a finite set $\mathcal W_c$ of
evaluation forecast windows $w$, all with the same horizon $H$. \bgaa\ selects a
single set of views $\widehat{\mathcal A}_c\subseteq\mathcal G$ per config and
holds it fixed across $\mathcal W_c$. Selection uses only the observations $\mathcal O_c$ available before the first
forecast origin in $\mathcal W_c$, from which we sample a finite set
$\mathcal W_c^{\mathrm{gate}}$ of historical \emph{gate windows}. This pools historical evidence across gate windows and amortizes
selection cost across evaluation forecasts.

\textbf{Problem definition.} For a loss $\ell$, \bgaa\ seeks a forecaster
$\tilde f$, chosen using only the frozen black-box $f$ and $\mathcal O_c$, whose
\emph{relative risk} $\mathcal R_c$ (its mean per-window loss divided by the
base's, so each window counts equally regardless of scale) satisfies
\begin{equation}
\mathcal R_c(\tilde f)
=
\frac{1}{|\mathcal W_c^{+}|}
\sum_{w\in\mathcal W_c^{+}}
\frac{
\ell\!\left(\tilde f(x^{(w)}_{1:T_w},H),y^{(w)}_{1:H}\right)}
{
\ell\!\left(f(x^{(w)}_{1:T_w},H),y^{(w)}_{1:H}\right)}
<
\mathcal R_c(f)=1,
\label{eq:problem}
\end{equation}
where $\mathcal W_c^{+}\subseteq\mathcal W_c$ contains the windows with
positive base loss. Since their targets are unobserved at selection time,
\bgaa\ scores each view by the gate-window surrogate $1-\widehat\epsilon(g)$
(\cref{eq:gate-score}).

\textbf{Method overview.}
\bgaa\ generates alternative forecasts from a frozen model, evaluates them on
historical forecasting tasks, and combines selected alternatives with the base
forecast. Each historical task holds out an already-observed segment and
forecasts it using only preceding observations. These backtests select views
once per config, before any evaluation window is observed. If historical
evidence is insufficient or no alternative improves over the base, \bgaa\
returns the base forecast. \Cref{fig:method} illustrates the pipeline, and
\cref{alg:bgaa-main} summarizes the procedure. Note: \(m_{\min}=8\) is the standard gate, with a short-history exception, described in \cref{sec:setup}.

\textbf{Candidate generation.}
The candidate generator produces alternative forecasts for the same target
horizon using the frozen forecaster. A \emph{view family} specifies a type of
modification, with different parameter settings defining individual views.
For example, cropping is a family; retaining the latest $50\%$ or $75\%$ of
the history defines two views. A view need not be an invertible input
transformation: it may modify the context, post-process a forecast, or make
multiple model calls, provided its output is returned to the same timestamps
and units as the target. The fixed pool $\mathcal G$ contains lookback crops,
patch-phase offsets, differencing and seasonal differencing, decomposition and
frequency-based views, distributional transforms, recursive rollout, and small
perturbations. The complete catalog and pull-back rules are in
\cref{sec:app:views}.

\textbf{Selection using historical backtests.}
To evaluate a view without using any evaluation-window target, we recreate the
forecasting problem inside an observed history. We index gate windows as
$\mathcal W^{\mathrm{gate}}_c=\{w_1,\ldots,w_M\}$ and write $x^{(j)}=x^{(w_j)}$
for the observed history of $w_j$. Each gate window provides $n_f=3$
backtests, one per rolling origin $t_{jr}$, $r\in\{1,\ldots,n_f\}$; at each
origin, we forecast the next $H$ observations using only $x^{(j)}_{1:t_{jr}}$.
For the
loss $\ell$, instantiated below, the empirical loss of view $g$ is
\begin{equation}
\widehat L_j(g)
=
\frac{1}{n_f}
\sum_{r=1}^{n_f}
\ell\!\left(
f_g(x^{(j)}_{1:t_{jr}},H),
x^{(j)}_{t_{jr}+1:t_{jr}+H}
\right).
\label{eq:backtest-loss}
\end{equation}
For point selection, $\ell$ is the mean absolute error of the median forecast.
Because the MASE scale factor is common to a view and the base within a
window, their relative comparison is unchanged, so we call this the MASE gate.
The distributional alternative uses mean pinball loss over $\mathcal L$, a
finite-quantile surrogate for CRPS, and is called the CRPS gate.

Windows with $\widehat L_j(\mathrm{id})=0$ are excluded because relative
improvement would be undefined. With
$\mathcal J=\{j\in\{1,\ldots,M\}:\widehat L_j(\mathrm{id})>0\}$ and
$m=|\mathcal J|$, each candidate view receives the score
\begin{equation}
\widehat{\epsilon}(g)
=
\frac{1}{m}
\sum_{j\in\mathcal J}
\frac{\widehat L_j(\mathrm{id})-\widehat L_j(g)}
     {\widehat L_j(\mathrm{id})}.
\label{eq:gate-score}
\end{equation}
Thus, $\widehat{\epsilon}(g)>0$ means $g$ improves over the base on average
across usable historical windows.

The deployed rule selects up to the $k=3$ views with the largest positive
scores,
\begin{equation}
\widehat{\mathcal A}_c
=
\operatorname{Top}_k
\left\{
g\in\mathcal G:
\widehat{\epsilon}(g)>0
\right\}
\quad\text{ranked by }\widehat{\epsilon}(g).
\label{eq:topk-gate}
\end{equation}
If $m<m_{\min}=8$, or no view has positive improvement,
$\widehat{\mathcal A}_c=\varnothing$ and the base forecast is returned. We
also evaluate a margin rule that requires a minimum mean improvement and a
minimum fraction of wins over $\mathcal J$; its full definition is in
\cref{sec:safety}.

\begin{algorithm}[t!]
\caption{\bgaa\ with deployed top-$k$ selection}
\label{alg:bgaa-main}

\centering
\resizebox{0.9\linewidth}{!}{
\begin{minipage}{\linewidth}
\begin{algorithmic}[1]
\Require frozen forecaster $f$, views $\mathcal G$, horizon $H$
\Statex \textbf{Selection, once per config}
\For{sampled historical gate windows $w_j\in\mathcal W^{\mathrm{gate}}_c$}
    \State evaluate $\widehat L_j(g)$ for
    $g\in\mathcal G\cup\{\mathrm{id}\}$ on $n_f$ rolling origins
\EndFor
\State discard windows with $\widehat L_j(\mathrm{id})=0$
\If{$m<m_{\min}$}
    \State $\widehat{\mathcal A}_c\gets\varnothing$
\Else
    \State compute $\widehat{\epsilon}(g)$ and retain up to $k$
    highest-scoring views with $\widehat{\epsilon}(g)>0$
\EndIf
\Statex \textbf{Application, per forecast window}
\State $\widehat Q^{\mathrm{id}}\gets f(x_{1:T},H)$
\If{$\widehat{\mathcal A}_c=\varnothing$}
    \State \Return $\widehat Q^{\mathrm{id}}$
\EndIf
\State \Return $\operatorname{Aggregate}
\bigl(\widehat Q^{\mathrm{id}},
\{\widehat Q^{g}:g\in\widehat{\mathcal A}_c\}\bigr)$
\end{algorithmic}
\end{minipage}
}

\end{algorithm}

\textbf{Forecast aggregation.}
Aggregation combines the base forecast and the candidate forecasts of
$\widehat{\mathcal A}_c$, each with equal weight
$\lambda_0=1/(|\widehat{\mathcal A}_c|+1)$, into the \emph{aggregate forecast}
\[
\widehat Q^{\mathrm{agg}}
=
\operatorname{Aggregate}\bigl(\widehat Q^{\mathrm{id}},
\{\widehat Q^{g}:g\in\widehat{\mathcal A}_c\}\bigr)
\in\mathbb{R}^{H\times|\mathcal L|}.
\]
\textbf{Vincent averaging} averages corresponding quantiles. \textbf{CDF
mixture} converts each finite quantile fan to an interpolated CDF, averages
the resulting distributions, and inverts the mixture at the nine levels in
$\mathcal L$. We interpolate linearly between reported quantiles and
extrapolate linearly in the tails using the slope of the nearest end segment,
truncating the CDF at 0 and 1. Our deployed policy, \texttt{mixtop3-mase},
combines the MASE gate, top-3 admission, and CDF mixing; \texttt{top3-mase}
uses the same selected set with Vincent averaging.

\textbf{Properties of the aggregation rules.}
\phantomsection\label{sec:gate-main}
For an aggregation rule convex in the gate's loss, \cref{thm:backtest} shows
that admitted historical improvements carry over to the aggregate on the same
backtest windows. Let
$f_{\mathrm{agg}}\colon(x_{1:T},H)\mapsto\widehat Q^{\mathrm{agg}}$ denote the
aggregate forecaster (\bgaa's $\tilde f$), $\widehat L_j(\mathrm{agg})$ its
backtest loss, given by
\cref{eq:backtest-loss} with $f_{\mathrm{agg}}$ in place of $f_g$, and
\[
\widehat L_{\mathrm{rel}}(\mathrm{agg})
=
\frac{1}{m}\sum_{j\in\mathcal J}
\frac{\widehat L_j(\mathrm{agg})}{\widehat L_j(\mathrm{id})}
\]
its mean backtest loss relative to the base. Under uniform weights, if
$\epsilon'=\min_{g\in\widehat{\mathcal A}_c}\widehat{\epsilon}(g)>0$ is the
smallest admitted relative improvement, then
$
\widehat L_{\mathrm{rel}}(\mathrm{agg})
\le
1-(1-\lambda_0)\epsilon'
\le 1.
$
This covers Vincent averaging under the MASE or CRPS gate,
but not the deployed \texttt{mixtop3-mase} policy, because CDF-mixture
quantiles are not convex combinations of member quantiles. The result bounds
$\widehat L_{\mathrm{rel}}(\mathrm{agg})$ only; it does not bound
$\mathcal R_c(f_{\mathrm{agg}})$ or a benchmark's aggregation rule.

CDF mixing has a separate exact-CRPS property: for member CDFs $F_i$ with
weights $\lambda_i\ge0$, $\sum_i\lambda_i=1$, and an outcome $y\in\mathbb{R}$,
$\operatorname{CRPS}(\sum_i\lambda_iF_i,y)$ is at most the weighted average of
the members' CRPS values (\cref{thm:ambiguity}). This does not imply the
mixture beats the base or the best member, and the identity applies to the
full CDF before conversion back to nine quantiles. Full assumptions and proofs
are in \cref{sec:app:proofs}.

\textbf{Computational cost.}
\bgaa\ separates selection from application. Selection evaluates the view
pool on historical backtests once per config; the resulting
$\widehat{\mathcal A}_c$ is reused across its forecast windows. Application
must still recompute the base and admitted candidate forecasts for each new
context--horizon pair. Individual views can require zero additional calls
when they post-process a cached base forecast, one call for a transformed
context, or multiple sequential or seeded calls for recursive or stochastic
views. A cached base forecast is reusable only for the same input, not across
forecast windows. The four \emph{thinking modes} vary the number of candidate
views, gate windows and rolling origins, and admitted views. Across the four
models they span roughly $2.3$--$9.5\times$ the cost of a single pass on
\gift\ (\cref{tab:ladder}); exact forward-pass accounting is in
\cref{sec:app:ladder}.

%% file: sections/05_setup.tex
\section{Experimental setup}
\label{sec:setup}

\textbf{Models.}
We evaluate four public frozen TSFMs spanning different architectures and
pretraining corpora: \mchronos\ \citep{chronos2}, \mtirex\
\citep{tirex2}, \mrone\ \citep{wen2026patchtstfm}, and \mtoto\
\citep{toto2}, all  evaluated in their univariate
settings. \mtoto\ is the 313M family member used in our model-scaling comparisons.

\textbf{Benchmarks.}
\gift\ \citep{aksu2024gifteval} contains $97$ configs.
\boom\ \citep{toto2024} contains $7413$ dataset$\times$term observability
tasks, of which the official leaderboard evaluates a $6750$-task scaled
subset. \fev\ \citep{fevbench2025} contains $100$ forecasting tasks.

\textbf{Metrics and statistics.}
We report MASE for point accuracy and CRPS on \gift\ and \boom, with SQL on
\fev. \gift\ uses the geometric mean of metric-to-seasonal-naive ratios,
\boom\ uses shifted geometric mean normalized by DataDog's
seasonal-naive run, and \fev\ uses raw metric means and skill against its published
seasonal-naive baseline. We report relative changes, paired win/loss counts,
and paired Wilcoxon tests over matched benchmark configs/tasks.

\textbf{Evaluated method.}
All main results use \texttt{mixtop3-mase}, with $k=3$ and $m_{\min}=8$. The candidate pool and global design choices were chosen once
using \mchronos\ on a labelled $15$-config subset of the \gift\
validation split and then frozen. No test, \boom, or \fev\ data were used for
this choice, and nothing is retuned by model or configuration.
This global design choice is distinct from BGAA's inference-time selection,
which selects views from each configuration's observed history. For \gift\ and \boom, we use a light gate of one rolling origin per gate window for configurations with $m<8$, which account for a third of \boom\ tasks. For 16 of the 100 \fev\ tasks, the history before the first evaluation window is too short for a single backtest fold. For these tasks, we forecast the first 23 evaluation windows with the base model and score them as base forecasts in all reported results. Once their targets are observed, the remaining windows are forecast with BGAA, whose gate backtests only on these initial windows, is run once. These initial windows receive no benefit from BGAA.

\textbf{Baselines.}
The primary baseline is each model's unchanged-input forecast. We also compare against alternative \bgaa\ selection and aggregation policies, compute-matched random, noise, and identity controls, and published larger-model results used only for the model-scaling comparisons.

%% file: sections/06_main_results.tex
\section{Main results}
\label{sec:results}

\textbf{\bgaa\ improves all evaluated model--benchmark combinations.}
\Cref{tab:main} shows that \bgaa\ improves both point and probabilistic
forecasting across all $12$ model$\times$benchmark combinations without
per-model or per-benchmark retuning. MASE improves by $0.5$--$1.0\%$ on
\gift, $1.7$--$5.9\%$ on \boom, and $1.3$--$5.0\%$ on \fev.
These are benchmark-level aggregate improvements; individual configurations
can regress, as analyzed below.

\begin{table}[t!]
\centering
\small
\caption{\textbf{\bgaa\ improves performance across all $12$
model$\times$benchmark combinations.}
Frozen base versus the deployed \texttt{mixtop3-mase} policy, with no
per-model or per-benchmark retuning. Boldface marks the lower score within
each matched base--\bgaa\ comparison; it does not indicate a ranking across
models. $\Delta$\% reports relative change in MASE and the probabilistic
metric, respectively; negative values indicate improvement.}
\label{tab:main}
\resizebox{\linewidth}{!}{
\begin{tabular}{llrrrcr}
\toprule
benchmark & model &
base (MASE / CRPS$^{\dagger}$) &
\bgaa\ (MASE / CRPS$^{\dagger}$) &
$p$ (MASE / CRPS$^{\dagger}$) &
$\Delta$\% (MASE / CRPS$^{\dagger}$) &
win/loss \\
\midrule
\multirow{4}{*}{\gift}
 & \mchronos
 & 0.7043 / 0.4830
 & \textbf{0.6975 / 0.4778}
 & 0.004 / 0.004
 & $-0.97 / -1.07$
 & 55/38 \\
 & \mtirex
 & 0.6972 / 0.4791
 & \textbf{0.6940 / 0.4776}
 & 0.021 / 0.028
 & $-0.45 / -0.32$
 & 54/38 \\
 & \mrone
 & 0.7052 / 0.4820
 & \textbf{0.7018 / 0.4811}
 & $<$0.001 / $<$0.001
 & $-0.49 / -0.19$
 & 64/28 \\
 & \mtoto
 & 0.7032 / 0.4838
 & \textbf{0.6984 / 0.4806}
 & 0.023 / 0.042
 & $-0.69 / -0.66$
 & 60/32 \\
\midrule

\multirow{4}{*}{\boom}
 & \mchronos
 & 0.6547 / 0.3935
 & \textbf{0.6259 / 0.3803}
 & $<$0.001 / $<$0.001
 & $-4.39 / -3.38$
 & 4165/1691 \\
  & \mtirex
 & 0.6418 / 0.3855
 & \textbf{0.6191 / 0.3763}
 & $<$0.001 / $<$0.001
 & $-3.54 / -2.39$
 & 3917/1941 \\
 & \mrone
 & 0.6812 / 0.4071
 & \textbf{0.6410 / 0.3885}
 & $<$0.001 / $<$0.001
 & $-5.91 / -4.58$
 & 4283/1580 \\
 & \mtoto
 & 0.5853 / 0.3532
 & \textbf{0.5755 / 0.3501}
 & $<$0.001 / 0.008
 & $-1.67 / -0.89$
 & 3456/2299 \\
\midrule

\multirow{4}{*}{\makecell[l]{\fev\\{\scriptsize(MASE / SQL$^{\dagger}$)}}}
 & \mchronos
 & 1.6332 / 1.3416
 & \textbf{1.5513 / 1.2836}
 & $<$0.001 / 0.002
 & $-5.02 / -4.32$
 & 56/38 \\
 & \mtirex
 & 1.6192 / 1.3653
 & \textbf{1.5989 / 1.3436}
 & 0.005 / 0.009
 & $-1.26 / -1.59$
 & 54/39 \\
 & \mrone
 & 1.5852 / 1.3150
 & \textbf{1.5461 / 1.2845}
 & $<$0.001 / $<$0.001
 & $-2.46 / -2.32$
 & 65/27 \\
 & \mtoto
 & 1.6652 / 1.3881
 & \textbf{1.6444 / 1.3622}
 & 0.003 / 0.005
 & $-1.25 / -1.87$
 & 56/37 \\
\bottomrule
\end{tabular}
}
\end{table}

\begin{figure}[t]
  \centering
  \includegraphics[width=\textwidth]{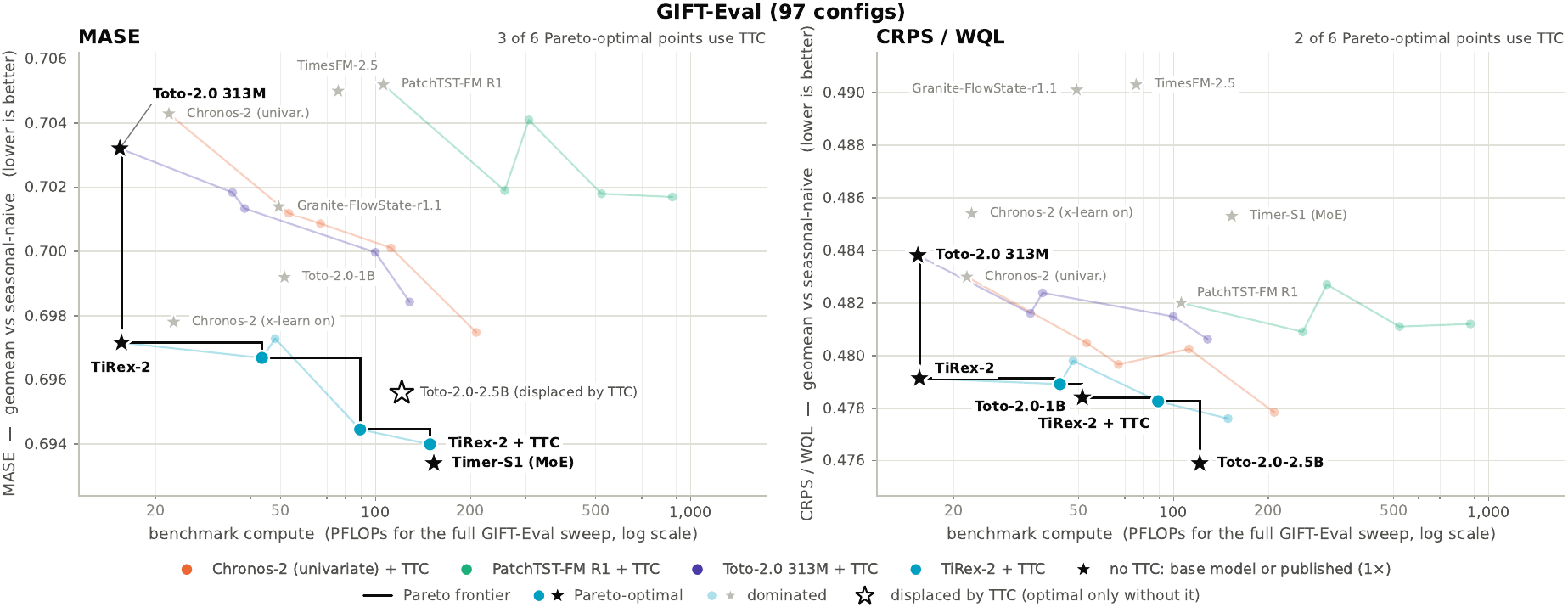}
  \caption{\textbf{Test-time compute improves MASE under estimated inference-FLOPs scaling.}
  On \gift\ ($97$ configurations), lines connect four \ttc\ budgets for each
  frozen model; stars denote published models at $1\times$ compute. Each rung
  uses the model's measured benchmark PFLOPs scaled by its measured
  forward-pass ratio
  $(\text{base}+\text{gate}+\text{apply})/\text{base}$.
  The CRPS ladders also improve over their corresponding
  bases.}
  \label{fig:frontier}
\end{figure}

\textbf{Test-time compute can dominate a larger model under the same
compute--accuracy comparison.}
\label{sec:vs-upgrade}
On \gift, the \texttt{high} \mtirex{} + \bgaa\ configuration achieves lower
MASE than \textsc{Toto-2-2.5B} with approximately $26\%$ less estimated
inference compute: $0.6945$ versus $0.6956$ MASE at approximately $89$ versus
$121$ PFLOPs (\cref{fig:frontier}). This is a pairwise dominance result under
our estimated inference-FLOPs accounting. Separately, on \boom, \bgaa\ on \mtoto\ achieves lower MASE than the published $1$B and $2.5$B members of the same family (\cref{tab:vs-upgrade}).

\Cref{fig:frontier} compares the four \emph{thinking modes}, which vary
test-time compute while keeping the BGAA procedure fixed
(\cref{tab:ladder}). Returns diminish with budget: \texttt{high} captures
$61$--$96\%$ of the \texttt{xhigh} MASE gain while using $54$--$78\%$ of its
compute. The CRPS ladders improve over their corresponding bases but do not
establish the same dominance result. Cross-learning-\emph{on}
\textsc{Chronos-2} also improves over its univariate counterpart at similar
base compute, illustrating that test-time compute is not the only useful
scaling direction. The FLOPs axis ignores batching and should be interpreted
as estimated inference-compute accounting rather than latency. Full accounting
is given in \cref{sec:app:ladder}. Complementary H100 measurements show that
\mchronos+\bgaa\ requires 7.02--14.3 seconds to
produce 10,000 forecasts across thinking budgets,
compared with 28.5 seconds for Toto-2-2.5B,
while using substantially less GPU memory
(0.48 versus 10.30 GiB).
Although the larger model retains better absolute
accuracy, these results show that historical verification
can be more efficient than deploying a much larger model
for high-throughput forecasting (\cref{tab:app:latency-all,tab:app:latency-buckets}).

\begin{wraptable}{r}{0.45\textwidth}
\centering
\small
\caption{\textbf{\bgaa\ on \mtoto\ ($313$M) exceeds the published larger
members of the same family on \boom\ MASE.}
Excerpt of the full recomputed leaderboard
(\cref{tab:app:boom-full}). This table compares accuracy; model size alone
does not establish inference-compute efficiency after adding \bgaa.}
\label{tab:vs-upgrade}
\resizebox{\linewidth}{!}{%
\begin{tabular}{@{}lrrl@{}}
\toprule
system & MASE & CRPS & note \\
\midrule
\textbf{\mtoto\ + \bgaa}
    & \textbf{0.575} & \textbf{0.350} & \textbf{ours} \\
\textsc{Toto-2-2.5B}
    & 0.581 & 0.349 & published \\
\textsc{Toto-2-1B}
    & 0.582 & 0.349 & published \\
\textsc{Toto-2-313M}
    & 0.585 & 0.351 & published \\
\mtoto\ base
    & 0.585 & 0.353 & our reproduction \\
\textsc{Toto (v1)}
    & 0.617 & 0.375 & published \\
\textbf{\mtirex + \bgaa}
    & \textbf{0.619} & \textbf{0.376} & \textbf{our run} \\
\textsc{TimesFM-2.5}
    & 0.632 & 0.382 & published \\
\mtirex\ base
    & 0.642 & 0.386 & our run \\
\textsc{Moirai-2-small}
    & 0.672 & 0.427 & published \\
\bottomrule
\end{tabular}%
}
\end{wraptable}

We additionally evaluate longer horizons
($H=96$--$720$) across five ETT/Weather datasets
(\cref{tab:app:long-horizon}). \bgaa\ improves both
normalized MSE and MAE on 18/20 dataset--horizon settings
for \mchronos\ and 15/20 for \mtoto, with gains persisting
at $H=720$, although they are not monotonic with horizon.

\textbf{Matched controls isolate the role of gating and structured views.}
\label{sec:controls}
Averaging the full candidate pool without gating worsens MASE by $4.97\%$,
whereas \bgaa\ improves it by $0.95\%$ on the same pool
(\cref{fig:gate}). On \boom, a lighter gate is active when fewer than
$m_{\min}=8$ backtest windows are available; median improvement is therefore
upto $0.3\%$ for $2929$ such tasks, rises to $0.44\%$ around $N=12$, and increases further for larger buckets (\cref{fig:where}).

\textbf{Compute-matched controls show that extra inference alone is insufficient.}
With $k=2$ and the pool and aggregation fixed, random selection worsens
MASE by $4.82\%$ and CRPS by $10.73\%$, while averaging jittered copies
worsens MASE by $0.03\%$ (\cref{tab:app:compute-matched}).
Restricted-family ablations further show that lookback-only and
offset-only pools recover just $30\%$ and $1\%$ of the full-pool MASE
gain, while cleaning-only views worsen MASE by $0.53\%$
(\cref{tab:app:controls}). More importantly, a 22-view pool curated
on \mchronos/\gift\ retains only $69\%$ of the full-pool gain when
transferred to other models and benchmarks, with a significant
disadvantage (\cref{tab:pool-restriction}). These results support
using a diverse, shared candidate pool rather than tuning a
restricted pool for individual models or benchmarks. The verifier
also remains useful with older evidence: backtests lagged by $4H$
still improve MASE by $0.41\%$, compared with $1.60\%$ at lag zero
(\cref{tab:app:temporal}). Further pool and aggregation ablations
are reported in \cref{tab:policy-ablation}.

\textbf{Historical selection also compares favorably with stronger
forecast-combination baselines.}
On a targeted 30-configuration subset, \bgaa\ achieves the largest
aggregate MASE and WQL improvements and substantially fewer severe
regressions than historical top-1 selection and cross-validated
convex combinations (\cref{tab:bgaa-combiners}). Although its mean
advantage over L2-regularized combinations is not statistically
significant, \bgaa\ requires far fewer candidate forecasts
at application time.

\begin{table}[t!]
\centering
\caption{Statistically significant regressions are rare under both policies. Per-configuration MASE improvement for \mchronos\ under
\texttt{mixtop3-mase} and \texttt{routemix05-crps}, relative to the frozen
base. Improvement is defined so that positive values are better; therefore,
negative values in the ``worst improvement'' columns denote degradation.
Full CDFs are in \cref{fig:noharm}.}
\label{tab:noharm}
\small
\resizebox{\linewidth}{!}{
\begin{tabular}{lrrrrrrr}
\toprule
 & & \multicolumn{3}{c}{final policy} & \multicolumn{3}{c}{conservative (\texttt{routemix05-crps})} \\
\cmidrule(lr){3-5}\cmidrule(lr){6-8}
benchmark & configs & equal or better & no sig. regression & worst & equal or better & no sig. regression & worst \\
\midrule
GIFT-Eval & 97 & 61\% & 95\% & $-8\%$ & 67\% & 93\% & $-6\%$ \\
BOOM & 6750 & 69\% & 92\% & $-199\%$ & 85\% & 95\% & $-98\%$ \\
fev-bench & 100 & 70\% & 94\% & $-7\%$ & 77\% & 97\% & $-6\%$ \\
\bottomrule
\end{tabular}}
\end{table}

\textbf{Most configurations do not regress.}
\Cref{tab:noharm} shows that $61\%$, $69\%$, and $70\%$ of configurations on
\gift, \boom, and \fev, respectively, are equal to or better than their frozen base
under the deployed policy, and $95\%$, $92\%$, and $94\%$ of configurations show no statistically significant regressions. The worst regressions are $8\%$ on \gift\ and
$7\%$ on \fev. On \boom, the larger negative tail is driven by a small number
of near-constant series for which the scaled metric is unstable. These are
empirical observations, not consequences of the backtest guarantee, which
does not extend to unseen forecast horizons or benchmark-level metrics
(\cref{sec:gate-main}).

%% file: sections/07_mechanism.tex
\section{Interpretation and supporting evidence}
\label{sec:mechanism}

The results above establish that \bgaa\ improves frozen forecasters, but do not
explain why \emph{re-presenting} the same observed history can produce useful
alternatives. We therefore ask: \emph{what kind of model behavior would make
structured views useful rather than merely different?} The argument below is
not required by \bgaa\ and is not an accuracy theorem. Instead, a restricted
transport result provides a concrete interpretation that yields empirical
predictions we test across the four models.

A deterministic view adds no observations. Nevertheless, a frozen forecaster
can return a different predictive distribution when the same history is
presented in another representation. Let $\Pi$ be an idealized distribution
over complete trajectories and let
$Q_P(\tau\mid x_{1:T})\in\mathbb{R}^{H}$ denote the conditional future
$\tau$-quantiles under trajectory distribution $P$, and write
$f(x_{1:T},H)_{\cdot,\tau}\in\mathbb{R}^{H}$ for the column of the frozen
forecast at level $\tau$.

\begin{assumption}[PFN idealization]
\label{ass:pfn-main}
For an observed context $x_{1:T}$, the frozen model returns the conditional
quantiles under $\Pi$:
$ f(x_{1:T},H)_{\cdot,\tau}=Q_{\Pi}(\tau\mid x_{1:T}), \qquad \tau\in\mathcal L.$
\end{assumption}

Now let $\psi\colon\mathbb{R}\to\mathbb{R}$ be a fixed strictly increasing bijection
applied pointwise, and let $\Pi_\psi$ be the trajectory distribution obtained by
applying $\psi^{-1}$ pointwise to trajectories drawn from $\Pi$. Let
$(\Pi_\psi)_{1:T}$ denote its marginal over contexts.

\begin{proposition}[Transport identity]
\label{prop:transport-main}
Under \cref{ass:pfn-main},
$\psi^{-1}\!\bigl(f(\psi(x_{1:T}),H)_{\cdot,\tau}\bigr) =
Q_{\Pi_\psi}(\tau\mid x_{1:T})$
for $(\Pi_\psi)_{1:T}$-almost every context $x_{1:T}$ and $\tau\in\mathcal L$.
\end{proposition}

For this restricted class of transformations, forecasting a transformed
history and mapping its quantiles back is equivalent to prediction under a
transported trajectory distribution. This does not imply greater accuracy, nor
do we claim that released TSFMs satisfy \cref{ass:pfn-main} exactly. None of
our non-identity deployed views satisfies all of the proposition's conditions:
Rank-to-Gaussian is context-fitted and clamped, crops and offsets alter the
observations presented to the model, and differencing is not pointwise and uses
an approximate quantile-path reconstruction. The full statement, proof, and
scope are in \cref{sec:transport}. \Cref{prop:transport-main} instead provides
a tractable case where re-presenting the same observations changes the
effective predictive distribution without adding information, motivating us to
ask whether \bgaa's broader views induce similarly structured changes.

This yields a testable hypothesis: if structured views systematically change
a frozen model's forecasting behavior rather than acting as arbitrary ensemble
noise, then \emph{which} views help should depend on the model, individual views
should induce repeatable changes in representation space, and these changes
should produce recognizable forecast-level signatures. We test these
predictions below; full protocols and results are in
\cref{sec:app:mechanism}.

\textbf{Useful view families depend on both the model and the data.}
\label{sec:exp2}
Across the six model pairs, Kendall's $\tau_{\mathrm{K}}$ between
per-family solo-gated benefit vectors ranges from $0.01$ to $0.35$,
indicating weak agreement on which transformations are beneficial
(\cref{tab:app:tau}). For example, differencing is strongest for
\mchronos\ ($-0.85\%$ MASE) but slightly hurts \mtirex\
($+0.01\%$), while \texttt{lookback} is most effective for
\mrone\ and \mtoto\ (\cref{tab:app:fingerprint}).
Transformation preferences also vary across domains
(\cref{fig:fingerprint}) and with series characteristics such as
seasonality, trend strength, autocorrelation, and volatility
(\cref{tab:app:char-families}). However, these statistical
relationships do not provide a reliable alternative to historical
verification: feature-based routers trained leave-one-configuration-out
fail to outperform historical top-1 selection at matched application
cost (\cref{tab:app:char-router}).
Furthermore, a pool curated on \mchronos/\gift\ excludes
$30$--$41\%$ of the views selected in other model--benchmark
pairs (\cref{tab:pool-views}). Together, these results motivate
using a diverse, shared candidate pool with historical verification
rather than relying on model-specific pool tuning or statistical
characteristics alone.

\textbf{Views induce systematic, model-specific changes.}
\label{sec:exp9}
Across $626$ \gift\ contexts, we compare view-induced displacements with rollout samples from each model, used as a proxy for its learned forecasting behavior. Some views move contexts consistently: \texttt{diff} and \texttt{smooth} have directional consistency of $0.75$--$0.78$, versus $0.04$--$0.07$ for crops and offsets (\cref{tab:app:directionality}). Different views are also non-redundant, with mean absolute pairwise displacement cosine of $0.36$--$0.48$ (\cref{tab:app:viewview}). Finally, \texttt{diff} aligns with a mode of each model's rollout distribution (cosine $+0.92$ to $+0.97$), but with different modes across models (\cref{tab:app:viewmode,fig:priorspace}). These patterns indicate repeatable, model-specific effects rather than arbitrary perturbations.

\textbf{The structure persists in forecast space.}
\label{sec:exp10}
The final prediction is that a view's effect should remain recognizable in the
forecast itself. On a $15$-configuration subset, \texttt{diff} widens
predictive intervals by a log-ratio of $+2.10$ to $+2.25$, with the same sign
on $14$--$15$ of $15$ configurations (\cref{tab:app:signatures}). This is a
property of the complete forecasting view, including its cumulative-sum
pull-back, rather than only of the transformed input. The broader signature
analysis in \cref{sec:app:exp10} shows the same family-level structure across
the other forecast statistics and models.

Taken together, this evidence
forms a consistent chain: useful views differ by model, individual views move
contexts in repeatable and distinct directions, and those changes remain
visible in the resulting predictive distributions. This evidence strengthens
the interpretation that structured views expose different behaviors of the
same frozen TSFM rather than merely being generic ensembling.
It does not establish causal prior transport: the rollout distributions are proxies for learned forecasting behavior, and we do not intervene on the models' pretraining distributions.

%% file: sections/08_limits.tex
\section{Limitations and future directions}
\label{sec:limits}

This section focuses on the main limitation revealed
by our diagnostics: historical backtesting supports reliable config-level selection, but provides much weaker guidance for individual forecast windows. We also discuss what this means for future test-time compute methods.

The backtest guarantee in \cref{sec:safety} is deliberately limited:
it controls loss on historical gate windows, not on future forecasts.
The key question is therefore whether historical rankings remain
informative at deployment. At the configuration level, they often do:
on \mchronos/\gift, the median Spearman correlation between backtest
and realized view rankings is $0.63$ per configuration, but only
$0.25$ per forecast window (\cref{fig:capturability-main}).
Configuration-level selection also benefits from pooling historical
evidence across series. On 41 \gift\ configurations, pooled selection
outperforms independent per-series selection on average across all
three tested models, particularly when individual series have limited
backtest evidence (\cref{tab:app:pooling}).
Together, these findings support pooling historical evidence for
more reliable configuration-level verification, while highlighting
the harder problem of identifying which view will help on a
particular future window using only its observed context.

Hindsight selection confirms that the best-performing view can vary substantially across windows. The backtest's top-$1$ contains the realized best view for $19\%$ of configurations but only about $11\%$ of individual windows, and its per-window top-$1$ recovers only about $27\%$ of the hindsight-available MASE gain (\cref{fig:topk}). Cross-fitted analyses likewise indicate that better fixed per-configuration choices remain possible (\cref{tab:capturability}). These quantities are diagnostic rather than deployable targets: hindsight uses the realized future, and the cross-fitted splits remain temporally dependent (\cref{sec:app:exp8}). They therefore show where the current verifier loses information, not how much improvement a future context-only selector could necessarily recover.

Our evaluation is also limited to univariate forecasting, and \bgaa\ incurs additional inference cost, primarily from evaluating candidate views on historical windows (\cref{tab:modes}). The main open question is whether information available before the forecast can support more accurate window-specific selection without making verification prohibitively expensive. This could require better window-level signals, better candidate views, or both. Other promising directions include: \emph{learned verifiers} that predict useful views from context and view features; \emph{verification-set design} that selects informative past windows; and \emph{adaptive budgets} that spend more compute when views disagree.

%% file: sections/10_conclusion.tex
\section{Conclusion}
\label{sec:conclusion}

In this paper, we show that test-time compute can improve frozen time series foundation models without retraining or future labels. \bgaa\ generates forecasts from multiple views of the same context and uses a leak-free rolling-origin backtest to select which views to aggregate. With one fixed pool and policy, \bgaa\ improves all $12$ model--benchmark combinations, with MASE gains of $0.5$--$1.0\%$ on \gift, $1.7$--$5.9\%$ on \boom, and $1.3$--$5.0\%$ on \fev. Mechanism experiments show structured, model-dependent view effects and consistent forecast signatures, supporting our prior-transport interpretation. The main remaining limitation is window-level selection: backtest rankings are strong at the configuration level but much weaker per window, with median Spearman correlation $0.63$ versus $0.25$ for \mchronos\ on \gift. This motivates better window-specific selection alongside improved candidate views.

\section*{AI Use Statement}

We used large language models as coding and writing assistants during the
development of \bgaa\ and while writing the paper. The initial base implementation of \bgaa\ was developed primarily by the authors. LLMs were used to scale
the implementation, reproduce benchmark pipelines, configure environments,
refactor code, and support model- and benchmark-specific integration. LLMs
were also used for \mtirex-specific kernel modifications to improve evaluation
efficiency on H100 GPUs. The paper was primarily written by the authors. LLMs were used to review the text, check consistency with the final
implementation, and compress the manuscript. The compression was done line-by-line or
paragraph-by-paragraph rather than by regenerating the manuscript as a whole.
Every resulting line was reviewed by the authors before inclusion,
and any incorrect, ambiguous, or overstated language introduced during
compression was revised manually. We take full responsibility for
the implementation, experiments, analysis, theoretical claims, and final text
of the paper.

%% file: appendix/A_proofs.tex
\section{Theory and proofs}
\label{sec:app:proofs}

\Cref{sec:safety} proves the backtest-dominance property used in
\cref{sec:gate-main}. \Cref{sec:convexity} gives the aggregation results behind
Vincent and CDF-mixture aggregation and states which aggregator each score
covers. \Cref{sec:transport} formalizes the transported-prior interpretation
tested in \cref{sec:mechanism}.

\begingroup
\let\section\subsection
\let\subsection\subsubsection
\input{theory/01_safety_theorem}
\input{theory/02_aggregation_convexity}
\input{theory/03_prior_transport}
\endgroup

%% file: theory/01_safety_theorem.tex
\section{The backtest referee}
\label{sec:safety}

This section gives the formal statement and proof of the gate guarantee
summarized in \cref{sec:gate-main}.

Fix a dataset configuration $c$. Let $\mathrm{id}$ denote the identity view,
which returns the base forecast, and $g\in\mathcal G$ a candidate view. On gate
window $w_j$, let $\widehat L_j(g)\ge 0$ be the loss of view
$g\in\mathcal G\cup\{\mathrm{id}\}$, averaged over the
$n_f$ rolling forecast origins in that window. The gate uses
absolute error of the median forecast for \texttt{*-mase} policies and the
mean pinball loss over $\mathcal L$ for \texttt{*-crps} policies. Windows with
$\widehat L_j(\mathrm{id})=0$ are excluded; let $\mathcal J$ be the set of
usable windows and $m=|\mathcal J|$ their number. Define
the mean per-window loss relative to the base as
\[
\widehat L_{\mathrm{rel}}(g)
=
\frac{1}{m}\sum_{j\in\mathcal J}
\frac{\widehat L_j(g)}{\widehat L_j(\mathrm{id})},
\qquad
\widehat L_{\mathrm{rel}}(\mathrm{id})=1.
\]
The relative improvement of candidate $g$ is
$\widehat{\epsilon}(g)=1-\widehat L_{\mathrm{rel}}(g)$. The margin rule requires
$\widehat{\epsilon}(g)\ge\epsilon_{\min}$, where $\epsilon_{\min}>0$ is the minimum improvement, and a
win on at least a fraction $\phi$ of usable windows. Top-$k$ keeps the $k$
candidates with largest positive $\widehat{\epsilon}(g)$. Both rules require at least
$m_{\min}$ usable windows.

Let $\widehat{\mathcal A}_c$ denote the admitted candidate set and $f_{\mathrm{agg}}$ the resulting
aggregate, with loss $\widehat L_j(\mathrm{agg})$. The aggregate assigns weight $\lambda_0\ge0$ to the base and nonnegative
weights $\lambda_g$ to admitted candidates, with
\[
\lambda_0+\sum_{g\in\widehat{\mathcal A}_c} \lambda_g = 1.
\]
If $\widehat{\mathcal A}_c=\varnothing$, the policy returns the base forecast. The result below
depends only on the gate losses and aggregation rule, not on the underlying
forecaster.

\begin{theorem}[Backtest dominance]
\label{thm:backtest}
Suppose that, on every usable gate window, the aggregate loss satisfies
\[
\widehat L_j(\mathrm{agg})
\le
\lambda_0\widehat L_j(\mathrm{id})
+
\sum_{g\in\widehat{\mathcal A}_c} \lambda_g\widehat L_j(g).
\]
This holds for Vincent averaging under pinball loss or absolute error of the
median forecast. More generally, if $\ell$ is exact CRPS, CDF mixing also
satisfies the required inequality. If
$\widehat{\mathcal A}_c\neq\varnothing$, let
\[
\epsilon'=\min_{g\in\widehat{\mathcal A}_c}\widehat{\epsilon}(g)
\]
be the smallest improvement among admitted candidates. Then
\[
\widehat L_{\mathrm{rel}}(\mathrm{agg})
\le
1-(1-\lambda_0)\epsilon'
\le
1
=
\widehat L_{\mathrm{rel}}(\mathrm{id}).
\]
Under the margin rule, $\epsilon'\ge\epsilon_{\min}$; under top-$k$,
$\epsilon'>0$. If $\widehat{\mathcal A}_c=\varnothing$, the aggregate equals the base.
\end{theorem}

\begin{proof}
For a non-empty admitted set, the assumed aggregation inequality holds on each
usable gate window. For Vincent averaging, it follows from convexity of
pinball loss; absolute error of the median forecast is proportional to the
$\tau=\tfrac12$ pinball loss. For CDF mixing under exact CRPS, it follows from
\cref{thm:ambiguity}. Since $\widehat L_j(\mathrm{id})>0$ on every usable window, dividing
by $\widehat L_j(\mathrm{id})$ and averaging gives:
\[
\widehat L_{\mathrm{rel}}(\mathrm{agg})
\le
\lambda_0+\sum_{g\in\widehat{\mathcal A}_c}\lambda_g\widehat L_{\mathrm{rel}}(g)
=
1-\sum_{g\in\widehat{\mathcal A}_c}\lambda_g\widehat{\epsilon}(g)
\le
1-(1-\lambda_0)\epsilon'.
\]
The empty-set case follows directly from returning the base forecast.
\end{proof}

\Cref{thm:backtest} applies to the gate windows and to the mean per-window
relative loss $\widehat L_{\mathrm{rel}}$ used by the gate. It does not guarantee improvement
under the benchmarks' reported aggregation rules, which generally differ from
$\widehat L_{\mathrm{rel}}$ and can weight configurations or forecast errors differently. It also
does not guarantee
improvement on the future forecast horizon without assumptions linking the
backtest to that horizon. Thus, we measure horizon performance directly in \cref{tab:main,tab:noharm}. \cref{tab:app:temporal} probes
the backtest-to-horizon link itself: when the gate's evidence is restricted to
observations at least $\mathrm{lag}\times H$ before the forecast start
($\mathrm{lag}\le 4$), the horizon MASE gain of the Vincent top-2 policy stays at
0.4--1.7\% even though the admitted set changes substantially
(Jaccard 0.44--0.59 against lag 0), on a 15-config \gift\ test subset for \mchronos.

Note: The theorem covers the Vincent \texttt{top3-mase} policy. It does not
cover the deployed \texttt{mixtop3-mase}, since CDF-mixture quantiles are not
convex combinations of member quantiles. Similarly, the \texttt{*-crps} gate
uses mean pinball loss over $\mathcal L$ rather than exact CRPS, so CDF mixing under that
gate does not inherit the exact-CRPS guarantee. The positive margin
$\epsilon_{\min}$, win-fraction threshold $\phi$, and minimum-window requirement
$m_{\min}$ are robustness checks against small or noisy backtests, not
finite-sample guarantees on the forecast horizon.

%% file: theory/02_aggregation_convexity.tex
\section{Aggregation}
\label{sec:convexity}

This section gives the full aggregation result summarized in
\cref{sec:gate-main}.

Let $F_1,\dots,F_n$ be the CDFs of $n$ forecast distributions, with weights
$\lambda_i\ge0$ and $\sum_{i=1}^n \lambda_i=1$, where the weights are determined
independently of the realized outcome $y\in\mathbb{R}$. Multi-horizon scores average the corresponding
one-step scores, so the result below applies at each horizon. In our
implementation, each $F_i$ is the interpolated CDF constructed from the
model's nine-quantile forecast (\cref{sec:app:views}). Vincent averaging
combines quantiles level by level, while CDF mixing averages the predictive
distributions.

\begin{theorem}[CRPS decomposition for CDF mixtures]
\label{thm:ambiguity}
Let
\[
\bar F=\sum_{i=1}^n \lambda_iF_i
\]
be the mixture CDF, and assume the member forecasts have finite CRPS. Then,
for every realized outcome $y$,
\[
\CRPS(\bar F,y)
=
\sum_{i=1}^n \lambda_i\CRPS(F_i,y)
-
\underbrace{
\sum_{i=1}^n \lambda_i\int_{\mathbb R}
\bigl(F_i(z)-\bar F(z)\bigr)^2\,\dg z
}_{\text{diversity term}\ge0},
\]
where $z$ indexes the forecast value. The diversity term depends only on the
forecast distributions, not on $y$.
\end{theorem}

\begin{proof}
For outcome $y$, define
\[
a_i(z)=F_i(z)-\mathbf 1\{y\le z\},
\qquad
\bar a(z)=\sum_{i=1}^n \lambda_i a_i(z),
\]
where $\mathbf 1\{\cdot\}$ is the indicator function. Since
$\bar a(z)=\bar F(z)-\mathbf 1\{y\le z\}$, the weighted-mean identity gives
\[
\bar a(z)^2
=
\sum_{i=1}^n \lambda_i a_i(z)^2
-
\sum_{i=1}^n \lambda_i\bigl(a_i(z)-\bar a(z)\bigr)^2.
\]
Moreover, $a_i(z)-\bar a(z)=F_i(z)-\bar F(z)$. Integrating both sides over
$z\in\mathbb R$ and using the definition of CRPS gives the result.
\end{proof}

Since the diversity term is non-negative, the CRPS of a CDF mixture is no
worse than the weighted average CRPS of its members for every outcome, with
equality when all positive-weight member CDFs agree almost everywhere. The
term can also be computed without labels, giving a label-free measure of
forecast disagreement and motivating the spread-routed policy in
\cref{sec:app:views}.

This identity applies to the mixture $\bar F$ itself, before we read it back
as a nine-quantile forecast for deployment. It is specific to exact CRPS. 

The guarantee in \cref{thm:backtest} requires the aggregation rule to be
convex in the score used by the gate. Vincent averaging satisfies this
condition for the absolute-error score used by \texttt{*-mase} and for
pinball loss. CDF mixing satisfies it for exact CRPS. Our \texttt{*-crps}
policies instead rank views using mean pinball loss over $\mathcal L$, so their CDF-mixture
variants do not inherit the exact-CRPS guarantee.

The deployed \texttt{mixtop3-mase} policy similarly combines a MASE gate with
CDF mixing, and thus does not inherit formal backtest dominance on either MASE
or CRPS. We use it because it gives the best mean MASE in the policy ablation. The corresponding Vincent policy, \texttt{top3-mase}, uses the same
gate and admitted set and is covered by the backtest guarantee in
\cref{thm:backtest}.

%% file: theory/03_prior_transport.tex
\section{Restricted transport interpretation}
\label{sec:transport}

A trajectory is $z=(z_1,\dots,z_{T+H})\in\mathbb R^{T+H}$, with context
$z_{1:T}$ and future $z_{T+1:T+H}$. Let $\Pi$ denote an idealized trajectory distribution associated with the
frozen model, $x_{1:T}\in\mathbb R^T$ a context, and
$\mathcal L$ the reported quantile levels. We write
$Q_{P}(\tau\mid x_{1:T})$ for the vector of conditional $\tau$-quantiles
of the future given $z_{1:T}=x_{1:T}$ under trajectory distribution $P$, and
$f(x_{1:T},H)_{\cdot,\tau}$ for the corresponding
model forecast, the column of $f(x_{1:T},H)$ at level $\tau$.

\noindent\textbf{Assumption~\ref{ass:pfn-main} (PFN idealization).}
Given a context $x_{1:T}$, the frozen model returns the conditional future
quantiles under $\Pi$:
\[
f(x_{1:T},H)_{\cdot,\tau}
=
Q_{\Pi}(\tau\mid x_{1:T}),
\qquad \tau\in\mathcal L.
\]

The PFN interpretation motivates this idealization \citep{muller2022}.
Pretrained probabilistic forecasters such as Chronos motivate asking whether
an analogous interpretation is useful for TSFMs \citep{ansari2024}, but we do
not assume that it holds exactly. We use it only to motivate a mechanism with
observable predictions, which we test in \cref{sec:mechanism}.

\medskip
\noindent\textbf{Proposition~\ref{prop:transport-main} (Transport identity).}
Let $\psi\colon\mathbb R\to\mathbb R$ be a fixed strictly increasing bijection applied
pointwise, and define its trajectory-level extension
\[
\Psi(z)=(\psi(z_1),\dots,\psi(z_{T+H})).
\]
Let $\Pi_\psi$ be the distribution of $\Psi^{-1}(z)$ for $z\sim\Pi$, and
let $(\Pi_\psi)_{1:T}$ denote its marginal distribution over contexts. Here,
$\psi(x_{1:T})$ and $\psi^{-1}(\cdot)$ are also applied coordinatewise. Under
\cref{ass:pfn-main}, for $(\Pi_\psi)_{1:T}$-almost every context $x_{1:T}$ and every
$\tau\in\mathcal L$,
\[
\psi^{-1}\!\bigl(f(\psi(x_{1:T}),H)_{\cdot,\tau}\bigr)
=
Q_{\Pi_\psi}(\tau\mid x_{1:T}).
\]
Thus, forecasting the transformed context and mapping its quantiles back is
equivalent to prediction under the transported trajectory distribution
$\Pi_\psi$.

\begin{proof}
Let $z\sim\Pi_\psi$ and define $\tilde z=\Psi(z)$. By the definition of
$\Pi_\psi$,
\[
\tilde z\sim\Pi.
\]
Since $\Psi$ applies $\psi$ coordinatewise,
\[
\tilde z_{1:T}=\psi(z_{1:T}),
\qquad
\tilde z_{T+1:T+H}=\psi(z_{T+1:T+H}).
\]

Because Euclidean trajectory space is standard Borel, regular conditional
distributions exist. Now fix a context $x_{1:T}$. Since $\psi$ is a bijection,
$z_{1:T}=x_{1:T}$ corresponds exactly
to $\tilde z_{1:T}=\psi(x_{1:T})$. Also,
\[
z_{T+1:T+H}=\psi^{-1}(\tilde z_{T+1:T+H})
\]
coordinatewise. Therefore, for $(\Pi_\psi)_{1:T}$-almost every $x_{1:T}$, the
conditional
distribution of $z_{T+1:T+H}$ given $z_{1:T}=x_{1:T}$ is obtained by applying $\psi^{-1}$
coordinatewise to the conditional distribution of $\tilde z_{T+1:T+H}$ given
$\tilde z_{1:T}=\psi(x_{1:T})$.

Because $\psi$ is strictly increasing, conditional quantiles transform
coordinatewise:
\[
Q_{\Pi_\psi}(\tau\mid x_{1:T})
=
\psi^{-1}\!\left(
Q_{\Pi}(\tau\mid \psi(x_{1:T}))
\right).
\]

Finally, \cref{ass:pfn-main} gives
\[
f(\psi(x_{1:T}),H)_{\cdot,\tau}
=
Q_{\Pi}(\tau\mid \psi(x_{1:T})).
\]
Substituting this into the previous display yields
\[
Q_{\Pi_\psi}(\tau\mid x_{1:T})
=
\psi^{-1}\!\bigl(f(\psi(x_{1:T}),H)_{\cdot,\tau}\bigr),
\]
which is the claimed identity.
\end{proof}

\paragraph{Observation-map views.}
\label{rem:crop}
\label{rem:orbit}
Lookback crops and patch-phase offsets do not satisfy the transport
construction. Instead, they change the information presented to the model
without changing the frozen forecaster. The gate treats them like any other
candidate, and
\cref{thm:backtest} does not require a transport interpretation.

\paragraph{Scope of the transport interpretation.}
\label{rem:scope}
\Cref{prop:transport-main} applies only to fixed pointwise strictly increasing
bijections. The identity view satisfies these conditions trivially, but none
of our non-identity deployed views satisfies them all exactly.
Rank-to-Gaussian is the closest case, but it is fitted to the current context
and clamps forecasts to the observed range. Lookback and offset views instead
change the observation map. The remaining families are structured
perturbations motivated by the same picture rather than exact transported
trajectory distributions.

Differencing requires one additional caveat. Its pull-back constructs a
level-$\tau$ path by cumulatively summing level-$\tau$ increment forecasts.
Since the quantile of a sum is not generally the sum of its marginal
quantiles, the resulting fan need not correspond to a transported posterior
predictive. This can widen intervals with horizon, so part of the
\texttt{diff} effect in \cref{tab:app:signatures} comes from the pull-back
itself. This does not affect \cref{thm:backtest}, which treats each view as a
forecast-generating policy.

\paragraph{Selection over transported views.}
\label{rem:eb}
For views satisfying \cref{prop:transport-main}, the backtest can be interpreted as
selecting among forecasts induced by different transported trajectory
distributions. We use this exact case to motivate the broader structured-view
interpretation. The gate
ranks views by predictive loss; its weights are not Bayesian model
probabilities.

%% file: appendix/B_views.tex
\section{The view catalogue}
\label{sec:app:views}

\paragraph{What makes a legal view.}
A view must (i) use no information beyond the available context, (ii) return a
forecast fan on the original data scale, through pull-back, restriction, or
output post-processing, and (iii) change something that is not removed by the
model's per-instance normalization. Requirement (iii) excludes affine warps:
with RevIN-style normalization, rescaling and shifting the context are exact
no-ops, which we verified empirically before constructing the pool.

\begin{table}[h]
\centering
\small
\caption{\textbf{The deployed pool of views}. Tiers are strictly nested,
crop $\subset$ micro $\subset$ ideas $\subset$
ideas+decomp, which makes \cref{tab:policy-ablation}'s left
column a decomposition rather than a list. ``\#'' counts \emph{views}; ``passes''
counts forward passes per window, which differs when a view is seeded
(\texttt{jitter}/\texttt{jitternorm} average $3$ seeded draws) or sequential
(\texttt{recursive\_f0.5} needs $2$ calls, \texttt{recursive\_f0.25} needs $4$).
Thus, the gate uses up to $58$ passes for the identity plus $51$ non-identity candidates ($52$ total views); where the decomposition family deduplicates to a single view, e.g.\ period-$1$ configs, this becomes $56$ passes and $50$ total views.}
\label{tab:app:pool}
\begin{tabular}{llrrp{5.1cm}}
\toprule
tier & family & \# & passes & degrees, and how the fan is inverted \\
\midrule
crop  & identity        & 1  & 1  & -- (cached base forecast) \\
crop  & lookback        & 18 & 18 & keep the last $\alpha T$ samples, $\alpha\in\{$0.98, 0.95, 0.925, 0.9, 0.875, 0.85, 0.8, 0.75, 0.7, 0.625, 0.55, 0.5, 0.4375, 0.375, 0.3125, 0.25, 0.1875, 0.125$\}$; restriction map, no fan inversion \\
crop  & offset          & 14 & 14 & shift the patch grid by $s\in\{$1, 2, 3, 4, 6, 8, 10, 12, 14, 16, 20, 24, 32, 48$\}$ samples ($\texttt{patch\_len}=16$); restriction map \\
micro & jitter          & 1  & 3  & additive $\sigma{=}0.002{\cdot}\mathrm{std}$ noise, $3$ seeded draws averaged in quantile space; median-neutral \\
micro & jitternorm      & 1  & 3  & as above, rescaled to preserve the context's scale \\
micro & winsor          & 1  & 1  & clip at the $0.5\%$ tails \\
ideas & diff            & 1  & 1  & first differencing; cumulative-sum inversion along each quantile path (comonotone pull-back, an approximation; \cref{rem:scope}) \\
ideas & sdiff           & 1  & 1  & seasonal differencing at the config's period $m$; seasonal cumulative-sum inversion \\
ideas & multideseason   & 1  & 1  & remove the multiplicative seasonal profile at period $m$; multiply it back onto every quantile \\
ideas & seasontile      & 2  & 2  & tile the last seasonal period, $K\in\{1,2\}$ \\
ideas & fft             & 3  & 3  & keep the top-$k$ frequencies, $k\in\{4,8,12\}$; low-pass of the context only \\
ideas & rankgauss       & 1  & 1  & empirical ranks $\to$ Gaussian order statistics; invert through the context's empirical quantile function, clamped to its observed range; context-fitted, degree $0.0$\footnotemark \\
ideas & nonneg          & 1  & 0  & project the fan onto $[0,\infty)$; reuses the cached identity pass \\
ideas & supportfloor    & 1  & 0  & project the fan onto the context's observed support; reuses the identity pass \\
ideas & recursive       & 2  & 6  & short-step rollout composed to $H$, step fraction $f\in\{0.5,0.25\}$ ($2$ and $4$ sequential calls) \\
\midrule
decomp & decomp         & $\le3$ & $\le3$ & seasonal decomposition at the base period and its sub/super harmonics $\{m, m/2, 2m\}$; deduplicated, so $1$--$3$ views per config \\
\bottomrule
\end{tabular}
\end{table}

\footnotetext{\texttt{rankgauss} falls back to the identity forecast when the
context has fewer than $16$ observations, contains non-finite values, or has
zero range.}

\paragraph{Two views are free.}
\texttt{nonneg} and \texttt{supportfloor} post-process the identity fan rather
than re-running the model, so they require \emph{zero} additional forward
passes in both the gate and apply steps once the identity forecast is memoized.
Together they account for $16\%$ of rank-weighted admits, giving them the
largest admit share per additional pass in the pool and motivating their
inclusion in the cheapest thinking mode in \cref{tab:modes}.

\paragraph{Families that were tried and rejected.}
For completeness, we also report families that were evaluated but not included
in the deployed candidate set. These include detrending, STL decomposition,
coarse resampling, motif retrieval (using either raw or model-embedding
distance), regime/changepoint crops, anomaly cleaning, ARIMA-residual views,
log-difference and reflected log-difference, moving-average and EWMA
smoothing, despiking, and power transforms. Smoothing is informative:
\cref{sec:app:exp9} finds that its context displacement points away from the
measured modes of the models' rollout distributions, a pattern consistent with
moving contexts away from regions favored by the frozen model.

\paragraph{How rejection was decided, and why it is not post-hoc.}
Every family, whether deployed or rejected, was evaluated using the same
procedure before deployment decisions were made. We constructed the family's
full degree grid, gated it \emph{solo} (that family plus the identity) using the leak-free backtest, and scored the resulting policy against the
\mchronos\ base on a labelled $15$-config subset of the \gift\
\emph{validation} split. A family was admitted to the deployed tier only if
its solo-gated $\Delta$MASE was negative and its worst-config regression was
bounded on this \mchronos\ \emph{validation} subset. This is the only stage at which
labels informed a design choice: the pool was then frozen before any test-split
evaluation or evaluation with the other models, and neither \boom\ nor \fev\
was used for an admit/reject decision. Rejected families were evaluated under
the same protocol and continue to be measured by it; harmful variants
(\texttt{smooth}, \texttt{ewma}, and \texttt{winsor} above the smallest
degree) remain in the \emph{ranking} pool but are excluded at deployment.
Their reported results are therefore regenerated under the same protocol
rather than taken from a one-off sweep. \Cref{sec:exp2} reports the resulting
per-family fingerprints across all four models.

\paragraph{Two families appear only in the fingerprint.}
\Cref{tab:app:fingerprint} also gates two fixed-pool families that are not in the
deployed tier. \texttt{changepoint} is the regime crop listed above: it finds the
strongest single mean shift in the most recent $1-\rho$ of the context,
$\rho\in\{0.2,0.3,0.4,0.5,0.6,0.7\}$, and starts the context at that break if the
shift exceeds $0.5$ standard deviations. \texttt{warp} resamples the context
by linear interpolation to $\mathrm{round}(wT)$ points and back to $T$, with
$w\sim\mathcal{U}[1-f,1+f]$, $f\in\{0.01,0.02,0.035,0.05,0.075,0.1,0.15\}$, and
averages $3$ seeded draws. Both resamplings span the whole context, so the round
trip keeps every time index in place.

\paragraph{Aggregation details.}
Let $\admit$ be the admitted set, $\qfan_0$ the identity fan, and $w_0$ the
weight on the base. The first three rules assign uniform weight to the base and
admitted members, giving $w_0=1/(|\admit|+1)$; the soft-gate instead reweights
the admitted set.
\begin{itemize}[leftmargin=1.2em,itemsep=1pt,topsep=2pt]
  \item \textbf{Vincent (quantile averaging).}
    $\bar q(\tau)=\sum_i w_i q_i(\tau)$ at each level $\tau$. Since each member
    is monotone in $\tau$, their weighted average is also a valid quantile fan.
  \item \textbf{CDF mixture.}
    $\bar F=\sum_i w_i F_i$, where each $F_i$ is the interpolated CDF of its
    nine-quantile fan. We invert the mixture on a shared grid by pooling and
    sorting the members' quantiles, evaluating $\bar F$ on this grid, and
    reading back the $9$ deployed levels by linear interpolation.
    \Cref{thm:ambiguity} applies exactly to $\bar F$ before this nine-quantile
    read-back. Its quantiles are not convex combinations of member quantiles,
    so the MASE-gated mixture is not covered by \cref{thm:backtest} on MASE or
    CRPS (\cref{sec:convexity}); the Vincent policy over the same admitted set,
    \texttt{top3-mase}, is the variant covered by the historical-backtest guarantee in \cref{thm:backtest}.
  \item \textbf{Spread-routed mixture} (\texttt{routemix} $\theta$).
    For each window, compute the pairwise-CRPS spread
    $s=\tfrac{2}{k(k-1)}\sum_{i<j}\crps(F_i,F_j)$ among admitted members,
    normalized by the window's own label scale. Use the CDF mixture when
    $s>\theta$ and Vincent otherwise. We report
    $\theta\in\{0.3,0.5,1.0\}$, with $\theta=0.5$ as the reported safe
    alternative. The statistic is a label-free disagreement measure closely
    related to the diversity term in \cref{thm:ambiguity}, but differs in both
    the routing set and scale normalization.
  \item \textbf{Temperature soft-gate.}
    $w_i\propto\exp(-\bt(g_i)/T)$ over the admitted set, with $T$ set to the
    spread of admitted backtest scores. We report this policy for completeness
    but do not deploy it: it gives the best CRPS and the worst MASE in
    \cref{tab:policy-ablation}.
\end{itemize}
Policies are named \texttt{<rule><k>-<objective>}, where the suffix denotes the
backtest ranking objective: \texttt{mase} uses median absolute error, while the
legacy \texttt{crps} suffix uses the nine-level pinball sum rather than exact
CRPS. The deployed policy is \texttt{mixtop3-mase}.

%% file: appendix/B2_frontier_controls.tex
\section{Compute ladder and additional controls}
\label{sec:app:frontier}

We provide the additional details supporting \cref{sec:results}: the thinking-mode ladder and compute accounting used for the compute frontier, together with the pool, policy, rival, compute-matched, and temporal-robustness controls from \cref{sec:controls}. The \bgaa\ procedure itself is defined in \cref{sec:method}.

\subsection{Thinking modes and the compute frontier}
\label{sec:app:ladder}

We use the four \emph{thinking modes} in \cref{tab:ladder} to trace the compute--performance frontier in \cref{fig:frontier}. Across modes, we vary only the test-time compute budget: the number of candidates scored by the gate, the number of gate windows and folds, and the number of views retained for aggregation. All modes draw from the same fixed candidate catalogue and use the same selection procedure and frozen model; lower-compute modes evaluate nested subsets of that catalogue.

\begin{table}[h]
\centering
\small
\caption{\textbf{Thinking modes define a compute ladder over a fixed candidate catalogue}
(\gift, $97$ configs). Across modes, we vary only the test-time compute budget:
the number of candidates scored by the gate, the number of gate windows and
folds, and the number of views retained for aggregation. Each mode evaluates a
nested subset of the same candidate catalogue, and the selection procedure is fixed. Cost is reported in forward passes relative
to the no-\ttc\ baseline. The left block gives the full \mchronos\ results; the
right block reports cost\,/\,$\Delta$MASE\% for the other three models.
\texttt{low} and \texttt{medium} use \texttt{mixtop2-mase}, while
\texttt{high} and \texttt{xhigh} use \texttt{mixtop3-mase}; these are compute-scaling variants for the frontier rather than changes to the deployed primary policy.}
\label{tab:ladder}
\footnotesize
\resizebox{\linewidth}{!}{
\begin{tabular}{lrrrrr@{\hskip 1.4em}rrr}
\toprule
& & \multicolumn{4}{c}{\mchronos} & \mtirex & \mrone & \mtoto \\
\cmidrule(lr){3-6}\cmidrule(lr){7-9}
mode & cand & cost & $\Delta$MASE\% & $\Delta$CRPS\% & worst\% & \multicolumn{3}{c}{cost / $\Delta$MASE\%} \\
\midrule
base            & $0$  & $1.00\times$ & ---     & ---     & ---     & $1.00\times$ & $1.00\times$ & $1.00\times$ \\
\texttt{low}    & $9$  & $2.40\times$ & $-0.44$ & $-0.52$ & $+5.62$ & $2.79\times$ / $+0.02$ & $2.40\times$ / $-0.45$ & $2.28\times$ / $-0.27$ \\
\texttt{medium} & $19$ & $3.03\times$ & $-0.49$ & $-0.69$ & $+5.67$ & $3.08\times$ / $-0.07$ & $3.01\times$ / $-0.20$ & $2.49\times$ / $-0.20$ \\
\texttt{high}   & $31$ & $5.07\times$ & $-0.59$ & $-0.57$ & $+7.92$ & $5.72\times$ / $-0.39$ & $4.90\times$ / $-0.47$ & $6.48\times$ / $-0.46$ \\
\texttt{xhigh}  & $51$ & $9.46\times$ & $\mathbf{-0.97}$ & $\mathbf{-1.07}$ & $+7.92$ & $9.54\times$ / $-0.45$ & $8.38\times$ / $-0.49$ & $8.32\times$ / $-0.68$ \\
\bottomrule
\end{tabular}
}
\end{table}

\begin{table}[h]
\centering\footnotesize
\caption{\textbf{Thinking modes provide a controlled test-time compute ladder.}
All four modes draw from the same fixed augmentation catalogue and use the same BGAA procedure; only the
test-time compute budget changes. Lower-compute modes evaluate nested subsets of the catalogue. \textbf{(a) Budget parameters:} $P$ denotes
the number of forward passes per backtest window, and
$W=\mathrm{clip}(\alpha N/(PF),W_{\min},W_{\max})$ is the median number of
gate windows. Gate and apply costs are reported relative to the base run, with
gate\% denoting the fraction of total cost spent on gating.
\textbf{(b) Candidate families:} entries show the number of degrees included
at each mode. The sets are nested. The resulting total cost and performance for
each mode are reported in \cref{tab:ladder}.}
\label{tab:modes}

\textbf{(a) Budget knobs}\\[2pt]
\begin{tabular}{lrrrrcrrrrrr}
\toprule
& \multicolumn{3}{c}{candidates} & \multicolumn{4}{c}{gate} & aggregate & \multicolumn{3}{c}{cost split, \mchronos} \\
\cmidrule(lr){2-4}\cmidrule(lr){5-8}\cmidrule(lr){9-9}\cmidrule(lr){10-12}
mode & families & views & $P$ & $\alpha$ & $W_{\min}/W_{\max}$ & $W$ & $F$ & top-$k$ & gate & apply & gate\% \\
\midrule
\texttt{low}    & $7$  & $9$  & $9$  & $1$      & $24/24$ & $24$  & $1$ & $2$ & $0.05$ & $2.07$ & $1.7$ \\
\texttt{medium} & $11$ & $19$ & $19$ & $6$      & $32/48$ & $48$  & $2$ & $2$ & $0.37$ & $2.15$ & $10.4$ \\
\texttt{high}   & $13$ & $31$ & $36$ & $16$     & $64/96$ & $64$  & $3$ & $3$ & $1.81$ & $3.66$ & $28.0$ \\
\texttt{xhigh}  & $15$ & $51$ & $58$ & $\infty$ & $16/128$& $128$ & $3$ & $3$ & $4.14$ & $2.87$ & $51.7$ \\
\bottomrule
\end{tabular}

\vspace{8pt}
\textbf{(b) Families admitted}\\[2pt]
\begin{tabular}{lcrrrr}
\toprule
family & passes/view & \texttt{low} & \texttt{medium} & \texttt{high} & \texttt{xhigh} \\
\midrule
\texttt{supportfloor}  & $0$      & $1$ & $1$ & $1$  & $1$  \\
\texttt{nonneg}        & $0$      & $1$ & $1$ & $1$  & $1$  \\
\texttt{rankgauss}     & $1$      & $1$ & $1$ & $1$  & $1$  \\
\texttt{winsor}        & $1$      & $1$ & $1$ & $1$  & $1$  \\
\texttt{multideseason} & $1$      & $1$ & $1$ & $1$  & $1$  \\
\texttt{recursive}     & $2$--$4$ & $1$ & $1$ & $2$  & $2$  \\
\texttt{lookback}      & $1$      & $3$ & $6$ & $10$ & $18$ \\
\midrule
\texttt{diff}          & $1$      & --- & $1$ & $1$  & $1$  \\
\texttt{jitter}        & $n_s$    & --- & $1$ & $1$  & $1$  \\
\texttt{jitternorm}    & $n_s$    & --- & $1$ & $1$  & $1$  \\
\texttt{offset}        & $1$      & --- & $4$ & $7$  & $14$ \\
\midrule
\texttt{sdiff}         & $1$      & --- & --- & $1$  & $1$  \\
\texttt{decomp}        & $1$      & --- & --- & $3$  & $3$  \\
\midrule
\texttt{fft}           & $1$      & --- & --- & ---  & $3$  \\
\texttt{seasontile}    & $1$      & --- & --- & ---  & $2$  \\
\midrule
jitter seeds $n_s$     &          & --- & $1$ & $2$  & $3$  \\
\bottomrule
\end{tabular}

{\scriptsize On yearly configs ($\text{period}=1$), \texttt{decomp} contributes
$1$ degree, so \texttt{high}/\texttt{xhigh} present $30/50$ views there.}
\end{table}

\paragraph{Compute accounting.}
\label{sec:app:flops}

\begin{figure}[t]
  \centering
  \includegraphics[width=\textwidth]{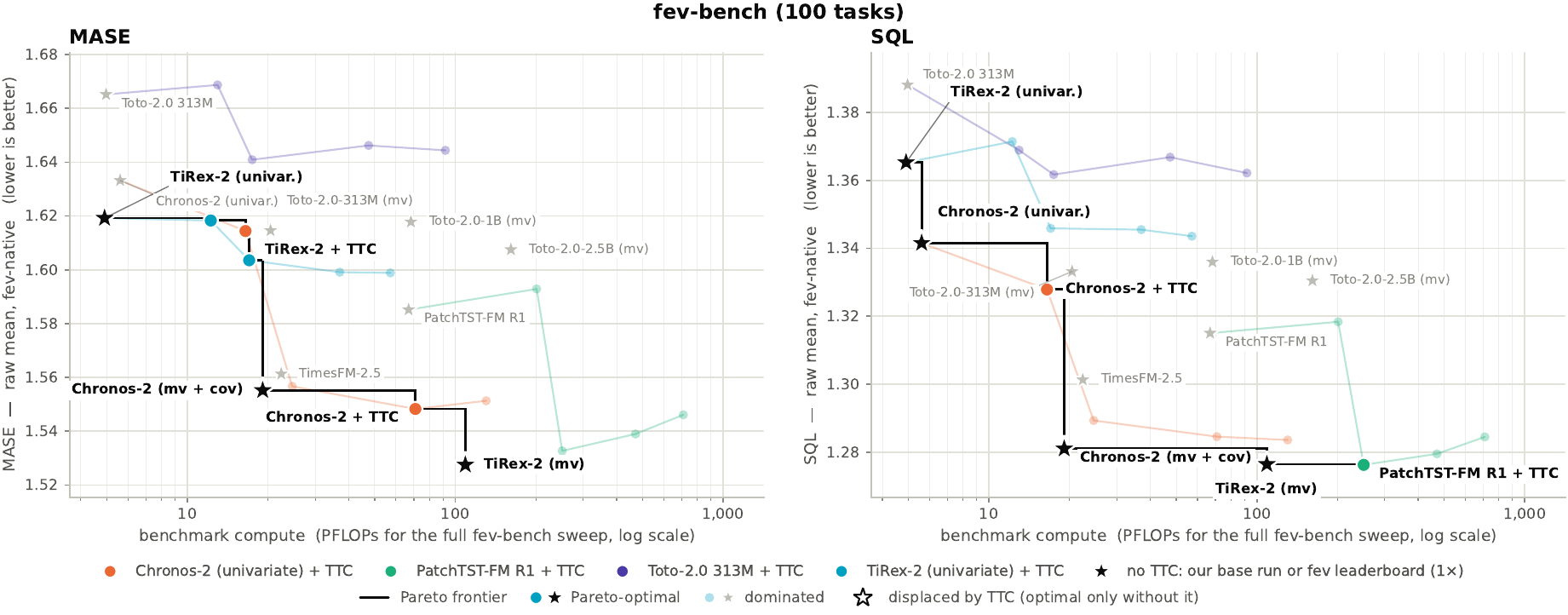}
  \caption{\textbf{Pareto curves for accuracy vs computational cost.} (\fev,
  $100$ tasks). Lines are our \ttc\ ladders over a frozen model, cheapest rung
  first; stars are published models on \fev. A rung sits at its model's
  measured $1\times$ benchmark PFLOPs times the measured forward-pass ratio
  $(\text{base}+\text{gate}+\text{apply})/\text{base}$. As in
  \cref{sec:app:flops}, charging each extra pass at the base-pass cost is a
  conservative accounting convention for models with variable-length computation, since some views shorten the context.}
  \label{fig:frontier-fev}
\end{figure}

\Cref{fig:frontier,fig:frontier-fev} and \cref{tab:ladder} place the \ttc\ ladders and published systems on a common compute axis. We use a conservative accounting scheme so that the comparison is consistent across models and compute budgets.

\paragraph{Compute axis.}
For each model, the $1\times$ point is its measured benchmark PFLOPs for a full \gift\ sweep, using the same profiling table for our models and the published baselines. We place each \ttc\ rung at $1\times\text{PFLOPs}\times c,$ where $c=(\text{base}+\text{gate}+\text{apply})/\text{base}$ is the forward-pass ratio recorded during evaluation. We charge every additional pass at the measured cost of a base pass. This can overestimate computation
for models that process shorter transformed contexts
more cheaply. However, \mtirex\ pads every input
to its full context window, so shortened contexts do
not reduce its per-call model computation. We retain
the same accounting convention across models and
compute budgets for consistency.

\paragraph{Interpretation of the compute estimate.}

The axis measures estimated FLOPs rather than wall-clock time and does not account for parallel batching. Charging shortened-context views at the full base-pass cost is conservative for models with variable-length computation, but not for fixed-length models such as \mtirex. The gate cost can also be amortized across multiple evaluation forecasts from the same configuration. We therefore distinguish this estimated-FLOPs accounting from the measured latency and memory reported in \cref{sec:app:latency}.

\paragraph{Paired and published baselines.}
The ladder values come from our \ttc\ runs over the $97$ \gift\ configs, while the published stars use the reported benchmark results from the compute table. We keep these sources separate. For example, our paired \mtoto\ base is $0.7032/0.4838$, compared with $0.7028/0.4814$ in the published table, while our \mrone\ base is $0.7052/0.4820$, compared with $0.7069/0.4829$. We use the base from each \ttc\ run to anchor its corresponding ladder, so the hollow base marker does not necessarily coincide exactly with the published star.

\paragraph{Chronos-2 configurations.}
Our \ttc\ results use Chronos-2 with cross-learning disabled, corresponding to the univariate $22.05$ PFLOPs configuration. Since that row does not have a published benchmark score, we use our base evaluation to anchor the ladder. We show the standard cross-learning-enabled Chronos-2 separately at $22.83$ PFLOPs. The two configurations have nearly identical compute, so this comparison separates gains from cross-series conditioning from gains obtained by spending the same budget on \ttc. On \gift, the cross-learning-enabled model performs better.

\paragraph{Compute scaling on \fev.}
\Cref{fig:frontier-fev} shows the same compute--accuracy analysis on \fev.
Across all four frozen models, increasing the \ttc\ budget improves over the
corresponding single-pass baseline on both MASE and SQL. The size of the gain
varies across models and metrics. The larger-budget \mchronos\ and
\mtirex\ settings give the strongest MASE improvements, while the other two
models still improve consistently over their own bases.

Thus, we use \fev\ as a second check that the compute--accuracy trade-off
extends beyond \gift, rather than as evidence that a single model defines the
Pareto frontier in every setting. As in \cref{fig:frontier}, each rung is placed
using the model's measured $1\times$ benchmark PFLOPs multiplied by its measured
forward-pass ratio, $(\text{base}+\text{gate}+\text{apply})/\text{base}$, following the conservative accounting described above.

\subsection{Measured Latency and Memory}
\label{sec:app:latency}

We complement our estimated-FLOPs analysis with wall-clock
latency and peak GPU-memory measurements on an 80GB H100
using one CPU core. We use synthetic contexts representative
of the median \gift\ configuration ($H=48$, context lengths
648--696), with throughput-oriented batching for all models.
Batch sizes are capped at 128, 512, and 256 for \mchronos,
\mtirex, and Toto-2-2.5B, respectively. Cold \bgaa\
includes historical backtesting, candidate forecasts,
view transformations, and CDF aggregation, while warm
inference reuses the previously selected views. GPU memory includes
model weights; component-based latency estimates agree
with end-to-end replays within 8\% for \mchronos\ and
\mtirex, and within 2\% for Toto-2-2.5B.

\begin{table}[h!]
\centering
\small
\setlength{\tabcolsep}{3pt}
\caption{\textbf{Accuracy, compute, latency, and memory across
thinking modes.}
MASE, WQL, and PFLOPs are from the full \gift\ sweep
(97 configurations); Toto-2-2.5B accuracy is published.
Latency and peak GPU memory are measured on an H100
using synthetic contexts representative of the median
\gift\ configuration. BGAA times include historical
selection, view transformations, and aggregation.
ms/fc is the per-forecast application cost after
selection at $F=10^4$; GPU is peak allocated memory
at $F=10^4$, including model weights.
\textbf{\mchronos+\bgaa\ is faster and more memory-efficient
than Toto-2-2.5B at high forecast volumes, although
the larger model retains better absolute accuracy.}}
\label{tab:app:latency-all}
\resizebox{\linewidth}{!}{%
\begin{tabular}{llrrrrrrrrr}
\toprule
 & & \multicolumn{3}{c}{\gift}
 & \multicolumn{4}{c}{Time for $F$ forecasts (s)}
 & & \\
\cmidrule(lr){3-5}
\cmidrule(lr){6-9}
System & Mode
& MASE & WQL & PFLOPs
& $F=1$ & $10^2$ & $10^3$ & $10^4$
& ms/fc & GPU (GiB) \\
\midrule
\texttt{toto\_2.0\_2.5b} & ---
& 0.6956 & $\mathbf{0.4759}$ & 121.0
& 0.133 & 0.319 & 2.86 & 28.5
& 2.85 & 10.30 \\
\midrule
\mtirex & base
& 0.6972 & 0.4791 & 15.6
& 0.0562 & 0.212 & 1.77 & 17.7
& 1.77 & 5.96 \\
\mtirex+\bgaa & \texttt{xhigh}
& $\mathbf{0.6940}$ & 0.4776 & 148.8
& 40.4 & 41.1 & 47.7 & 114
& 7.40 & 5.96 \\
\midrule
\mchronos & base
& 0.7043 & 0.4830 & 22.1
& 0.0143 & 0.0203 & 0.173 & 1.71
& 0.17 & 0.48 \\
\mchronos+\bgaa & \texttt{low}
& 0.7012 & 0.4805 & 53.0
& 0.125 & 0.160 & 0.780 & 7.02
& 0.69 & 0.48 \\
\mchronos+\bgaa & \texttt{medium}
& 0.7009 & 0.4797 & 66.8
& 0.280 & 0.358 & 1.11 & 8.59
& 0.83 & 0.48 \\
\mchronos+\bgaa & \texttt{high}
& 0.7001 & 0.4802 & 111.9
& 0.993 & 1.09 & 1.99 & 11.0
& 1.00 & 0.48 \\
\mchronos+\bgaa & \texttt{xhigh}
& 0.6975 & 0.4778 & 208.7
& 4.49 & 4.58 & 5.47 & 14.3
& 0.99 & 0.48 \\
\bottomrule
\end{tabular}%
}
\end{table}

An implementation detail affects \mtirex's computational
cost. Its \texttt{gifteval-zs} checkpoint pads every input
to a fixed length of $8192+928=9120$ time steps,
corresponding to 285 patches of length 32.
Consequently, even much shorter contexts and lookback
views incur approximately the same model computation
per forward call. Unlike models that benefit from
shorter transformed contexts, \mtirex\ therefore cannot
reduce its per-call cost through context cropping.
This also contributes to its relatively high historical
backtesting latency.

\begin{table}[h!]
\centering
\small
\setlength{\tabcolsep}{3pt}
\caption{\textbf{Detailed inference latency and memory across
forecast volumes.}
Measurements use one H100 GPU and one CPU core with
throughput-oriented batching. \mchronos\ and \mtirex\
use \bgaa\ in \texttt{xhigh} mode. Cold includes the
full historical backtest (22,272 forward passes);
warm reuses previously selected views.
Multipliers are relative to each model's frozen base;
Passes gives the corresponding forward-pass count ratio.
GPU denotes peak allocated memory in GiB, including weights.
\textbf{Batching substantially reduces verification overhead,
which can be amortized across repeated forecasts.}}
\label{tab:app:latency-buckets}

\resizebox{\linewidth}{!}{%
\begin{tabular}{llrrrrrrrr}
\toprule
& & \multicolumn{2}{c}{Time (s)}
& \multicolumn{4}{c}{Relative to frozen base ($\times$)}
& \multicolumn{2}{c}{GPU (GiB)} \\
\cmidrule(lr){3-4}
\cmidrule(lr){5-8}
\cmidrule(lr){9-10}
Model & $F$
& Base & BGAA cold
& Cold batched & Cold unbatched & Passes & Warm batched
& Base & BGAA \\
\midrule
\mchronos
& 1
& 0.0143 & 4.49 & 313 & 22,594 & 22,276 & 1.12
& 0.26 & 0.46 \\
& 10
& 0.0158 & 4.49 & 285 & 2,260 & 2,231 & 1.30
& 0.27 & 0.46 \\
& 100
& 0.0203 & 4.58 & 226 & 230 & 227 & 5.55
& 0.42 & 0.47 \\
& 500
& 0.0864 & 4.97 & 57.5 & 49.2 & 48.5 & 5.77
& 0.47 & 0.47 \\
& 1,000
& 0.173 & 5.47 & 31.7 & 26.6 & 26.3 & 5.78
& 0.47 & 0.48 \\
& 2,000
& 0.345 & 6.45 & 18.7 & 15.4 & 15.1 & 5.74
& 0.48 & 0.48 \\
& 10,000
& 1.71 & 14.3 & 8.38 & 6.32 & 6.23 & 5.77
& 0.48 & 0.48 \\
\midrule
\mtirex
& 1
& 0.0562 & 40.4 & 718 & 22,131 & 22,276 & 1.04
& 0.35 & 5.96 \\
& 10
& 0.0659 & 40.4 & 613 & 2,222 & 2,231 & 1.70
& 0.45 & 5.96 \\
& 100
& 0.212 & 41.1 & 194 & 226 & 227 & 3.53
& 1.44 & 5.96 \\
& 500
& 0.884 & 44.0 & 49.8 & 48.3 & 48.5 & 4.19
& 5.83 & 5.96 \\
& 1,000
& 1.77 & 47.7 & 27.0 & 26.1 & 26.3 & 4.19
& 5.96 & 5.96 \\
& 2,000
& 3.54 & 55.1 & 15.6 & 15.1 & 15.1 & 4.19
& 5.96 & 5.96 \\
& 10,000
& 17.7 & 114 & 6.46 & 6.20 & 6.23 & 4.18
& 5.96 & 5.96 \\
\bottomrule
\end{tabular}%
}

\vspace{7pt}

\begin{tabular}{rrr}
\toprule
\multicolumn{3}{c}{
\texttt{toto\_2.0\_2.5b} base} \\
\midrule
$F$ & Time (s) & GPU (GiB) \\
\midrule
1 & 0.133 & 9.47 \\
10 & 0.133 & 9.50 \\
100 & 0.319 & 9.79 \\
500 & 1.44 & 10.28 \\
1,000 & 2.86 & 10.29 \\
2,000 & 5.70 & 10.30 \\
10,000 & 28.5 & 10.30 \\
\bottomrule
\end{tabular}
\end{table}

Table~\ref{tab:app:latency-all} shows that the practical
cost of \bgaa\ depends strongly on forecast volume and
thinking budget. At $F=10^4$, \mchronos+\bgaa\
requires 7.02--14.3 seconds across its four modes,
compared with 28.5 seconds for Toto-2-2.5B, even
including historical selection. Its peak GPU memory
is approximately $21\times$ lower (0.48 versus
10.30 GiB). However, Toto-2-2.5B retains better
absolute \gift\ accuracy than \mchronos+\bgaa.
Conversely, \mtirex+\bgaa\ achieves lower MASE
(0.6940 versus 0.6956) but requires more wall-clock
time at the xhigh budget (114 versus 28.5 seconds).
Table~\ref{tab:app:latency-buckets} further shows
that batching reduces historical backtesting from
5.4 minutes to 4.5 seconds for \mchronos\ and
from 20.7 minutes to 40 seconds for \mtirex.
Cold-start overhead remains substantial for isolated
forecasts but can be amortized over repeated predictions.
These results distinguish estimated inference FLOPs
from practical latency and demonstrate that
verifier-guided inference can offer a favorable
accuracy--efficiency trade-off in high-throughput
forecasting workloads.

\subsection{Controls and ablations}
\label{sec:app:controls}

We use the following controls to test alternative explanations for the observed
gains. In \cref{tab:app:compute-matched}, we compare the gate against
random-$k$, noise-$k$, and identity-$k$ baselines at matched $k=2$ using paired
tests. \Cref{tab:app:controls} restricts the candidate pool to specific view
families, while \cref{tab:app:temporal} progressively ages the backtest windows
to test whether the gate mainly exploits the most recent regime.

\Cref{tab:policy-ablation} separates the effects of the candidate pool and the
aggregation rule. The pool ablation is nested: crops and offsets recover 56\% (\mchronos) and 17\%
(\mtoto) of the final MASE gain, adding the ideas-tier families improves further, and the decomposition family contributes an additional 0.37\% relative
MASE improvement for \mchronos\ and 0.11\% for \mtoto. The
aggregation ablation shows a trade-off between average improvement and
regression frequency. Among the fixed top-$k$ policies, the CDF mixture gives
the best mean MASE. The spread-routed variant gives up some of this gain but
improves the win/loss count ($63/29$) and reduces the worst regression relative
to the deployed mixture. \Cref{thm:ambiguity} motivates the label-free spread
statistic used by this policy, but its exact CRPS identity applies to the
mixture CDF before the nine-quantile read-back, not necessarily to the
reported CRPS after read-back. Finally, the temperature soft-gate improves
over the base and is competitive on CRPS, but underperforms the CDF mixture on MASE.

\begin{table}[h]
\centering
\small
\caption{\textbf{Pool and aggregation ablations separate the sources of improvement.}
\textbf{Left:} nested pool ablations for \mchronos\ (top) and \mtoto\ (bottom)
on \gift, using CDF mixture top-3 unless otherwise noted. \textbf{Right:}
aggregation ablations using the deployed pool on \gift. Expanding the candidate
pool generally improves MASE, while the aggregation rule determines the
trade-off between MASE and CRPS. For \mchronos, switching to the mixture is
approximately MASE-neutral. The decomposition family provides the final MASE
gain of the deployed configuration over the preceding pool.}
\label{tab:policy-ablation}
\begin{tabular}{lrr@{\hskip 2em}|lrr}
\toprule
pool tier & MASE & CRPS & policy & MASE & CRPS \\
\midrule
base (no TTC) & 0.7043 & 0.4830 & margin (conservative) & 0.7019 & 0.4823 \\
crop core & 0.7005 & 0.4811 & top-2 Vincent & 0.6990 & 0.4792 \\
+ micro views & 0.7007 & 0.4809 & top-3 Vincent & 0.6978 & 0.4780 \\
+ 9 ideas families & 0.7000 & 0.4809 & CDF mixture top-3 & 0.6975 & \textbf{0.4778} \\
+ mixture aggregation & 0.7001 & 0.4805 & spread-routed mixture & 0.6996 & 0.4796 \\
+ decomposition family & \textbf{0.6975} & \textbf{0.4778} & tempvinc-mase & 0.6991 & 0.4798 \\
\bottomrule
\end{tabular}
\begin{tabular}{lrr@{\hskip 2em}|lrr}
\toprule
pool tier & MASE & CRPS & policy & MASE & CRPS \\
\midrule
base (no TTC) & 0.7032 & 0.4838 & margin (conservative) & 0.7010 & 0.4822 \\
crop core & 0.7024 & 0.4827 & top-2 Vincent & 0.6998 & 0.4816 \\
+ micro views & 0.7021 & 0.4821 & top-3 Vincent & 0.6993 & 0.4810 \\
+ 9 ideas families & 0.7005 & 0.4820 & CDF mixture top-3 & 0.6984 & \textbf{0.4806} \\
+ mixture aggregation & 0.6992 & 0.4814 & spread-routed mixture & 0.6990 & \textbf{0.4806} \\
+ decomposition family & \textbf{0.6984} & \textbf{0.4806} & tempvinc-mase & 0.6997 & 0.4815 \\
\bottomrule
\end{tabular}
\end{table}

\begin{figure}[h]
  \centering
  \begin{subfigure}[t]{0.85\textwidth}
    \includegraphics[width=\textwidth]{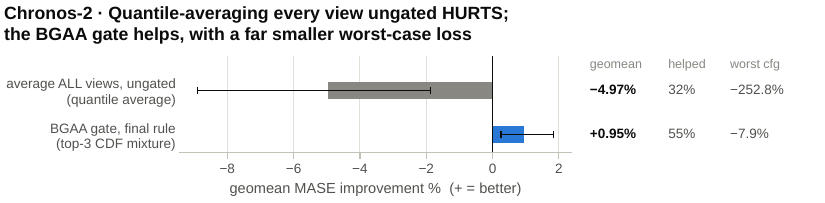}
    \caption{Ungated versus gated aggregation.}
    \label{fig:gate}
  \end{subfigure}
  \vspace{1em}
  \begin{subfigure}[t]{0.85\textwidth}
    \includegraphics[width=\textwidth]{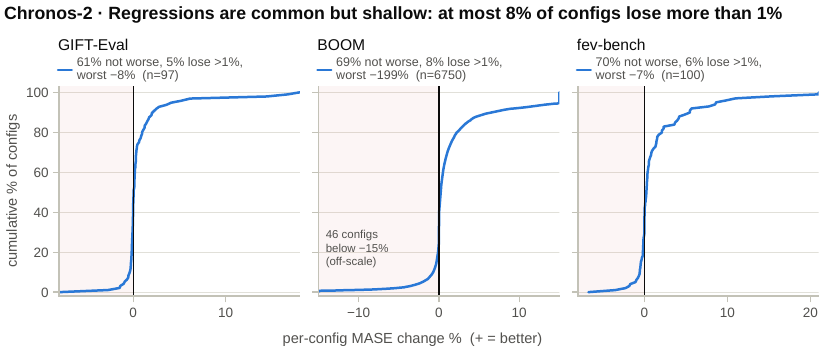}
    \caption{CDF of per-config MASE changes.}
    \label{fig:noharm}
  \end{subfigure}
  \vspace{1em}
  \begin{subfigure}[t]{0.85\textwidth}
    \includegraphics[width=\textwidth]{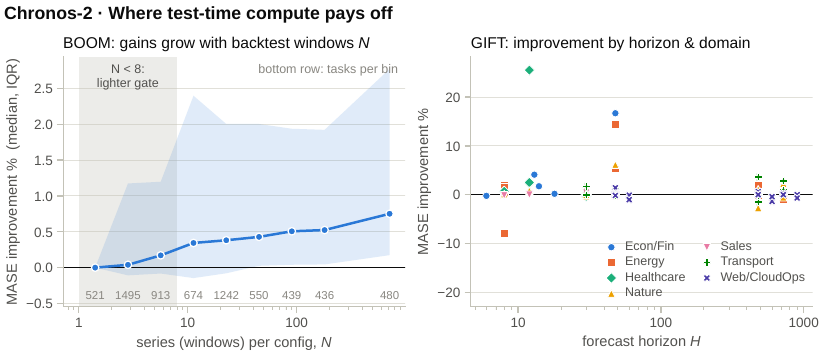}
    \caption{Gain as a function of backtest windows $N$ and horizon $H$.}
    \label{fig:where}
  \end{subfigure}
  \caption{\textbf{The gate drives the gains and benefits from sufficient
backtest evidence.}
\textbf{(a)} Ungated aggregation over the full candidate pool compared with
the gated policy. \textbf{(b)} CDFs of per-config MASE changes for the
policies in \cref{tab:noharm}. \textbf{(c)} MASE improvement as a function of
the number of available backtest windows $N$ and forecast horizon $H$. When
$N<8$, a lighter gate is applied, since this accounts for a third of the benchmark tasks.}
  \label{fig:capturability}
\end{figure}

\subsubsection{Rival controls}
\label{sec:app:exp6}
We restrict the candidate pool to lookback-only, offset-only, and cleaning-only
views, while keeping the same gated top-2 Vincent policy on \mchronos. We use
the same evaluation subset and report paired comparisons between each restricted
pool and the full candidate pool.

\begin{table}[h]
\centering\small
\caption{\textbf{Restricted-pool controls} (\mchronos, gated top-2 Vincent, subset
scope). ``share'' denotes the fraction of the full-pool gain recovered by each
restricted pool. The right block reports paired full-vs.-control comparisons on
the same configs. We use these paired tests for the main comparison,
since a restricted pool can recover part of the aggregate gain while still
underperforming the full pool on the same examples.}
\label{tab:app:controls}
\begin{tabular}{lrrrrrrr}
\toprule
 & \multicolumn{4}{c}{control alone} & \multicolumn{3}{c}{full pool vs control} \\
\cmidrule(lr){2-5}\cmidrule(lr){6-8}
pool & $\Delta$MASE\% & w/l & $p$ & share & $\Delta$\% & w/l & $p$ \\
\midrule
full pool & $-1.60$ & 9/5 & 0.011 & 100\% & --- & --- & --- \\
lookback-only & $-0.48$ & 8/2 & 0.037 & 30\% & $-1.13$ & 7/6 & 0.065 \\
offset-only & $-0.02$ & 3/4 & 0.100 & 1\% & $-1.58$ & 10/4 & 0.001 \\
cleaning-only & $+0.53$ & 3/5 & 0.484 & $-33$\% & $-2.12$ & 10/4 & 0.030 \\
\bottomrule
\end{tabular}
\end{table}

\subsubsection{Compute-matched controls}
\label{sec:app:exp5}
This control tests whether the gain comes simply from spending $k\times$ more
inference compute. We keep the candidate catalogue fixed to the deployed tier
(the identity plus $51$ non-identity candidates, $52$ total views at period $24$), use the same $k=2$ members per window, the same
equal-weight Vincent aggregation with the base, and the same \texttt{gluonts}
scoring. The only change is how the $k$ members are selected. For
\texttt{random\_k}, we average over three seeds to reduce variance from a
single random draw.

\begin{table}[h]
\centering
\small
\caption{\textbf{Compute-matched controls} (\mchronos, \gift, $n=97$, $k=2$).
The left block compares each control with the frozen base. The right block
reports the paired comparison between \bgaa\ and that control on the same
configs. We use the latter as the main test, since a control may improve
over the base while still underperforming the gated selection rule. All
significance tests use paired Wilcoxon tests across configs.}
\label{tab:app:compute-matched}
\begin{tabular}{lrrrrrrrrr}
\toprule
 & \multicolumn{3}{c}{vs base (MASE)} & \multicolumn{3}{c}{vs base (CRPS)} & \multicolumn{3}{c}{BGAA vs arm (MASE)} \\
\cmidrule(lr){2-4}\cmidrule(lr){5-7}\cmidrule(lr){8-10}
arm & $\Delta$\% & w/l & $p$ & $\Delta$\% & w/l & $p$ & $\Delta$\% & w/l & $p$ \\
\midrule
BGAA top-2 & $-0.76$ & 63/29 & $<0.001$ & $-0.79$ & 64/29 & $0.02$ & --- & --- & --- \\
random-$k$     & $+4.82$ & 35/60 & $4{\times}10^{-6}$ & $+10.73$ & 31/63 & $8{\times}10^{-8}$ & $-5.58$ & 77/20 & $1{\times}10^{-12}$ \\
noise-$k$ & $+0.03$ & 47/50 & 0.079 & $-0.05$ & 45/52 & 0.86 & $-0.80$ & 65/32 & $8\times10^{-5}$ \\
\bottomrule
\end{tabular}
\end{table}

First, \textbf{noise-$k$ gives no statistically significant gain}. Averaging $k$
tiny-jitter copies of the identity ($\sigma=0.002$) worsens MASE by $0.03\%$
($p=0.079$), while improving CRPS by $0.05\%$ ($p=0.86$).
This suggests that generic variance reduction does not explain
the overall effect. The gated policy still performs
substantially better head to head, by $0.80\%$ MASE.

Second, \textbf{random-$k$ performs worse than the base forecast}, by $4.82\%$
MASE and $10.73\%$ CRPS. This is expected for a broad candidate pool that
contains views which can be poorly aligned with a given series
(\cref{sec:app:exp9}): uniformly sampling from the pool often includes harmful
candidates. We therefore do not report a ``share of gain recovered'' for this
arm, since the corresponding ratio would be negative. The result also shows
that additional test-time compute is not useful by itself; without selection,
it can degrade performance.

Note: the reference arm in this experiment is gated top-$2$ Vincent rather
than the deployed \texttt{mixtop3-mase} policy. This gives a MASE improvement of
$-0.76\%$, compared with $-0.97\%$ for the deployed policy in
\cref{tab:main} on the same $97$ configs. We use the simpler top-$2$
Vincent policy for all arms so that the selection rule is the only factor that
changes across the comparison.

\subsubsection{Temporal robustness}
\label{sec:app:exp7}

We test how sensitive the gate is to the recency of its backtest evidence. For a
given lag, we restrict the backtest to observations at least
$\text{lag}\times H$ before the forecast start, while the final forecast still
uses the full context. If the gate mainly relies on the most recent regime, we
would expect its performance to deteriorate quickly as the lag increases.

\begin{table}[t]
\centering
\caption{Temporal robustness of the final selection policy (\textsc{Chronos-2}, CDF mixture top-3,
subset scope, $n=15$). We progressively restrict the backtest to older windows. Jaccard reports the
overlap between the admitted view set at each lag and the set selected at lag 0.}
\label{tab:app:temporal}
\small
\begin{tabular}{rrrrr}
\toprule
lag ($\times H$) & $\Delta$MASE\% & $\Delta$CRPS\% & $p$(MASE) & Jaccard vs lag 0 \\
\midrule
0 & $-1.60$ & $-1.54$ & 0.011 & 1.00 \\
1 & $-1.23$ & $-1.45$ & 0.043 & 0.59 \\
2 & $-1.71$ & $-1.07$ & 0.015 & 0.48 \\
4 & $-0.41$ & $-0.86$ & 0.050 & 0.44 \\
\bottomrule
\end{tabular}
\end{table}

The gain persists across lags, ranging from $-0.41\%$ to $-1.71\%$, and is
significant at lags 0, 1, and 2, with borderline significance at lag 4
($p=0.050$), even though the admitted set changes substantially.
This suggests that the gate is exploiting a signal that is not limited to the
most recent regime. We do not interpret the non-monotonic pattern across lags
as evidence that older backtest windows are better. CRPS improvements decline
with lag, while MASE shows a non-monotonic pattern and remains the metric used
to rank the policy.

\subsubsection{Forecast Combination Under Matched Historical Evidence}
\label{sec:app:matched-combiners}

\begin{table}[h!]
\centering
\small
\caption{\textbf{BGAA versus alternative forecast combiners under matched
historical evidence.} All methods use identical candidate forecasts and
backtest windows on 30 configurations (15 \gift, 15 fev-bench) across
three TSFMs. \textbf{Top:} MASE and WQL changes versus the frozen base
(\%; lower is better; best per row in bold). \textbf{Bottom:} Aggregate
MASE gap to \bgaa\ (95\% bootstrap CI), worst regression, configurations
degrading by more than 3\%, and mean additional forecasts used.
\textbf{BGAA achieves the best aggregate accuracy and robustness while
using fewer forecasts than the regularized alternatives.}}
\label{tab:bgaa-combiners}
\begin{tabular}{lrrrrrrrrrr}
\toprule
 & \multicolumn{2}{c}{\bgaa}
 & \multicolumn{2}{c}{Top-1}
 & \multicolumn{2}{c}{Base + top-1}
 & \multicolumn{2}{c}{L2-convex}
 & \multicolumn{2}{c}{L2-anchored} \\
\cmidrule(lr){2-3}
\cmidrule(lr){4-5}
\cmidrule(lr){6-7}
\cmidrule(lr){8-9}
\cmidrule(lr){10-11}
 & MASE & WQL & MASE & WQL
 & MASE & WQL & MASE & WQL & MASE & WQL \\
\midrule
\mchronos, \gift
 & $\mathbf{-4.34}$ & $\mathbf{-3.99}$
 & $-0.96$ & $+0.46$
 & $-2.10$ & $-1.56$
 & $-3.27$ & $-2.66$
 & $-2.96$ & $-2.75$ \\
\mchronos, fev
 & $\mathbf{-7.83}$ & $\mathbf{-5.85}$
 & $-4.34$ & $-0.71$
 & $-6.71$ & $-4.09$
 & $-3.65$ & $-1.87$
 & $-4.05$ & $-1.86$ \\
Toto-2, \gift
 & $-1.93$ & $-1.77$
 & $+0.22$ & $+1.53$
 & $-1.14$ & $-0.53$
 & $-2.39$ & $\mathbf{-2.29}$
 & $\mathbf{-2.59}$ & $-2.03$ \\
Toto-2, fev
 & $-2.43$ & $\mathbf{-2.20}$
 & $-1.65$ & $-0.92$
 & $-2.26$ & $-1.95$
 & $\mathbf{-3.95}$ & $-1.63$
 & $-3.24$ & $-1.19$ \\
\mtirex, \gift
 & $-1.69$ & $-1.74$
 & $-1.42$ & $-0.40$
 & $-0.91$ & $-0.68$
 & $-2.49$ & $\mathbf{-2.91}$
 & $\mathbf{-2.96}$ & $-2.86$ \\
\mtirex, fev
 & $\mathbf{-4.69}$ & $\mathbf{-5.64}$
 & $-2.55$ & $-2.86$
 & $-2.82$ & $-2.97$
 & $-2.05$ & $-3.36$
 & $-2.25$ & $-3.30$ \\
\midrule
\textbf{All (90)}
 & $\mathbf{-3.84}$ & $\mathbf{-3.55}$
 & $-1.79$ & $-0.49$
 & $-2.68$ & $-1.97$
 & $-2.97$ & $-2.45$
 & $-3.01$ & $-2.33$ \\
\midrule
Gap to \bgaa
 & \multicolumn{2}{c}{---}
 & \multicolumn{2}{c}{$+2.11$}
 & \multicolumn{2}{c}{$+1.20$}
 & \multicolumn{2}{c}{$+0.91$}
 & \multicolumn{2}{c}{$+0.86$} \\
\quad 95\% CI
 & \multicolumn{2}{c}{}
 & \multicolumn{2}{c}{[$0.45$, $4.04$]}
 & \multicolumn{2}{c}{[$0.22$, $2.35$]}
 & \multicolumn{2}{c}{[$-0.48$, $2.61$]}
 & \multicolumn{2}{c}{[$-0.57$, $2.59$]} \\
Worst config
 & \multicolumn{2}{c}{$\mathbf{+1.38}$}
 & \multicolumn{2}{c}{$+27.39$}
 & \multicolumn{2}{c}{$+6.54$}
 & \multicolumn{2}{c}{$+12.30$}
 & \multicolumn{2}{c}{$+12.30$} \\
Configs $>3\%$ worse
 & \multicolumn{2}{c}{$\mathbf{0}$}
 & \multicolumn{2}{c}{$6$}
 & \multicolumn{2}{c}{$3$}
 & \multicolumn{2}{c}{$8$}
 & \multicolumn{2}{c}{$5$} \\
Candidates used
 & \multicolumn{2}{c}{$2.8$}
 & \multicolumn{2}{c}{$1.0$}
 & \multicolumn{2}{c}{$1.0$}
 & \multicolumn{2}{c}{$13.2$}
 & \multicolumn{2}{c}{$6.3$} \\
\bottomrule
\end{tabular}
\end{table}

To isolate the contribution of historical selection and aggregation,
we compare \bgaa\ against four alternative combiners using identical
candidate forecasts and backtest windows. The 30 configurations
(15 \gift, 15 fev-bench) were selected beforehand for high augmentation
potential, ranked by the deployed gate's MASE gain averaged over the three models with at most two per dataset family. This is because view effectiveness can only be compared across series characteristics where views actually change the forecast; this oversamples datasets where views matter, so effect sizes are larger than on a typical configuration.
Thus, this is a targeted subset rather than a representative benchmark
sample. Top-1 selects the historically best view using \bgaa's ranking
statistic and identity fallback; base + top-1 averages it with the
identity forecast. L2-convex fits per-quantile nonnegative, sum-to-one
weights by minimizing backtest pinball loss with an
$\ell_2$ penalty, while L2-anchored penalizes deviation from the
identity weight vector. Both select regularization by five-fold
cross-validation on historical evidence.

Table~\ref{tab:bgaa-combiners} shows that \bgaa\ achieves the largest
aggregate MASE ($-3.84\%$) and WQL ($-3.55\%$) improvements.
Its advantages over top-1 and base + top-1 are statistically
significant, whereas the differences from L2-regularized combinations
are not. However, \bgaa\ has a substantially smaller worst regression
($1.38\%$ versus $6.54$--$27.39\%$), no configurations degrading by
more than $3\%$, and uses only $2.8$ additional forecasts on average,
versus $6.3$--$13.2$ for the regularized methods. This suggests that
historical admission followed by simple aggregation offers a favorable
accuracy--robustness--compute trade-off without fitting combination
weights to historical outcomes.

\subsubsection{Transferability of Restricted Candidate Pools}
\label{sec:app:pool-transfer}

To examine whether the candidate pool can be pruned without
sacrificing transferability, we construct restricted pools using
only \mchronos/\gift\ backtests, ranking views by their selection
frequency (ties broken by positive backtest gains and reciprocal
rank). We then apply these pools unchanged to the other models
and benchmarks on the same 90 model--configuration pairs as
Table~\ref{tab:bgaa-combiners}. Only the candidate set changes;
the backtest windows, scoring, top-3 aggregation, and evaluation
forecasts are held fixed. We additionally compare pools selected
by leave-one-configuration-out ranking within the same
model--benchmark pair or across other pairs, and 200 randomly
sampled pools. The source pair is in-sample by construction,
since the 22-view pool contains every view selected there.

Table~\ref{tab:pool-restriction} shows that the 22-view pool
retains only $69\%$ of the full-pool MASE gain on the 75
transfer configurations, with a significant disadvantage of
$+1.18$ log points (95\% CI $[0.44,2.20]$). Even excluding
the three largest gaps, the disadvantage remains $+0.47$
$[0.23,0.76]$. Pools selected through leave-one-configuration-out
procedures retain approximately $79$--$80\%$ of the gain
at roughly $53$--$58\%$ of the forward-pass cost, but remain
significantly worse than the full pool on MASE.
Table~\ref{tab:pool-views} explains this transfer loss:
$30$--$41\%$ of selected views in the other model--benchmark
pairs are absent from the source-model pool. Thus, smaller
pools offer meaningful computational savings, but a diverse
shared pool avoids excluding transformations that become
useful for other models or datasets, without requiring
deployment-specific pool tuning.

\begin{table}[t]
\centering
\small
\setlength{\tabcolsep}{2.7pt}
\caption{\textbf{Accuracy--compute trade-off and transferability of
restricted candidate pools.}
Pools selected using Chronos-2/\gift\ backtests are applied
unchanged to other models and benchmarks.
\textbf{(a)} MASE and WQL changes versus the frozen base
(\%; lower is better), with retained MASE gain and paired
gap to the full pool (95\% bootstrap CI).
\textbf{(b)} Alternative pool-selection strategies across
90 model--configuration pairs, including regression risk
and relative forward-pass cost.
\textbf{Smaller pools reduce compute but significantly
lose accuracy relative to the full pool, particularly
when transferred across models and benchmarks.}}
\label{tab:pool-restriction}

\resizebox{\linewidth}{!}{%
\begin{tabular}{lrrrrrrrrr}
\toprule
\multicolumn{10}{l}{
\textbf{(a) Chronos-2/\gift\ pool transferred across models and benchmarks}} \\
\midrule
 & \multicolumn{2}{c}{Full pool (51)}
 & \multicolumn{4}{c}{Restricted pool (22)}
 & \multicolumn{3}{c}{Retained gain, top-$k$ (\%)} \\
\cmidrule(lr){2-3}
\cmidrule(lr){4-7}
\cmidrule(lr){8-10}
Model / benchmark
& MASE & WQL
& MASE & WQL & Kept (\%)
& Gap [95\% CI]
& $k=10$ & $k=5$ & $k=3$ \\
\midrule
\mchronos, \gift$^\ast$
& $-4.34$ & $-3.99$
& $-4.34$ & $-3.99$
& $100$ & $0.00$
& $57$ & $42$ & $43$ \\
\midrule
\mchronos, fev
& $-7.83$ & $-5.85$
& $-4.36$ & $-4.25$
& $56$ & $+3.70$ {\scriptsize[0.47, 8.42]}
& $32$ & $25$ & $33$ \\

Toto-2, \gift
& $-1.93$ & $-1.77$
& $-1.24$ & $-0.92$
& $64$ & $+0.70$ {\scriptsize[$-$0.13, 2.15]}
& $47$ & $16$ & $5$ \\

Toto-2, fev
& $-2.43$ & $-2.20$
& $-2.15$ & $-1.75$
& $88$ & $+0.29$ {\scriptsize[$-$0.04, 0.68]}
& $74$ & $80$ & $34$ \\

\mtirex, \gift
& $-1.69$ & $-1.74$
& $-1.22$ & $-1.38$
& $72$ & $+0.48$ {\scriptsize[0.03, 1.12]}
& $57$ & $65$ & $57$ \\

\mtirex, fev
& $-4.69$ & $-5.64$
& $-3.97$ & $-4.69$
& $85$ & $+0.75$ {\scriptsize[0.25, 1.33]}
& $39$ & $55$ & $50$ \\
\midrule
\textbf{Transfer (75)}
& $\mathbf{-3.74}$ & $\mathbf{-3.46}$
& $-2.60$ & $-2.61$
& $69$ & $\mathbf{+1.18}$ {\scriptsize[0.44, 2.20]}
& $43$ & $42$ & $37$ \\

\textbf{All (90)}
& $\mathbf{-3.84}$ & $\mathbf{-3.55}$
& $-2.89$ & $-2.84$
& $75$ & $\mathbf{+0.99}$ {\scriptsize[0.36, 1.84]}
& $45$ & $42$ & $38$ \\
\bottomrule
\end{tabular}%
}

\vspace{7pt}

\resizebox{\linewidth}{!}{%
\begin{tabular}{lrrrrrrrr}
\toprule
\multicolumn{9}{l}{
\textbf{(b) Alternative restricted-pool selection strategies}} \\
\midrule
Pool selection
& $k$
& MASE
& Kept (\%)
& Gap MASE [95\% CI]
& Gap WQL [95\% CI]
& Worst
& $>3\%$
& Cost \\
\midrule
\textbf{Full pool}
& $51$ & $\mathbf{-3.84}$ & $\mathbf{100}$
& --- & ---
& $\mathbf{+1.38}$ & $\mathbf{0}$ & $1.00$ \\
\addlinespace[2pt]

\mchronos/\gift
& $22$ & $-2.89$ & $75$
& $+0.99$ {\scriptsize[0.36, 1.84]}
& $+0.73$ {\scriptsize[0.30, 1.23]}
& $+2.80$ & $0$ & $0.60$ \\

& $20$ & $-2.83$ & $74$
& $+1.05$ {\scriptsize[0.32, 2.09]}
& $+0.79$ {\scriptsize[0.33, 1.32]}
& $+2.80$ & $0$ & $0.57$ \\

& $10$ & $-1.75$ & $45$
& $+2.15$ {\scriptsize[0.92, 3.73]}
& $+2.00$ {\scriptsize[0.93, 3.28]}
& $+6.17$ & $2$ & $0.43$ \\
\addlinespace[2pt]

Same-pair LOO
& $20$ & $-3.09$ & $80$
& $+0.78$ {\scriptsize[0.25, 1.45]}
& $+0.58$ {\scriptsize[$-$0.02, 1.23]}
& $+4.10$ & $2$ & $0.53$ \\

& $10$ & $-2.11$ & $55$
& $+1.78$ {\scriptsize[0.75, 3.00]}
& $+1.17$ {\scriptsize[0.09, 2.39]}
& $+25.10$ & $3$ & $0.34$ \\
\addlinespace[2pt]

Other-pairs LOO
& $20$ & $-3.03$ & $79$
& $+0.84$ {\scriptsize[0.28, 1.51]}
& $+0.73$ {\scriptsize[0.05, 1.59]}
& $+5.93$ & $3$ & $0.58$ \\

& $10$ & $-1.83$ & $48$
& $+2.07$ {\scriptsize[0.72, 3.81]}
& $+2.21$ {\scriptsize[0.94, 3.72]}
& $+6.11$ & $8$ & $0.39$ \\
\addlinespace[2pt]

Random$^\ddagger$
& $20$ & $-2.40$ & $62$
& $+1.49$ {\scriptsize[0.67, 2.35]}
& $+1.29$ {\scriptsize[0.75, 2.18]}
& $+4.39$ & $1.2$ & $0.47$ \\

& $10$ & $-1.61$ & $42$
& $+2.30$ {\scriptsize[1.43, 3.11]}
& $+2.07$ {\scriptsize[1.21, 2.92]}
& $+4.50$ & $1.5$ & $0.27$ \\
\bottomrule
\end{tabular}%
}
\end{table}

\begin{table}[t]
\centering
\small
\setlength{\tabcolsep}{3pt}
\caption{\textbf{Model- and benchmark-dependent view selection.}
Views selected by full-pool \bgaa\ in at least 3 of
15 configurations, with selection frequencies in parentheses.
Distinct counts views selected at least once; Outside is
the fraction of selections absent from the 22-view
Chronos-2/\gift\ pool. $^\dagger$ denotes an excluded view
and $m$ the seasonal period.
\textbf{Across transfer settings, 30--41\% of selected views
lie outside the source-model pool.}}
\label{tab:pool-views}

\begin{tabular}{llrr}
\toprule
Model / benchmark
& Selected views (frequency)
& Distinct
& Outside \\
\midrule

\mchronos, \gift$^\ast$
& \parbox[t]{0.54\linewidth}{\raggedright
\texttt{decomp}$_m$ (7),
\texttt{decomp}$_{2m}$ (6),
\texttt{rankgauss} (5),
\texttt{multideseason}$_m$ (4)}
& 22 & $0\%$ \\

\mchronos, fev
& \parbox[t]{0.54\linewidth}{\raggedright
\texttt{decomp}$_{2m}$ (5),
\texttt{decomp}$_m$ (5),
\texttt{diff1}$^\dagger$ (4),
\texttt{lookback}$_{0.125}^\dagger$ (3),
\texttt{rankgauss} (3),
\texttt{multideseason}$_m$ (3)}
& 24 & $30\%$ \\

Toto-2, \gift
& \parbox[t]{0.54\linewidth}{\raggedright
\texttt{recursive}$_{0.25}$ (5),
\texttt{decomp}$_m$ (3)}
& 27 & $34\%$ \\

Toto-2, fev
& \parbox[t]{0.54\linewidth}{\raggedright
\texttt{winsor}$_{0.005}$ (4),
\texttt{recursive}$_{0.5}^\dagger$ (4),
\texttt{multideseason}$_m$ (3),
\texttt{diff1}$^\dagger$ (3),
\texttt{rankgauss} (3),
\texttt{decomp}$_m$ (3)}
& 22 & $41\%$ \\

\mtirex, \gift
& \parbox[t]{0.54\linewidth}{\raggedright
\texttt{supportfloor}$^\dagger$ (3),
\texttt{nonneg}$^\dagger$ (3),
\texttt{decomp}$_m$ (3)}
& 27 & $40\%$ \\

\mtirex, fev
& \parbox[t]{0.54\linewidth}{\raggedright
\texttt{rankgauss} (5),
\texttt{recursive}$_{0.5}^\dagger$ (4),
\texttt{decomp}$_m$ (4),
\texttt{winsor}$_{0.005}$ (3),
\texttt{decomp}$_{m/2}$ (3),
\texttt{multideseason}$_m$ (3)}
& 25 & $36\%$ \\

\bottomrule
\end{tabular}
\end{table}

\clearpage

%% file: appendix/C_full_tables.tex
\section{Full results and verification diagnostics}
\label{sec:app:tables}

This section collects the full results underlying \cref{sec:results} and the
verification analyses from \cref{sec:limits}. We include the cross-model policy
grid for each benchmark, the complete official \boom\ leaderboard corresponding
to the excerpt in \cref{tab:vs-upgrade}, the per-domain breakdown, the
cross-fitted headroom and ranking results, and a small set of qualitative
forecast examples illustrating how \bgaa\ changes the prediction on selected
successful windows. Compute-ladder, gate-ablation, and matched-control analyses
are kept with their underlying experiments in
\cref{sec:app:ladder,sec:app:controls} rather than duplicated here.

\subsection{Benchmark-level results}

\begin{table}[h]
\centering\small
\caption{\textbf{\gift\ results across four models and four policies}
($97$ configs; geomean of metric/seasonal-naive). We report
$\Delta\%$ and MASE win/loss counts relative to each model's own base; all
MASE comparisons have $p<0.023$. \texttt{mixtop3-mase} is the deployed policy.
\texttt{routemix05-crps} gives the best win/loss ratio for \mrone\ and \mtirex, but with a smaller mean gain. \texttt{top3-mase}
has the best win/loss ratio for \mchronos\, while \texttt{routemix05-crps} has the best win/loss ratio for \mtoto. On MASE, \mrone\ and \mtirex are the only
models for which \texttt{top2-mase} strictly outperforms the deployed policy.}
\label{tab:app:gift-policies}
\begin{tabular}{llrrrrr}
\toprule
model & policy & MASE & CRPS & $\Delta$MASE\% & $\Delta$CRPS\% & w/l \\
\midrule
\multirow{5}{*}{\mchronos}
 & base                     & 0.7043 & 0.4830 & ---     & ---     & --- \\
 & \texttt{top2-mase}       & 0.6990 & 0.4792 & $-0.76$ & $-0.79$ & 63/29 \\
 & \texttt{top3-mase}       & 0.6978 & 0.4780 & $-0.93$ & $-1.04$ & 65/27 \\
 & \texttt{routemix05-crps} & 0.6996 & 0.4796 & $-0.67$ & $-0.70$ & 63/29 \\
 & \textbf{\texttt{mixtop3-mase}} & \textbf{0.6975} & \textbf{0.4778} & $\mathbf{-0.97}$ & $-1.07$ & 55/38 \\
\midrule
\multirow{5}{*}{\mtirex}
 & base                     & 0.6972 & 0.4791 & ---     & ---     & --- \\
 & \texttt{top2-mase}       & 0.6937 & 0.4776 & $-0.50$ & $-0.33$ & 57/33 \\
 & \texttt{top3-mase}       & 0.6939 & 0.4772 & $-0.46$ & $-0.39$ & 57/33 \\
 & \texttt{routemix05-crps} & 0.6949 & 0.4775 & $-0.33$ & $-0.33$ & 62/28 \\
 & \textbf{\texttt{mixtop3-mase}} & \textbf{0.6940} & \textbf{0.4776} & $-0.45$ & $-0.32$ & 54/38 \\
\midrule
\multirow{5}{*}{\mrone}
 & base                     & 0.7052 & 0.4820 & ---     & ---     & --- \\
 & \texttt{top2-mase}       & 0.7009 & 0.4810 & $-0.61$ & $-0.21$ & 62/30 \\
 & \texttt{top3-mase}       & 0.7020 & 0.4818 & $-0.45$ & $-0.05$ & 62/30 \\
 & \texttt{routemix05-crps} & 0.7036 & 0.4810 & $-0.21$ & $-0.21$ & 67/24 \\
 & \textbf{\texttt{mixtop3-mase}} & \textbf{0.7018} & \textbf{0.4811} & $-0.49$ & $-0.19$ & 64/28 \\
\midrule
\multirow{5}{*}{\mtoto}
 & base                     & 0.7032 & 0.4838 & ---     & ---     & --- \\
 & \texttt{top2-mase}       & 0.6998 & 0.4816 & $-0.49$ & $-0.46$ & 60/31 \\
 & \texttt{top3-mase}       & 0.6993 & 0.4810 & $-0.55$ & $-0.58$ & 60/31 \\
 & \texttt{routemix05-crps} & 0.6990 & 0.4806 & $-0.60$ & $-0.66$ & 58/29 \\
 & \textbf{\texttt{mixtop3-mase}} & \textbf{0.6984} & \textbf{0.4806} & $\mathbf{-0.69}$ & $-0.66$ & 60/32 \\
\bottomrule
\end{tabular}
\end{table}

\begin{table}[h]
\centering\small
\caption{\textbf{\fev\ results across all four models} (all $100$ tasks;
\texttt{fev}-native raw metric means and skill relative to the published
seasonal-naive baseline, where higher is better). SQL is the primary \texttt{fev}
metric and is used for the paired win/loss counts; every \ttc\ row has
is statistically significant against its corresponding base. \texttt{routemix05-crps} again trades
some mean improvement for a substantially better win/loss ratio, e.g.,
$64/27$ versus $54/39$ for \mtirex.}
\label{tab:app:fev}
\resizebox{\linewidth}{!}{
\begin{tabular}{llrrrrrr}
\toprule
model & policy & SQL & WQL & MASE & SQL skill & MASE skill & SQL w/l \\
\midrule
\multirow{3}{*}{\mchronos}
 & base                     & 1.3416 & 0.2193 & 1.6332 & 0.438 & 0.312 & --- \\
 & \texttt{routemix05-crps} & 1.2882 & 0.2179 & 1.5593 & 0.447 & 0.324 & 71/23 \\
 & \texttt{mixtop3-mase}    & 1.2836 & 0.2163 & 1.5513 & 0.447 & 0.325 & 56/38 \\
\midrule
\multirow{3}{*}{\mtirex}
 & base                     & 1.3653 & 0.2275 & 1.6192 & 0.426 & 0.305 & --- \\
 & \texttt{routemix05-crps} & 1.3460 & 0.2266 & 1.6098 & 0.431 & 0.309 & 64/27 \\
 & \texttt{mixtop3-mase}    & 1.3436 & 0.2263 & 1.5989 & 0.430 & 0.309 & 54/39 \\
\midrule
\multirow{3}{*}{\mrone}
 & base                     & 1.3150 & 0.1956 & 1.5852 & 0.446 & 0.328 & --- \\
 & \texttt{routemix05-crps} & 1.2778 & 0.1954 & 1.5386 & 0.455 & 0.338 & 70/22 \\
 & \texttt{mixtop3-mase}    & 1.2845 & 0.1954 & 1.5461 & 0.455 & 0.338 & 65/27 \\
\midrule
\multirow{3}{*}{\mtoto}
 & base                     & 1.3881 & 0.2151 & 1.6652 & 0.433 & 0.316 & --- \\
 & \texttt{routemix05-crps} & 1.3687 & 0.2156 & 1.6467 & 0.437 & 0.318 & 60/33 \\
 & \texttt{mixtop3-mase}    & 1.3622 & 0.2133 & 1.6444 & 0.439 & 0.319 & 56/37 \\
\bottomrule
\end{tabular}}
\end{table}

\begin{table}[h]
\centering\small
\caption{\textbf{Full official \boom\ leaderboard} ($6750$ scaled tasks from
the $7413$-task benchmark; DataDog seasonal-naive baseline; shifted geometric
mean with $\varepsilon=10^{-5}$). We recompute all $23$ published systems using
DataDog's evaluation pipeline and verify the reproduced scores against the
published leaderboard. Bold rows are our runs. \Cref{tab:vs-upgrade} shows the
corresponding excerpt used in the main text.}
\label{tab:app:boom-full}
\begin{tabular}{lrr@{\hskip 2.2em}lrr}
\toprule
system & MASE & CRPS & system & MASE & CRPS \\
\midrule
\textbf{\mtoto{}+\bgaa}       & \textbf{0.575} & \textbf{0.350} & \texttt{moirai\_2\_small}    & 0.672 & 0.427 \\
\texttt{toto\_2.0\_2.5b}      & 0.581 & 0.349 & \texttt{moirai\_1.1\_base}   & 0.710 & 0.428 \\
\texttt{toto\_2.0\_1b}        & 0.582 & 0.349 & \texttt{moirai\_1.1\_large}  & 0.720 & 0.436 \\
\texttt{toto\_2.0\_313m}      & 0.585 & 0.351 & \texttt{moirai\_1.1\_small}  & 0.729 & 0.442 \\
\textbf{\mtoto\ base}         & \textbf{0.585} & \textbf{0.353} & \texttt{timesfm\_2\_0\_500m} & 0.725 & 0.447 \\
\texttt{toto\_2.0\_22m}       & 0.601 & 0.363 & \texttt{chronos\_bolt\_base} & 0.726 & 0.451 \\
\texttt{toto} (v1)            & 0.617 & 0.375 & \texttt{chronos\_bolt\_small}& 0.733 & 0.455 \\
\textbf{\mtirex{}+\bgaa}      & \textbf{0.619} & \textbf{0.376} & \texttt{timer}               & 0.796 & 0.639 \\
\texttt{toto\_2.0\_4m}        & 0.624 & 0.377 & \texttt{time\_moe\_200M}     & 0.881 & 0.643 \\
\textbf{\mchronos{}+\bgaa}    & \textbf{0.626} & \textbf{0.380} & \texttt{time\_moe\_50M}      & 0.806 & 0.649 \\
\texttt{timesfm\_2\_5\_200m}  & 0.632 & 0.382 & \texttt{visionts}            & 0.988 & 0.673 \\
\textbf{\mtirex\ base}        & \textbf{0.642} & \textbf{0.386} & \texttt{autoarima}           & 0.824 & 0.736 \\
\textbf{\mrone{}+\bgaa}       & \textbf{0.641} & \textbf{0.388} & \texttt{seasonalnaive}       & 1.000 & 1.000 \\
\texttt{chronos\_2}           & 0.641 & 0.392  & \texttt{autotheta}           & 1.123 & 1.018 \\
\textbf{\mchronos\ base}      & \textbf{0.655} & \textbf{0.394} & \texttt{autoets}             & 0.842 & 1.975 \\
\textbf{\mrone\ base}         & \textbf{0.681} & \textbf{0.407} & & & \\
\bottomrule
\end{tabular}
\\[2pt]
{\footnotesize Ordered by CRPS, as the leaderboard is. \bgaa\ moves every one of
our four models up the table. \mtoto\ with \ttc\ ($0.575$) is the best
MASE of any \emph{zero-shot} row here.}
\end{table}

\begin{table}[h]
\centering
\small
\caption{\gift\ results for \mchronos\ across the deployed policy grid.
Changes are relative to the base score of $0.7043$ MASE / $0.4830$ CRPS;
``worst'' denotes the largest regression on any single config.
\textbf{Bold} marks the final deployed policy.}
\label{tab:app:policiesc2}
\resizebox{\linewidth}{!}{
\begin{tabular}{lrrrrrrrrr}
\toprule
policy & MASE & $\Delta$\% & $p$ & w/l & worst\% & CRPS & $\Delta$\% & $p$ & w/l \\
\midrule
base & 0.7043 & --- & --- & --- & --- & 0.4830 & --- & --- & --- \\
margin-mase & 0.7019 & $-0.35$ & 0.080 & 17/9 & $+12.07$ & 0.4823 & $-0.15$ & 0.980 & 15/11 \\
top2-mase & 0.6990 & $-0.76$ & $<$0.001 & 63/29 & $+6.08$ & 0.4792 & $-0.79$ & 0.021 & 64/29 \\
top3-mase & 0.6978 & $-0.93$ & $<$0.001 & 65/27 & $+8.39$ & 0.4780 & $-1.04$ & 0.002 & 66/27 \\
mixtop2-mase & 0.6991 & $-0.74$ & 0.020 & 58/35 & $+5.67$ & 0.4791 & $-0.81$ & 0.026 & 57/36 \\
\textbf{mixtop3-mase} & \textbf{0.6975} & \textbf{\boldmath$-0.97$} & \textbf{0.004} & \textbf{55/38} & \textbf{\boldmath$+7.92$} & \textbf{0.4778} & \textbf{\boldmath$-1.07$} & \textbf{0.004} & \textbf{60/33} \\
routemix05-crps & 0.6996 & $-0.67$ & 0.003 & 63/29 & $+6.08$ & 0.4796 & $-0.70$ & 0.004 & 67/26 \\
tempvinc-crps & 0.6997 & $-0.66$ & 0.005 & 59/33 & $+7.22$ & 0.4794 & $-0.74$ & 0.056 & 61/32 \\
tempvinc-mase & 0.6991 & $-0.73$ & $<$0.001 & 62/30 & $+13.31$ & 0.4798 & $-0.66$ & 0.039 & 65/28 \\
tempmix-mase & 0.6985 & $-0.82$ & 0.014 & 54/39 & $+9.61$ & 0.4789 & $-0.84$ & 0.045 & 56/37 \\
\bottomrule
\end{tabular}}
\end{table}

\begin{table}[h]
\centering\small
\caption{\textbf{Per-domain CRPS change on \gift} under the deployed
\texttt{mixtop3-mase} policy. We report the mean $\Delta\%$ across
configs in each domain, with win/loss counts in parentheses. This is the
domain-level counterpart of the fingerprint analysis in \cref{fig:fingerprint}
and \cref{sec:exp2}: the pattern varies substantially across models. For example,
Econ/Fin ranges from $-4.00\%$ for \mchronos\ to $+1.53\%$ for \mtirex, while
Web/CloudOps ranges from $-1.12\%$ for \mtoto\ to $+2.46\%$ for \mrone.
Healthcare, sales, and energy are the only domains where all four models improve.}
\label{tab:app:perdomain}
\begin{tabular}{lrrrrrrr}
\toprule
model & Econ/Fin & Energy & Health & Nature & Sales & Transport & Web/CloudOps \\
\midrule
\textsc{Chronos-2} & $-4.00$ & $-0.40$ & $-7.07$ & $-1.16$ & $-0.19$ & $-0.17$ & $-0.12$ \\
 & (4/2) & (20/12) & (4/1) & (8/7) & (3/0) & (11/4) & (10/7) \\[2pt]
\textsc{TiRex-2} & $+1.53$ & $-0.58$ & $-5.50$ & $+0.31$ & $-0.22$ & $-0.17$ & $+0.37$ \\
 & (1/5) & (19/12) & (5/0) & (6/9) & (3/0) & (9/6) & (10/7) \\[2pt]
\textsc{PatchTST-FM-R1} & $-0.75$ & $-0.54$ & $-1.64$ & $-0.34$ & $-0.17$ & $-0.63$ & $+2.46$ \\
 & (3/3) & (20/12) & (4/0) & (10/5) & (3/0) & (12/3) & (11/5) \\[2pt]
\textsc{Toto-2-313M} & $-0.87$ & $-0.16$ & $-5.52$ & $-0.28$ & $-0.42$ & $+0.33$ & $-1.12$ \\
 & (2/4) & (20/12) & (5/0) & (8/7) & (3/0) & (8/6) & (12/5) \\[2pt]
\bottomrule
\end{tabular}
\end{table}

Energy contains the largest number of configs and improves for all four
models, while Healthcare is the most consistent domain by win/loss count
(at most one loss for every model). Web/CloudOps is flat or harmful on \gift\ (except
\mtoto), but the same pattern does not appear on \boom, which consists
entirely of observability data and shows large overall gains in
\cref{tab:main}. One relevant difference is the amount of available backtest
evidence: the Web/CloudOps configs in \gift\ are relatively short, whereas the
\boom\ tasks provide substantially more windows for the gate. This comparison
is descriptive and does not isolate history length from domain effects.

\paragraph{Benchmark comparability notes.}
\begin{itemize}[leftmargin=1.2em,itemsep=1pt,topsep=2pt]
  \item \textbf{\boom.} Our \boom\ results are directly comparable to the public
    leaderboard. We evaluate the full official universe of $7413$
    dataset$\times$term tasks and use DataDog's evaluation code, which produces the
    $6750$-task scaled set, replaces non-finite raw metrics by column means,
    normalizes by the published seasonal-naive run, and then computes the shifted
    geometric mean with $\varepsilon=10^{-5}$. We also recompute all $23$ published
    systems through the same pipeline and verify the resulting scores against the \boom\ leaderboard; for example,
    \mtoto\ approximately reproduces at $0.585/0.353$.
    
    Note: our evaluations are univariate, whereas published
    leaderboard entries use each model's native configuration, although our
    \mtoto\ base approximately reproduces the published \mtoto\ result for the same checkpoint.
    Also, the $663$ tasks with zero seasonal-naive score, which the leaderboard
    handles separately, are not part of the scaled table reported here.
    \item \textbf{\fev.} We report \fev\ using its native raw metric means together
    with skill relative to the published seasonal-naive baseline. Temperature-family
    policies achieve higher skill but worse raw-mean CRPS and do not produce results
    for $8$ short-horizon tasks. The deployed mix, route, and top-$k$ policies
    improve both raw means and skill across all $100$ tasks.
\end{itemize}

\subsection{Generalization to Longer Forecasting Horizons}
\label{sec:app:long-horizon}

We evaluate whether \bgaa's improvements extend to longer
forecasting horizons on five standard long-horizon forecasting datasets: ETTh1,
ETTh2, ETTm1, ETTm2, and Weather. We test \mchronos\
and \mtoto\ at $H\in\{96,192,336,720\}$, using input
contexts of 2048 steps and the same daily test origins
across horizons. Each channel is normalized using
training-split statistics, and performance is measured
using normalized MSE and MAE. The frozen base and
\bgaa\ are evaluated on identical forecast windows.

\begin{table}[h!]
\centering
\small
\setlength{\tabcolsep}{5pt}
\caption{\textbf{BGAA generalizes to longer forecasting horizons.}
Normalized MSE and MAE for \mchronos\ and \mtoto\ on
ETTh1, ETTh2, ETTm1, ETTm2, and Weather, with context
length 2048 and horizons $H\in\{96,192,336,720\}$.
$\Delta\%$ denotes relative change versus the frozen base
(negative is better; computed from displayed scores).
Bold marks the better result within each comparison.
\textbf{BGAA improves both metrics on 18/20 and 15/20
dataset--horizon settings for \mchronos\ and \mtoto,
respectively, with gains persisting at $H=720$.}}
\label{tab:app:long-horizon}
\begin{tabular}{lrrrrrrr}
\toprule
& & \multicolumn{3}{c}{Normalized MSE}
& \multicolumn{3}{c}{Normalized MAE} \\
\cmidrule(lr){3-5}\cmidrule(lr){6-8}
Model & $H$
& Base & \bgaa & $\Delta\%$
& Base & \bgaa & $\Delta\%$ \\
\midrule
\mchronos
& 96
& 0.2405 & $\mathbf{0.2362}$ & $-1.79$
& 0.2723 & $\mathbf{0.2707}$ & $-0.59$ \\
& 192
& 0.2920 & $\mathbf{0.2870}$ & $-1.71$
& 0.3105 & $\mathbf{0.3083}$ & $-0.71$ \\
& 336
& 0.3328 & $\mathbf{0.3232}$ & $-2.88$
& 0.3402 & $\mathbf{0.3371}$ & $-0.91$ \\
& 720
& 0.3901 & $\mathbf{0.3863}$ & $-0.97$
& 0.3827 & $\mathbf{0.3815}$ & $-0.31$ \\
\midrule
\mtoto
& 96
& $\mathbf{0.2386}$ & 0.2392 & $+0.25$
& $\mathbf{0.2673}$ & 0.2677 & $+0.15$ \\
& 192
& 0.2867 & $\mathbf{0.2852}$ & $-0.52$
& 0.3055 & $\mathbf{0.3049}$ & $-0.20$ \\
& 336
& 0.3241 & $\mathbf{0.3139}$ & $-3.15$
& 0.3354 & $\mathbf{0.3333}$ & $-0.63$ \\
& 720
& 0.3830 & $\mathbf{0.3806}$ & $-0.63$
& 0.3793 & $\mathbf{0.3786}$ & $-0.18$ \\
\bottomrule
\end{tabular}
\end{table}

Table~\ref{tab:app:long-horizon} shows that \bgaa\
improves both metrics on 18 of 20 dataset--horizon
settings for \mchronos\ and 15 of 20 for \mtoto.
Aggregated MSE improves at every horizon for \mchronos,
with the largest relative reduction at $H=336$
($2.88\%$). For \mtoto, the improvement is also largest
at $H=336$ ($3.15\%$), although performance slightly
regresses at $H=96$. Both models retain improvements
at $H=720$, demonstrating that historical verification
can remain useful for substantially longer prediction
horizons. However, improvements are generally modest
and do not increase monotonically with horizon.

\subsection{Verification headroom and label-based diagnostics (\cref{tab:capturability})}
\label{sec:app:exp8}

We retain the per-window pinball loss on the realized test horizon for every
view and use it to construct three label-based diagnostics. The
\emph{cross-fitted best-view gain} selects one fixed view per config,
using disjoint halves of the windows for selection and evaluation. The
\emph{label-ranked gate} keeps the deployed policy unchanged but replaces the
backtest ranking with a ranking computed from labeled held-out windows. The
\emph{per-window hindsight selection} selects the lowest-loss view separately for
each forecast window using that window's labels. These diagnostics use future
labels only for analysis; the deployed policy does not.

\paragraph{Evaluation protocol.}
We score all diagnostics with the same gluonts MASE and CRPS metrics used in
\cref{tab:main} and apply the same aggregation procedure. Each reported value is
the relative change in the aggregate metric with respect to the base forecast,
computed over the same set of windows.

\begin{itemize}
\item \emph{Windows.} For each config, we use
  $W=\min(1024,N)$ evenly spaced test windows from the evaluation sampler used for the main results.
  We evaluate the identity forecast together with every available non-identity view in the deployed catalogue on these windows and retain
  the per-window losses. For MASE, we drop windows with zero seasonal error to
  match gluonts.

\item \emph{Deployed policy.} We evaluate the view set admitted by the
  deployed run using the deployed CDF mixture. This reproduces the
  per-config results underlying \cref{tab:main} to within $10^{-7}$
  relative error.

\item \emph{Cross-fitting split.} We divide the $W$ windows into two contiguous
  halves and remove the $\lceil H/\mathrm{stride}\rceil$ windows closest to the
  split boundary so that selection and evaluation labels do not overlap.
  We then swap the two halves and pool the held-out evaluations. We require at
  least $8$ retained windows per half, which leaves $53$ of the $97$
  configs. On the remaining $44$ shorter configs, the deployed
  policy improves by $-0.1$ to $1.2\%$ across models.

\item \emph{Label-ranked gate.} We keep the deployed top-$3$ selection rule
  but rank views using labeled test-window losses instead of backtest losses.
  The \emph{in} variant ranks on the same half used for evaluation; the reported
  \emph{x-fit} variant ranks on the opposite half.

\item \emph{Best-view gain.} We select a single view, including identity, with
  the lowest CRPS on the selection half and evaluate it on the other half.

\item \emph{Per-window hindsight selection.} For each evaluation window, we select
  the view with the lowest summed pinball loss using that window's realized
  labels.
\end{itemize}

On the $53$ cross-fittable configs, the deployed policy's all-window and
held-out improvements differ by less than $0.1$ percentage points for every
model.

\begin{table}[t]
\centering
\small
\caption{\textbf{Verification headroom is much larger at the forecast-window
level.}
CRPS improvement over the frozen base in \%. The first ``realized'' column uses
all $97$ configs; the others use the $53$ configs that can be cross-fit.
The \emph{label-ranked gate} keeps the deployed policy fixed but ranks views
with held-out labels. The reported realized/label-ranked ratio is the realized
improvement divided by the label-ranked improvement. The \emph{best-view gain (x-fit)} selects one view
per config on one half of its windows and evaluates it on the other.
The \emph{per-window hindsight selection} selects the best view per window using
its realized labels. These label-based quantities are diagnostics, not
deployable verifier targets.}
\label{tab:capturability}
\resizebox{\linewidth}{!}{
\begin{tabular}{lr@{\hskip 14pt}rrrrr}
\toprule
 & all $97$ configs & \multicolumn{5}{c}{$53$ configs long enough to cross-fit} \\
\cmidrule(lr){2-2}\cmidrule(lr){3-7}
model & realized (\cref{tab:main}) & realized & label-ranked gate &
realized / label-ranked & best-view gain (x-fit) & per-window hindsight \\
 & \% & \% & \% & ratio & \% & \% \\
\midrule
\mchronos & 1.07 & 1.32 & 2.06 & 0.64 & 4.10 & 19.4 \\
\mtirex   & 0.32 & 0.50 & 1.21 & 0.41 & 1.48 & 16.6 \\
\mrone    & 0.19 & 0.41 & 1.20 & 0.34 & 1.42 & 17.5 \\
\mtoto    & 0.66 & 0.61 & 1.16 & 0.52 & 1.37 & 16.6 \\
\bottomrule
\end{tabular}}
\end{table}

\begin{table}[t]
\centering
\small
\caption{\textbf{Supporting diagnostics for \cref{tab:capturability}.}
Results are reported on the same $53$ cross-fittable configs as
improvement over the base in percent (positive is better). \emph{In} selects
and evaluates on the same windows, while \emph{x-fit} selects on one half and
evaluates on the other. The MASE block reports the same diagnostics using the
point-forecast metric.}
\label{tab:capturability-appendix}
\resizebox{\linewidth}{!}{
\begin{tabular}{l rr rr @{\hskip 14pt} rrr rr r}
\toprule
 & \multicolumn{4}{c}{CRPS} & \multicolumn{6}{c}{MASE} \\
\cmidrule(lr){2-5}\cmidrule(lr){6-11}
 & \multicolumn{2}{c}{best-view gain} & \multicolumn{2}{c}{label-ranked gate} & realized & label-ranked & realized/ & \multicolumn{2}{c}{best-view gain} & per-window \\
model & in & x-fit & in & x-fit & & gate (x-fit) & label-ranked & in & x-fit & hindsight \\
\midrule
\mchronos  & 5.36 & 4.10 & 2.01 & 2.06 & 0.92 & 1.39 & 0.67 & 2.44 & 0.76 & 20.7 \\
\mtirex    & 3.43 & 1.48 & 1.06 & 1.21 & 0.23 & 0.75 & 0.31 & 1.79 & 1.09 & 19.0 \\
\mrone     & 3.54 & 1.42 & 1.09 & 1.20 & 0.35 & 1.23 & 0.28 & 3.46 & 2.36 & 21.3 \\
\mtoto     & 2.95 & 1.37 & 1.26 & 1.16 & 1.08 & 1.44 & 0.75 & 1.74 & 0.72 & 19.4 \\
\bottomrule
\end{tabular}}
\end{table}

\Cref{tab:capturability} summarizes these diagnostics for each model. The
realized/label-ranked ratio is the realized improvement divided by the improvement
from the label-ranked gate on the same $53$ configs. This measures how much of
the gain available to the fixed policy is recovered by the leak-free backtest
ranking. In-sample best-view selection is optimistic: it exceeds the
cross-fitted gain by $1.3$--$2.5\times$
(\cref{tab:capturability-appendix}), which is why we use the split above. The
label-ranked gate is less sensitive, with in-sample and x-fit results differing
by at most $0.2$ percentage points.

For MASE, the realized/label-ranked ratio ranges from $0.28$ to $0.75$
across models. We therefore interpret this
quantity as a ranking diagnostic, not as a ceiling: under distribution shift,
labels from the opposite half can rank views worse than the recent backtest.
The two halves are also contiguous and autocorrelated, and their contexts may
overlap near the split boundary, so cross-fitting reduces label reuse without
making the samples independent. The per-window hindsight best-view diagnostic should be read
similarly. It uses the realized outcome and selects only among individual views;
for CRPS, an aggregate can outperform every member
(\cref{thm:ambiguity}), so this hindsight best-view diagnostic is not an upper bound over aggregation
policies.

\begin{figure}[t]
  \centering
  \includegraphics[width=0.85\textwidth]{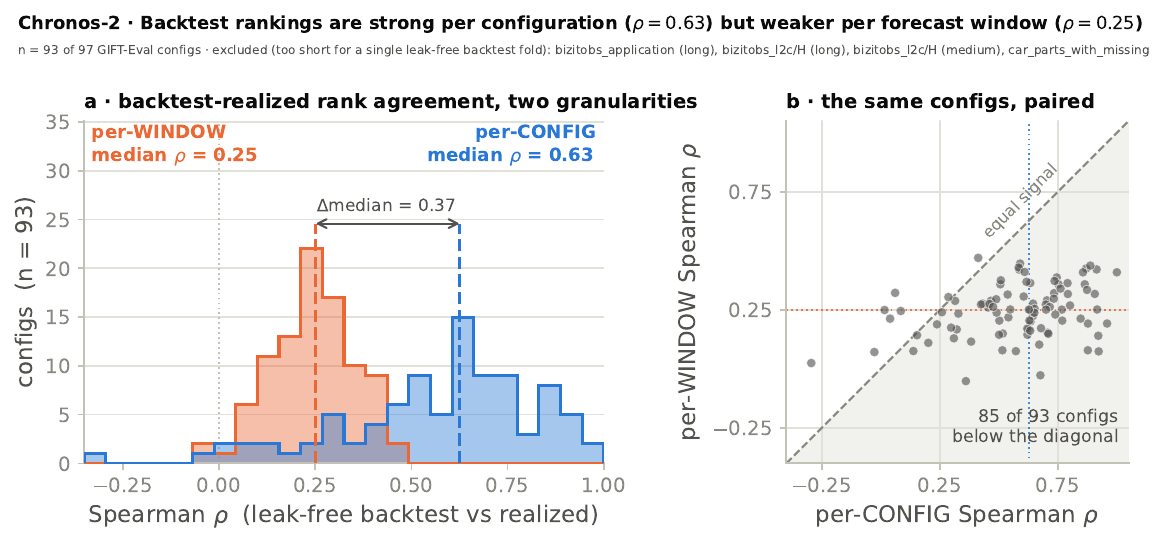}
  \caption{\textbf{Backtest rankings are strong per configuration but much
  weaker per forecast window.}
  For \mchronos\ on $93$ \gift\ configurations, median Spearman $\rho$
  between leak-free backtest and realized view rankings falls from $0.63$
  per configuration to $0.25$ per window. Results use $8{,}313$ sampled
  windows and MASE.}
  \label{fig:capturability-main}
\end{figure}

\begin{figure}[h]
  \centering
  \includegraphics[width=\textwidth]{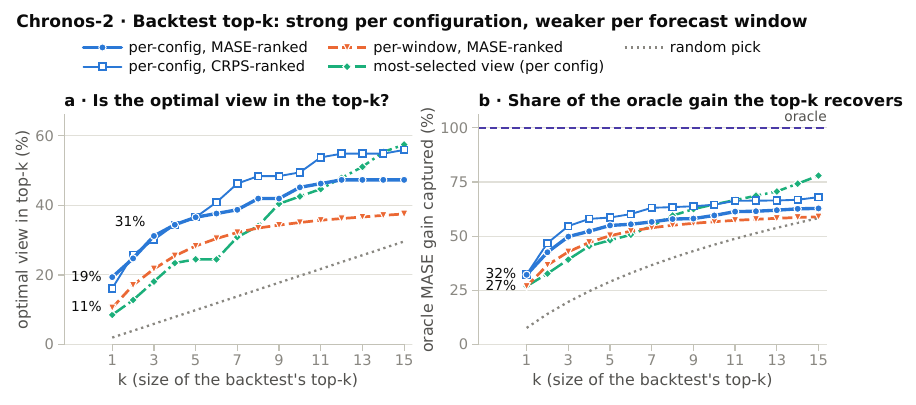}
  \caption{\textbf{Agreement between backtest and hindsight view rankings for \mchronos\ on \gift.}
(a)~Fraction of configs whose realized best view lies in the backtest's top-$k$.
Ranked per config, the backtest's top-$1$ contains the best view for $19$--$20\%$ of
configs and its top-$3$ for $31$--$33\%$, well above both the most-selected-view
heuristic ($10\%$ at $k{=}1$) and random selection ($1.3\%$ at $k{=}1$).
The top-$1$ hit rate is therefore about $15\times$ chance. Ranked per window, the
top-$1$ hit rate falls to about $11\%$.
(b)~Share of the hindsight MASE gain recovered by the top-$k$: the per-config
top-$1$ recovers $32$--$33\%$, while the per-window top-$1$ recovers about $27\%$.
Because the per-window target depends on the realized outcome, we use it only
as a diagnostic rather than as an achievable target for a verifier that
observes only $x$.}

  \label{fig:topk}
\end{figure}

\paragraph{Configuration-level versus window-level ranking.}
The ranking diagnostics show the same separation. On \mchronos, the median
Spearman correlation between backtest and realized view rankings is $0.63$
when rankings are aggregated per configuration, but only $0.25$ when computed
per forecast window (\cref{fig:capturability-main}). The top-$k$ hit-rate and
gain-recovery diagnostics show the same degradation at the window level
(\cref{fig:topk}). Because the per-window targets use the realized outcome,
they quantify ranking difficulty and heterogeneity rather than an achievable
target for a context-only verifier.

\paragraph{Why averaging more backtest windows does not solve window-level selection.}
The gate statistic averages relative loss over $m$ historical windows. Under
the usual independent-noise heuristic, the sampling component of its standard
error decreases on the order of $1/\sqrt{m}$. The configuration-level gate
uses $m=8$--$128$ windows, whereas a window-specific comparison based on a
single realized window has no analogous averaging. This heuristic helps
explain why configuration-level rankings are more stable, but it is not a
finite-sample guarantee: the gate windows are dependent time-series samples,
and the realized future window may come from a different local regime. More
historical windows can stabilize a configuration-level average; they do not by
themselves identify which view will be best on a particular future window.

\paragraph{Fine-grained weighting is not better supported by the current signal.}
The policy ablation is consistent with this limitation. The
temperature-weighted gate uses more of the relative backtest-score information:
it achieves the strong CRPS among the tested policies but the worst MASE
(\cref{tab:policy-ablation}). Fixed top-$k$ instead uses the backtest primarily
to identify a small set of promising views and is less sensitive to errors in
their fine ordering. This is evidence about the policies tested here, not a
general claim that weighting cannot help.

\subsection{Pooled versus Per-Series Historical Selection}
\label{sec:app:pooling}

We investigate whether selecting views independently for each
series improves over the deployed configuration-level gate.
On 41 \gift\ configurations containing 1,972 series,
we compare the pooled gate, which ranks views using
historical evidence shared across a configuration, against
per-series selection using only each series's own backtests.
Both approaches use the same sampled historical evidence
and candidate views; only the granularity of selection changes.
For configurations with fewer than eight usable historical
windows, we use the lighter gate described in the experimental
setup, rather than requiring the standard eight-window minimum.
We group configurations by series count to examine the effect
of sparse per-series evidence, with 1--47 backtests available
per series depending on the model and configuration.

\begin{table}[t]
\centering
\small
\setlength{\tabcolsep}{4pt}
\caption{\textbf{Configuration-pooled versus per-series historical
view selection.}
Results on 41 \gift\ configurations (1,972 series) across
three TSFMs, grouped by the number of series per configuration.
MASE changes are relative to the frozen base (\%; lower is
better). Gap is the paired per-series minus per-config MASE
log-ratio difference (95\% configuration-bootstrap CI);
positive values favor per-config selection. Backtests/series reports
the historical evidence available to individual selectors.
\textbf{Per-config selection performs better on average across all three
models, with the largest reliable gains when per-series
evidence is sparse.}}
\label{tab:app:pooling}
\resizebox{\linewidth}{!}{%
\begin{tabular}{llrrrrrl}
\toprule
Model & Series/config
& Configs & Series & Backtests/series
& Per-config MASE & Per-series MASE
& Gap [95\% CI] \\
\midrule
\mchronos
& Low ($2$--$7$)
& 17 & 114 & 47
& $-0.26$ & $+0.69$
& $+0.94$ {\scriptsize[$-1.27$, $4.41$]} \\
& Medium ($21$)
& 7 & 147 & 18
& $-0.88$ & $-1.01$
& $-0.13$ {\scriptsize[$-1.17$, $1.09$]} \\
& High ($30$--$42$)
& 6 & 204 & 12
& $-0.48$ & $-0.69$
& $-0.21$ {\scriptsize[$-0.95$, $0.39$]} \\
& Very high ($118$--$156$)
& 11 & 1,507 & 3
& $-2.15$ & $-0.67$
& $\mathbf{+1.50}$ {\scriptsize[0.50, 2.70]} \\
\cmidrule(lr){2-8}
& \textbf{All (41)}
& \textbf{41} & \textbf{1,972} & 3--47
& $\mathbf{-0.91}$ & $-0.17$
& $+0.74$ {\scriptsize[$-0.32$, $2.34$]} \\
\midrule

\mtirex
& Low ($2$--$7$)
& 17 & 114 & 47
& $-0.47$ & $+0.07$
& $+0.54$ {\scriptsize[$-0.22$, $1.56$]} \\
& Medium ($21$)
& 7 & 147 & 18
& $-0.05$ & $-0.28$
& $-0.23$ {\scriptsize[$-0.78$, $0.31$]} \\
& High ($30$--$42$)
& 6 & 204 & 12
& $-0.05$ & $+0.53$
& $+0.58$ {\scriptsize[$-0.55$, $2.05$]} \\
& Very high ($118$--$156$)
& 11 & 1,507 & 3
& $-1.62$ & $-1.15$
& $\mathbf{+0.48}$ {\scriptsize[0.08, 0.88]} \\
\cmidrule(lr){2-8}
& \textbf{All (41)}
& \textbf{41} & \textbf{1,972} & 3--47
& $\mathbf{-0.65}$ & $-0.25$
& $+0.40$ {\scriptsize[$-0.01$, $0.89$]} \\
\midrule

\mtoto
& Low ($2$--$7$)
& 17 & 114 & 6
& $-0.69$ & $+0.84$
& $\mathbf{+1.52}$ {\scriptsize[0.31, 3.05]} \\
& Medium ($21$)
& 7 & 147 & 6
& $-0.56$ & $-0.76$
& $-0.20$ {\scriptsize[$-0.82$, $0.44$]} \\
& High ($30$--$42$)
& 6 & 204 & 3
& $-0.14$ & $+0.04$
& $+0.18$ {\scriptsize[$-0.82$, $0.90$]} \\
& Very high ($118$--$156$)
& 11 & 1,507 & 1
& $-0.98$ & $+0.01$
& $+0.99$ {\scriptsize[$-0.05$, $2.29$]} \\
\cmidrule(lr){2-8}
& \textbf{All (41)}
& \textbf{41} & \textbf{1,972} & 1--6
& $\mathbf{-0.66}$ & $+0.22$
& $\mathbf{+0.89}$ {\scriptsize[0.23, 1.67]} \\
\bottomrule
\end{tabular}%
}
\end{table}

Table~\ref{tab:app:pooling} shows that per-config selection
achieves better average MASE across all three models:
$-0.91\%$ versus $-0.17\%$ for \mchronos,
$-0.65\%$ versus $-0.25\%$ for \mtirex, and
$-0.66\%$ versus $+0.22\%$ for \mtoto.
The clearest pattern
appears in configurations with 118--156 series,
where each series provides only 1--3 backtests:
pooling significantly outperforms per-series selection
for both \mchronos\ and \mtirex. These findings
suggest that sharing historical verification evidence
across related series can improve selection reliability,
particularly when individual series provide too little
evidence for stable view ranking.

\subsection{Qualitative examples}
\label{sec:app:qualitative}

We additionally inspect a small set of forecast windows where the deployed
BGAA policy improves over the frozen base model. These examples illustrate  the behavior of \bgaa, and each plot follows the same format. The left panel shows the base context
together with the inputs or forecast-side transformations associated with the
views admitted by the gate. The middle panel compares the forecasts produced
by these views before aggregation, together with the resulting BGAA mixture.
The right panel compares the base and final predictive distributions against
the realized forecast horizon. This makes it possible to inspect how the
admitted views alter the frozen model's prediction and how their combination
changes the final forecast in windows where BGAA is successful.

\begin{figure}[t]
    \centering

    \begin{subfigure}{0.9\linewidth}
        \centering
        \includegraphics[width=\linewidth]{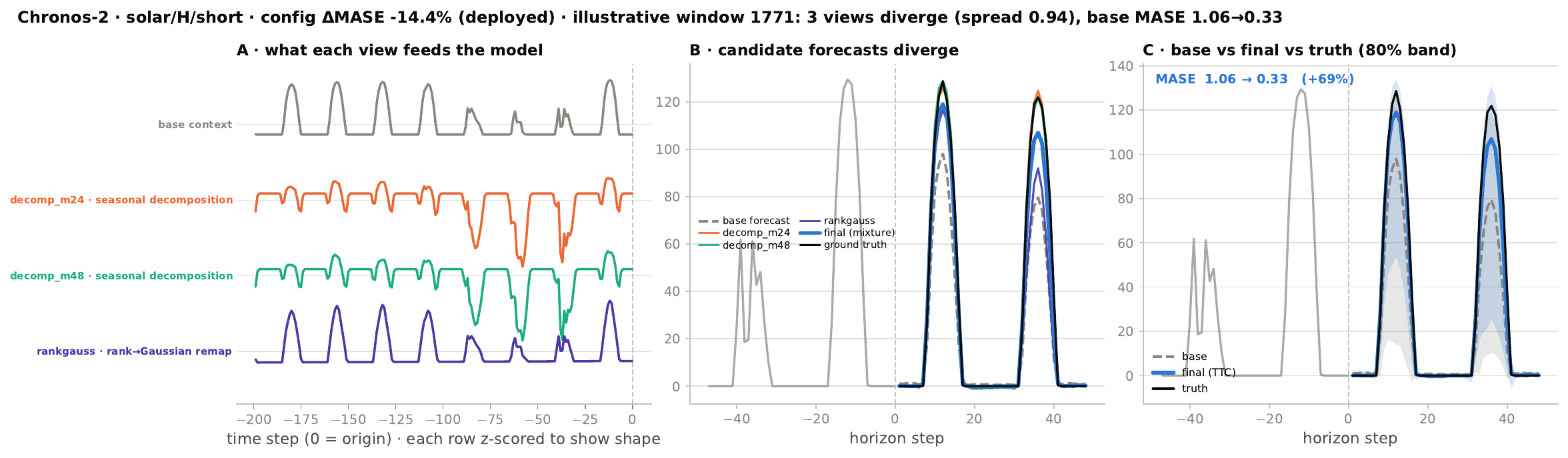}
        \label{fig:qualitative-1a}
    \end{subfigure}

    \begin{subfigure}{0.9\linewidth}
        \centering
        \includegraphics[width=\linewidth]{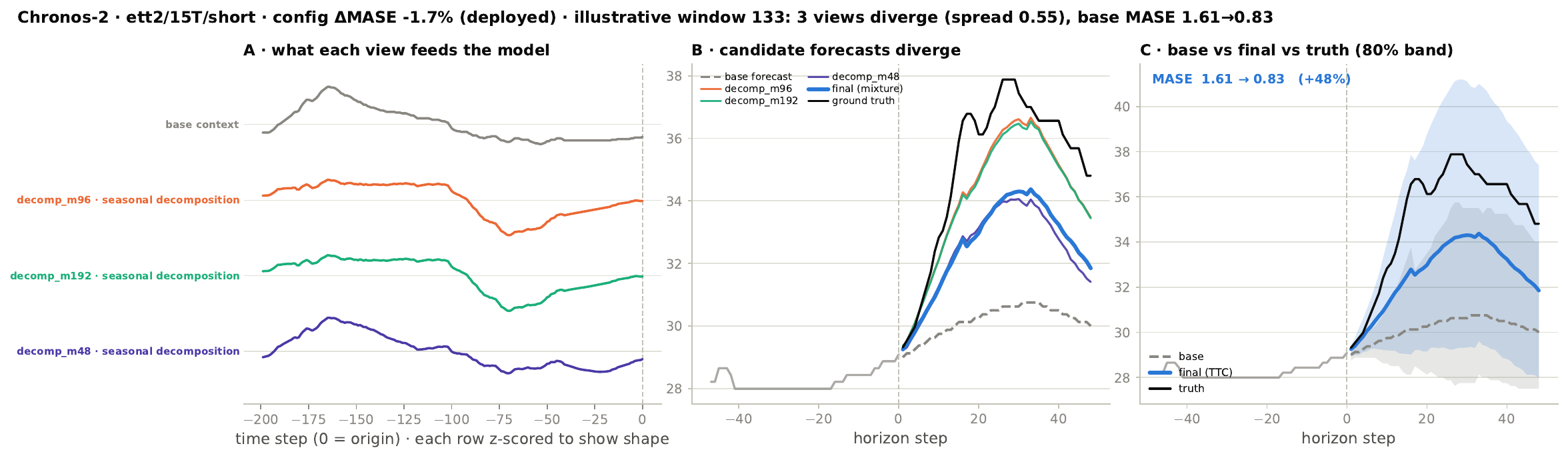}
        \label{fig:qualitative-1b}
    \end{subfigure}

    \begin{subfigure}{0.9\linewidth}
        \centering
        \includegraphics[width=\linewidth]{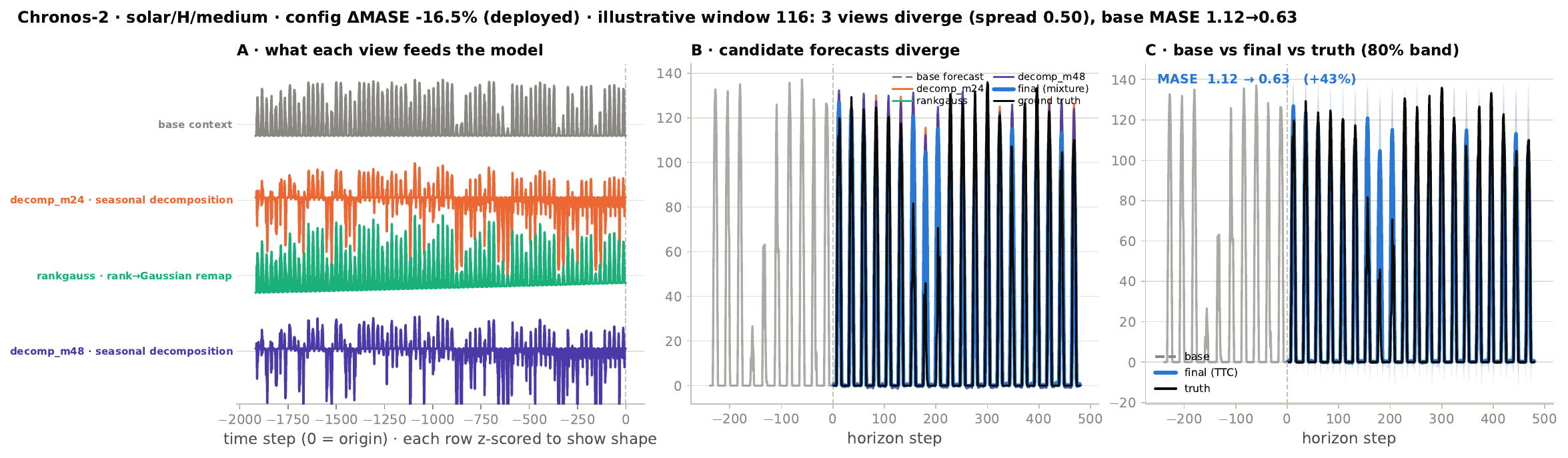}
        \label{fig:qualitative-1c}
    \end{subfigure}

    \begin{subfigure}{0.99\linewidth}
        \centering
        \includegraphics[width=\linewidth]{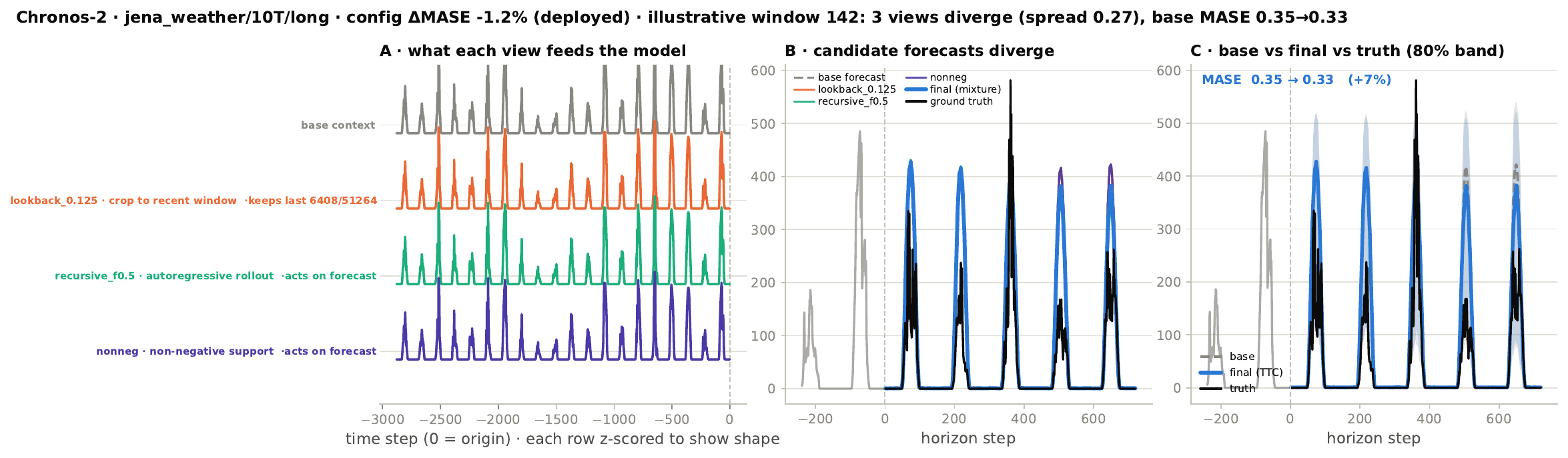}
        \label{fig:qualitative-1d}
    \end{subfigure}

    \caption{\textbf{Qualitative examples for \bgaa\ improvements.}
    Selected forecast windows for which BGAA improves over the frozen base
    model. Each subfigure shows the admitted views, their candidate forecasts,
    and the resulting BGAA prediction relative to the realized horizon.}
    \label{fig:qualitative-1}
\end{figure}

\begin{figure}[t]
    \centering

    \begin{subfigure}{0.9\linewidth}
        \centering
        \includegraphics[width=\linewidth]{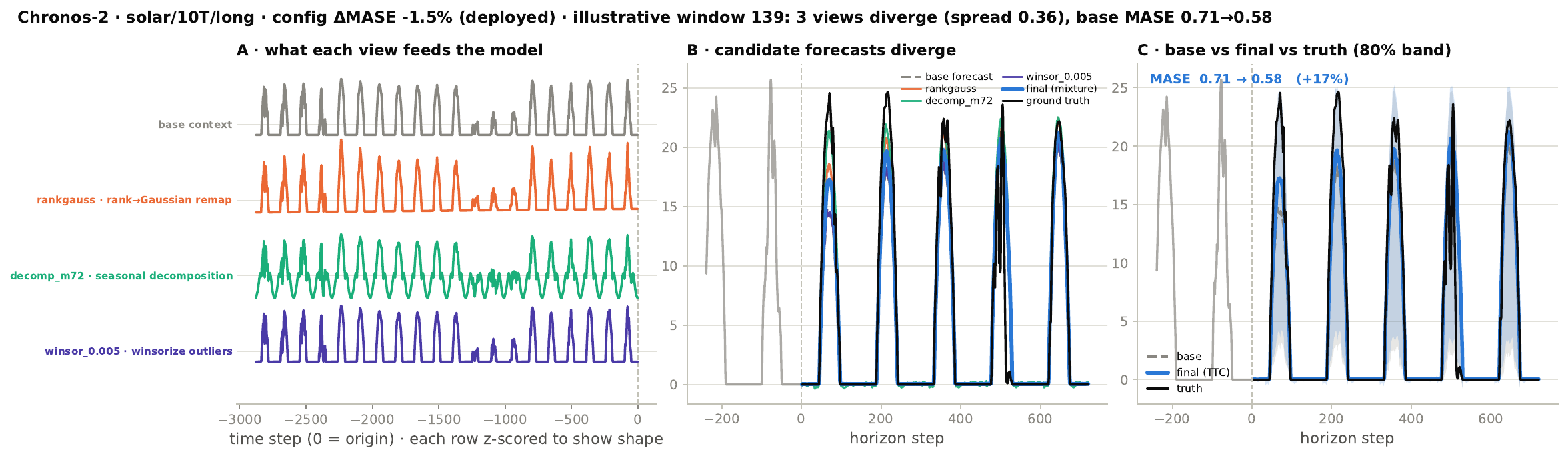}
        \label{fig:qualitative-2a}
    \end{subfigure}

    \begin{subfigure}{0.9\linewidth}
        \centering
        \includegraphics[width=\linewidth]{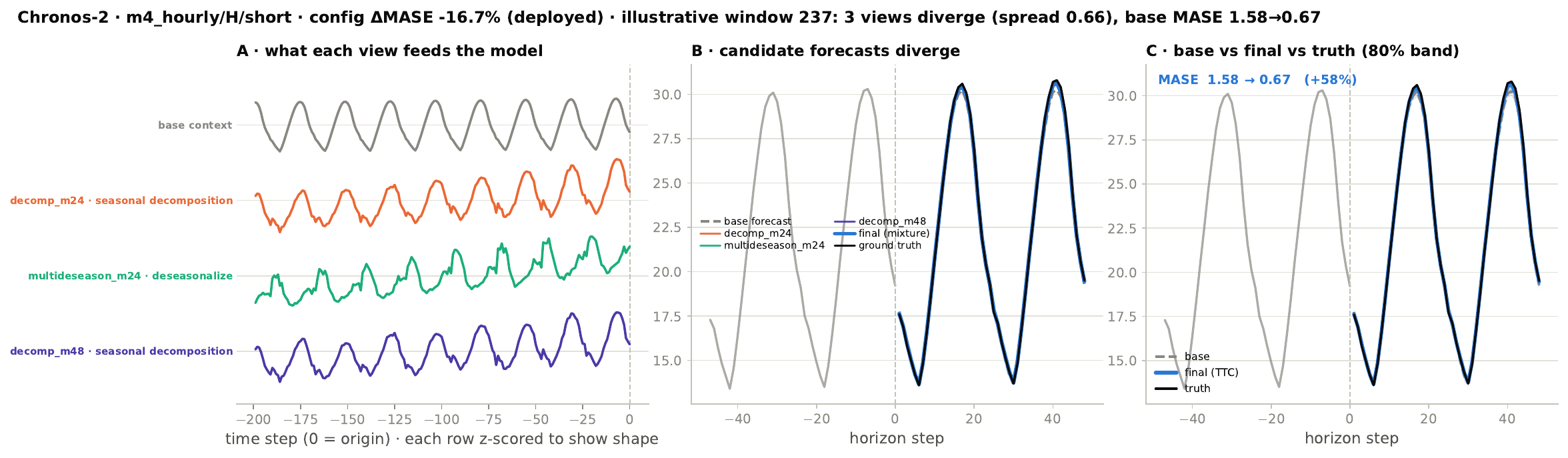}
        \label{fig:qualitative-2b}
    \end{subfigure}

    \begin{subfigure}{0.9\linewidth}
        \centering
        \includegraphics[width=\linewidth]{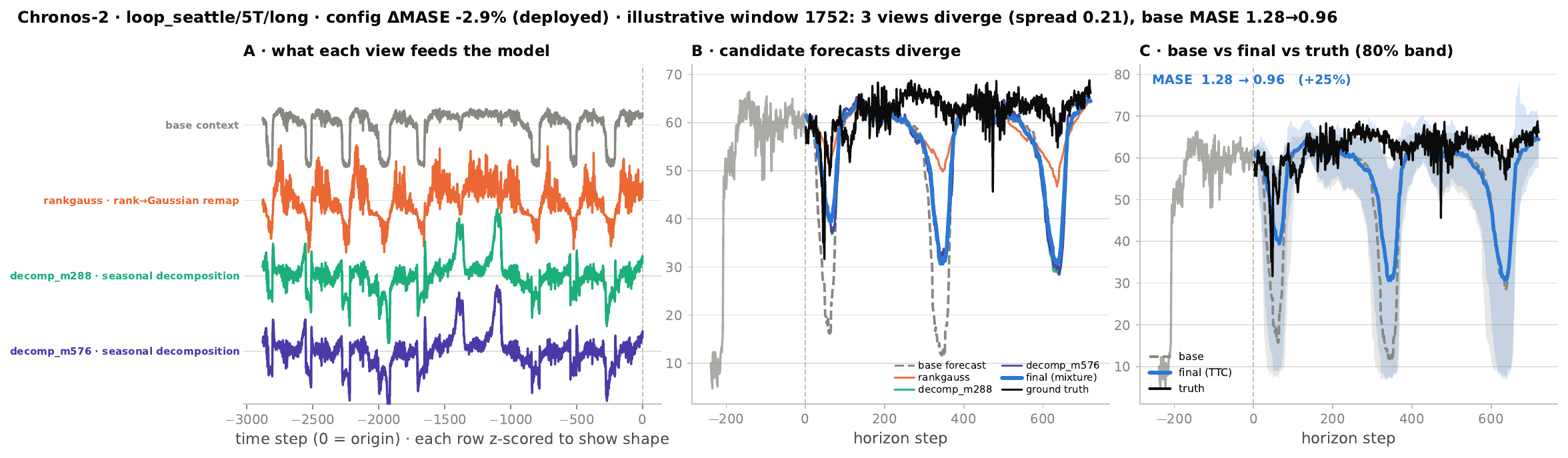}
        \label{fig:qualitative-2c}
    \end{subfigure}

    \begin{subfigure}{0.9\linewidth}
        \centering
        \includegraphics[width=\linewidth]{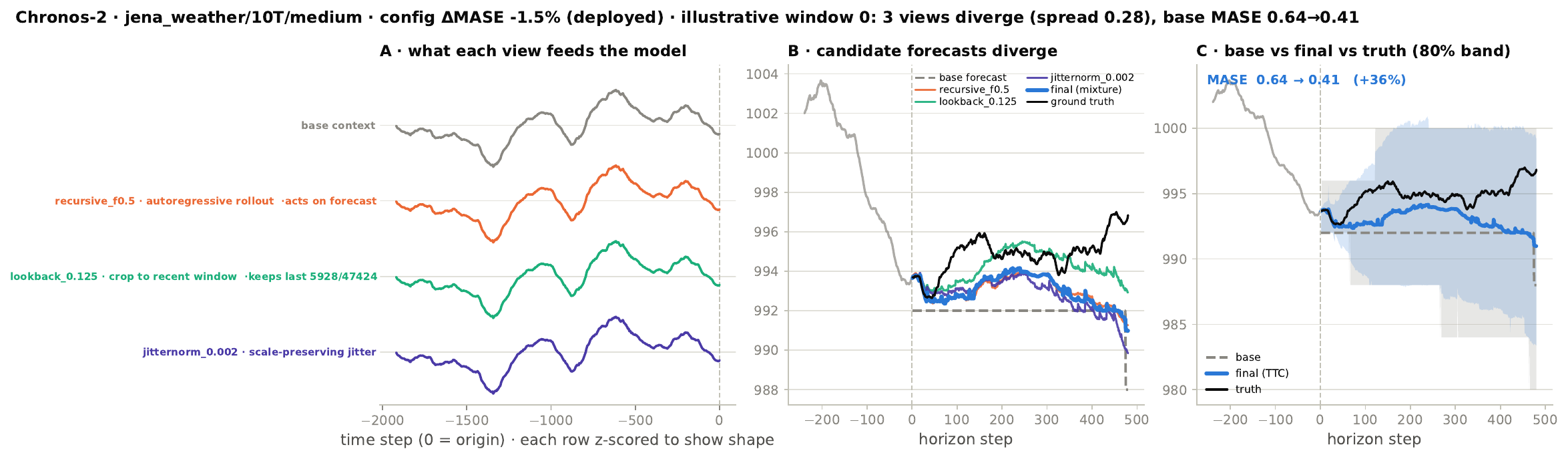}
        \label{fig:qualitative-2d}
    \end{subfigure}

    \caption{\textbf{Qualitative examples for \bgaa\ improvements.}
    Additional successful forecast windows under the deployed policy.}
    \label{fig:qualitative-2}
\end{figure}

\clearpage

%% file: appendix/D_mechanism_details.tex
\section{Mechanism experiments: protocols and full results}
\label{sec:app:mechanism}

This section gives the full protocols and results for the three mechanism
analyses in \cref{sec:mechanism}: fingerprinting, prior geometry, and forecast
signatures. When an experiment reuses forecast evaluation, we run it through
the same evaluation pipeline used for the main results, with the same candidate
pool, leak-free backtest, admission rules, aggregation, and scoring. We do not
modify the deployed outputs for these analyses.

Experiments that restrict the candidate pool or change the selection rule are reported separately with the controls in \cref{sec:app:controls}. Unless noted otherwise, the experiments in this section use a $15$-config \gift\ test \emph{subset} used for selected mechanism and forecast-signature analyses rather than the full set of $97$ configs.

\subsection{Fingerprinting (\cref{sec:exp2})}
\label{sec:app:exp2}

\begin{table}[h!]
\centering\small
\caption{\textbf{Solo-gated family benefit in $\Delta$MASE\%} (15 config \gift\ test subset;
more negative is better; $0$ indicates that the family was never admitted).
Each column gives the family-level fingerprint for one model; pairwise agreement
between these fingerprints is reported in \cref{tab:app:tau}. \textbf{Bold}
marks the best-performing family for each model. The preferred family differs
across models, and \mtirex\ has the flattest profile: its best family yields about a quarter of the gain of \mtoto's best family. Notably, \texttt{diff} is
the strongest family for \mchronos\ but slightly hurts \mtirex.}
\label{tab:app:fingerprint}
\begin{tabular}{lrrrr}
\toprule
family & \mchronos & \mtirex & \mrone & \mtoto \\
\midrule
\texttt{changepoint}   & $-0.034$ & $0$ & $0$ & $0$ \\
\texttt{decomp}        & $-0.814$ & $-0.084$ & $-0.109$ & $-0.214$ \\
\texttt{diff}          & $\mathbf{-0.849}$ & $+0.012$ & $-0.537$ & $-0.264$ \\
\texttt{fft}           & $-0.030$ & $0$ & $-0.094$ & $0$ \\
\texttt{jitter}        & $-0.370$ & $\mathbf{-0.362}$ & $-0.130$ & $-0.105$ \\
\texttt{jitternorm}    & $-0.362$ & $-0.357$ & $-0.127$ & $-0.104$ \\
\texttt{lookback}      & $-0.480$ & $-0.047$ & $\mathbf{-0.649}$ & $\mathbf{-1.390}$ \\
\texttt{multideseason} & $0$ & $0$ & $-0.065$ & $-0.035$ \\
\texttt{nonneg}        & $+0.002$ & $-0.068$ & $-0.075$ & $-0.003$ \\
\texttt{offset}        & $-0.080$ & $-0.016$ & $+0.002$ & $-0.980$ \\
\texttt{rankgauss}     & $-0.317$ & $-0.180$ & $-0.529$ & $0$ \\
\texttt{recursive}     & $+0.015$ & $-0.213$ & $-0.347$ & $-0.207$ \\
\texttt{seasontile}    & $-0.182$ & $-0.082$ & $-0.149$ & $0$ \\
\texttt{warp}          & $-0.397$ & $-0.113$ & $-0.104$ & $-0.270$ \\
\bottomrule
\end{tabular}
\end{table}

\begin{table}[h]
\centering\small
\caption{\textbf{Cross-model agreement between family fingerprints.}
We report Kendall's $\tau$ between the fingerprints in
\cref{tab:app:fingerprint}, with permutation $p$-values. Agreement is well below
$1$ for every model pair, indicating that useful families are not ranked
consistently across models. The two pairs involving \mtirex\ and a model whose
profile is dominated by \texttt{lookback} or \texttt{diff} are not
significantly different from zero agreement. This is consistent with a
model-specific ranking of view families, with some shared signal from families
that are broadly useful.}
\label{tab:app:tau}
\begin{tabular}{llrr}
\toprule
model A & model B & $\tau$ & perm.\ $p$ \\
\midrule
\mchronos & \mtirex & $0.089$ & $0.350$ \\
\mchronos & \mrone  & $0.297$ & $0.078$ \\
\mchronos & \mtoto  & $0.352$ & $0.049$ \\
\mtirex   & \mrone  & $0.223$ & $0.148$ \\
\mtirex   & \mtoto  & $0.012$ & $0.480$ \\
\mrone    & \mtoto  & $0.148$ & $0.253$ \\
\bottomrule
\end{tabular}
\end{table}

We evaluate each view family separately for each model by restricting the
candidate pool to that family plus the identity and applying the deployed
backtest gate unchanged. We then measure the resulting gain relative to that
model's own base forecast. Families are drawn from the full fixed pool, which overlaps the deployed tier in \cref{tab:app:pool} but differs from it in four ways: it adds \texttt{changepoint} and
\texttt{warp}, gates \texttt{jitter} and \texttt{jitternorm} over their full
7-degree grids ($\sigma\in[0.002,0.03]$) rather than the single deployed
$\sigma=0.002$, merges \texttt{sdiff} into \texttt{diff} and \texttt{supportfloor}
into \texttt{nonneg}, and omits the cleaning views (\texttt{smooth}, \texttt{ewma},
\texttt{winsor}). To compare family preferences across models, we
compute Kendall's $\tau$ between the resulting $14$-family benefit vectors.
Permutation $p$-values are estimated from $10^4$ random label shuffles.

\paragraph{Why the fingerprint uses the full pool.}
We fingerprint the full fixed pool rather than the deployed tier, for three
reasons. First, the deployed tier is itself the output of a label-informed admit
decision (\cref{sec:app:views}). Fingerprinting only its survivors would select
the family set on the quantity being compared. Second, that selection keeps the
families that helped. It truncates the benefit vectors, and truncation typically
pulls rank correlation toward zero, the direction our claim predicts. The full
pool removes the truncation and makes the test harder. Restricted to the
$12$ families it shares with the deployed tier, mean cross-model $\tau$ falls
from $0.19$ to $0.15$ (range $-0.09$ to $0.33$), and removing any single family
keeps every pair at $\tau\le0.45$. Third, gating each family over its full degree
grid measures it at the degree the gate prefers for each model, not at the one
fixed for deployment (\texttt{jitter}'s single $\sigma{=}0.002$). The grids are identical across models, so any advantage a larger grid gives is
shared by all four columns and would tend to add agreement, not remove it.

The differences are also qualitative. \mtirex's strongest family is
\texttt{jitter} ($-0.36\%$). \texttt{lookback} is strongest for \mrone\
($-0.65\%$) and \mtoto\ ($-1.39\%$), but gives little gain for \mtirex\
($-0.05\%$). Within \mchronos, admitted families also vary across domains
(\cref{fig:fingerprint}), supporting the use of config-specific selection.

We distinguish two notions of cross-model agreement. The \emph{solo-gated}
benefit vectors above have Kendall $\tau$ in the range $0.01$--$0.35$. In
contrast, when we compute the same comparison using per-config
\emph{admission} patterns over full \gift, the agreement is much higher
($0.50$--$0.72$). This is largely because admission is dominated by the two
crop families, which are used by all four models. For the mechanism analysis,
we therefore focus on agreement in realized family benefit rather than
agreement in whether a family is admitted.

\subsection{Series Characteristics and Feature-Based Selection}
\label{sec:app:char-routing}

We investigate whether simple statistical properties can predict
which transformation families are useful, using 30 configurations
(15 \gift, 15 \fev) across three TSFMs. For each series, we
compute STL seasonality strength, linear-trend $R^2$, lag-1
autocorrelation, and volatility lumpiness, then compare the lowest
and highest terciles. Table~\ref{tab:app:char-families} reports
the strongest family in each group using the realized performance
of an ungated identity--view mixture, alongside \bgaa's improvement
on the same windows. For example, \mchronos\ favors differencing
under low seasonality ($-6.4\%$) but decomposition under high
seasonality ($-4.3\%$), whereas \mtirex\ more often favors
multideseasoning. These are retrospective associations: the best
family is chosen using evaluation outcomes, and the confidence
intervals are not adjusted for this selection. Several effects
also fail to replicate across both benchmarks.

To test whether these properties provide a deployable alternative
to backtesting, we evaluate feature-based routers trained
leave-one-configuration-out. Table~\ref{tab:app:char-router}
compares them with historical top-1 selection at the same
one-view application budget. Across the three models,
the historical gate improves MASE by $1.7$--$4.4\%$,
whereas the four-characteristic router ranges from a
$1.5\%$ improvement to a $2.8\%$ regression.
None of the 36 router variants improves average MASE
over historical top-1 selection, and even the
post-hoc best-of-12 routers do not consistently match it.
Thus, statistical features help describe transformation
preferences, but these simple predictors do not reliably
replace the causal verification signal obtained from
historical forecast errors.

\begin{table}[t]
\centering
\small
\setlength{\tabcolsep}{3pt}
\caption{\textbf{View-family preferences across time-series
characteristics.}
For the lowest and highest terciles of four statistical
characteristics, we report the hindsight-best family
and its MASE change versus base (\%, 95\% bootstrap CI).
The second line in each cell gives the corresponding
\bgaa\ improvement on the same windows.
Results cover 30 configurations (15 \gift, 15 fev-bench)
across three TSFMs. Negative changes indicate improvement;
$^\dagger$ marks effects not replicated on both benchmarks.
\textbf{Preferred transformations vary across models and
series characteristics, motivating adaptive view selection.}}
\label{tab:app:char-families}
\resizebox{\linewidth}{!}{%
\begin{tabular}{llccc}
\toprule
Characteristic & Tercile
& \mchronos & \mtoto & \mtirex \\
\midrule
Seasonality
& Low
& \shortstack[l]{\texttt{diff} $-6.4$ $[-12.2,-0.9]$\\
  {\scriptsize BGAA $-6.9$}}
& \shortstack[l]{\texttt{winsor}$^\dagger$ $-3.2$ $[-13.7,+4.4]$\\
  {\scriptsize BGAA $-1.7$}}
& \shortstack[l]{\texttt{offset}$^\dagger$ $-0.4$ $[-1.7,+0.5]$\\
  {\scriptsize BGAA $-2.4$}} \\

& High
& \shortstack[l]{\texttt{decomp} $-4.3$ $[-7.6,-1.7]$\\
  {\scriptsize BGAA $-5.6$}}
& \shortstack[l]{\texttt{decomp} $-1.5$ $[-3.2,+0.1]$\\
  {\scriptsize BGAA $-1.2$}}
& \shortstack[l]{\texttt{multideseason} $-2.0$ $[-4.4,+0.6]$\\
  {\scriptsize BGAA $-3.4$}} \\
\addlinespace[3pt]

Trend strength
& Low
& \shortstack[l]{\texttt{decomp} $-2.6$ $[-5.4,-0.6]$\\
  {\scriptsize BGAA $-4.1$}}
& \shortstack[l]{\texttt{decomp} $-1.2$ $[-2.1,-0.2]$\\
  {\scriptsize BGAA $-0.9$}}
& \shortstack[l]{\texttt{multideseason} $-1.4$ $[-3.2,+0.4]$\\
  {\scriptsize BGAA $-2.0$}} \\

& High
& \shortstack[l]{\texttt{decomp} $-2.9$ $[-6.0,-0.5]$\\
  {\scriptsize BGAA $-8.1$}}
& \shortstack[l]{\texttt{winsor}$^\dagger$ $-2.5$ $[-11.0,+3.7]$\\
  {\scriptsize BGAA $-2.1$}}
& \shortstack[l]{\texttt{multideseason}$^\dagger$ $-0.9$ $[-3.7,+0.8]$\\
  {\scriptsize BGAA $-4.1$}} \\
\addlinespace[3pt]

Lag-1 autocorr.
& Low
& \shortstack[l]{\texttt{decomp} $-2.0$ $[-3.9,-0.3]$\\
  {\scriptsize BGAA $-2.0$}}
& \shortstack[l]{\texttt{winsor}$^\dagger$ $-1.8$ $[-5.5,+0.3]$\\
  {\scriptsize BGAA $-1.3$}}
& \shortstack[l]{\texttt{multideseason} $-0.7$ $[-1.8,+0.2]$\\
  {\scriptsize BGAA $-1.7$}} \\

& High
& \shortstack[l]{\texttt{diff} $-5.7$ $[-10.9,-0.8]$\\
  {\scriptsize BGAA $-7.7$}}
& \shortstack[l]{\texttt{decomp} $-1.2$ $[-3.2,+0.6]$\\
  {\scriptsize BGAA $-1.7$}}
& \shortstack[l]{\texttt{multideseason}$^\dagger$ $-2.6$ $[-6.3,+0.5]$\\
  {\scriptsize BGAA $-5.8$}} \\
\addlinespace[3pt]

Volatility
& Low
& \shortstack[l]{\texttt{diff} $-5.4$ $[-10.9,-0.3]$\\
  {\scriptsize BGAA $-7.6$}}
& \shortstack[l]{\texttt{winsor}$^\dagger$ $-1.9$ $[-11.9,+5.1]$\\
  {\scriptsize BGAA $-1.5$}}
& \shortstack[l]{\texttt{offset}$^\dagger$ $-0.3$ $[-1.5,+0.4]$\\
  {\scriptsize BGAA $-2.6$}} \\

& High
& \shortstack[l]{\texttt{decomp} $-3.0$ $[-5.1,-1.2]$\\
  {\scriptsize BGAA $-4.0$}}
& \shortstack[l]{\texttt{decomp} $-1.8$ $[-3.9,-0.1]$\\
  {\scriptsize BGAA $-2.7$}}
& \shortstack[l]{\texttt{winsor} $-0.4$ $[-1.0,-0.0]$\\
  {\scriptsize BGAA $-1.7$}} \\
\bottomrule
\end{tabular}%
}
\end{table}

\begin{table}[t]
\centering
\small
\setlength{\tabcolsep}{6pt}
\caption{\textbf{Historical verification versus statistical-feature
routing.}
MASE change versus the frozen base (\%; lower is better)
over 30 configurations (15 \gift, 15 \fev).
Routers are trained leave-one-configuration-out.
For each router, the second line gives its paired MASE
log-ratio gap to historical top-1 selection (95\%
configuration-bootstrap CI; positive favors the gate).
The best-of-12 row is selected post hoc and is not a
deployable model-selection procedure.
\bgaa\ with $k=3$ is shown as a reference with a larger
application budget.
\textbf{No tested router consistently outperforms
historical verification at matched application cost.}}
\label{tab:app:char-router}

\begin{tabular}{lccc}
\toprule
Selection method & \mchronos & \mtoto & \mtirex \\
\midrule
Historical top-1 + identity
& $\mathbf{-4.4}$
& $\mathbf{-1.7}$
& $\mathbf{-1.9}$ \\
\addlinespace[2pt]

Global view router
& \shortstack[c]{$-2.8$\\
  {\scriptsize $+1.7\ [0.3,3.3]$}}
& \shortstack[c]{$+2.0$\\
  {\scriptsize $+3.7\ [1.5,6.4]$}}
& \shortstack[c]{$+1.3$\\
  {\scriptsize $+3.2\ [1.4,5.5]$}} \\
\addlinespace[2pt]

Four-characteristic router
& \shortstack[c]{$-1.5$\\
  {\scriptsize $+3.0\ [0.0,7.8]$}}
& \shortstack[c]{$-0.7$\\
  {\scriptsize $+1.1\ [-0.7,2.7]$}}
& \shortstack[c]{$+2.8$\\
  {\scriptsize $+4.6\ [1.6,9.0]$}} \\
\addlinespace[2pt]

Best of 12 routers (post hoc)
& \shortstack[c]{$-4.3$\\
  {\scriptsize $+0.2\ [-1.1,1.2]$}}
& \shortstack[c]{$-1.6$\\
  {\scriptsize $+0.1\ [-2.6,2.1]$}}
& \shortstack[c]{$-0.7$\\
  {\scriptsize $+1.3\ [-0.8,3.7]$}} \\
\midrule

\bgaa\ ($k=3$; reference)
& $\mathbf{-6.1}$
& $\mathbf{-2.2}$
& $\mathbf{-3.2}$ \\
\bottomrule
\end{tabular}
\end{table}

\subsection{Prior geometry (\cref{sec:exp9})}
\label{sec:app:exp9}

\begin{figure}[h]
  \centering
  \includegraphics[width=\textwidth]{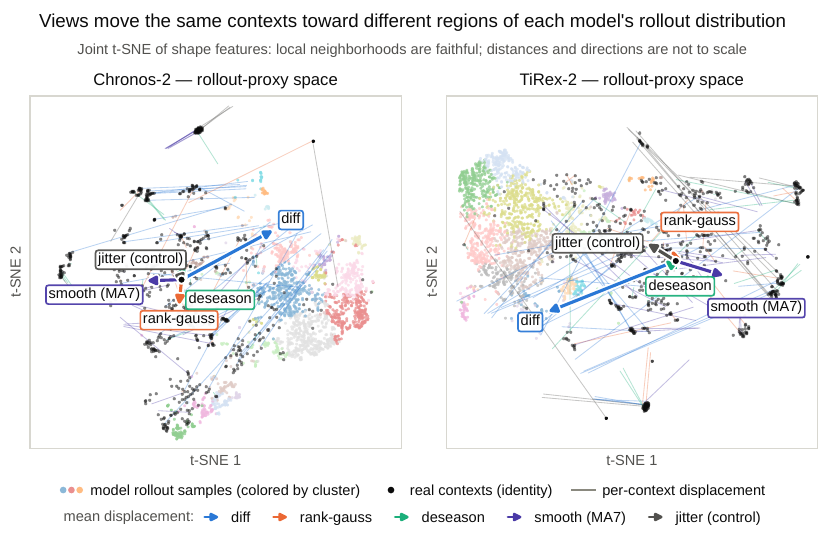}
  \caption{\textbf{Views move the same contexts toward different regions of each
model's rollout distribution.}
We use ancestral rollouts from each model's quantile head as a proxy for its
learned forecasting behavior and visualize the resulting $20$-dimensional shape-feature space
with joint t-SNE. Colored points are rollout samples, grouped by $k$-means
mode; black points are real \gift\ contexts; arrows show the mean displacement
induced by each view. We use the same contexts and views for both models, but
\texttt{diff} moves toward different rollout modes: for \mchronos, a
low-\texttt{iqr\_over\_std}, high-\texttt{step\_max} mode with mass $0.07$;
for \mtirex, a high-\texttt{kurtosis}, high-\texttt{skew} mode with mass
$0.02$. This provides the geometric counterpart to the fingerprint result.
t-SNE preserves local neighborhoods, but distances should not be interpreted
metrically.}
\label{fig:priorspace}
\end{figure}

We use ancestral rollouts as a proxy for each model's learned forecasting behavior, with
short-step samples drawn by inverse-CDF sampling from the model's own
$9$-quantile head. We study the resulting geometry by embedding both contexts and rollout samples in two spaces:
a $20$-dimensional shape-feature space and the model's own embedding space.
For each model, we standardize both using its rollout samples. We identify modes
by applying $k$-means to the rollout distribution and name each mode by its most
distinctive features. For context $i$ and view $g$, we define the displacement as
$d_i=z(g(x_i))-z(x_i)$, using $n=626$ \gift\ contexts.
\Cref{fig:priorspace} shows the joint t-SNE projection; we also compute PCA,
embedding-space, and three-dimensional projections.

\paragraph{Shape features.}
The $20$ shape features capture temporal dependence, trend, spectral structure,
distributional shape, and local variation (\cref{tab:shape-features}). We compute
them from the last $512$ points of each context or rollout after z-normalizing
the series, making the representation invariant to level and scale. For each
model, we standardize the features using the mean and standard deviation of its
own rollout samples. We then fit PCA and the $k$-means mode centroids in this
model-specific standardized space. Thus, the coordinates and modes in each
panel are defined relative to that model's own rollout distribution.

\begin{table}[h]
\centering
\small
\caption{\textbf{Shape features used for contexts and model rollouts.}
We compute all features on z-normalized series using at most the last $512$
observations. Bounded statistics are clipped and rescaled where needed so that
no single coordinate dominates the representation.}
\label{tab:shape-features}
\resizebox{\linewidth}{!}{
\begin{tabular}{p{2.8cm}p{5.1cm}p{6.0cm}}
\toprule
group & features & description \\
\midrule
Autocorrelation and memory
& \texttt{acf1}, \texttt{acf2}, \texttt{acf5}, \texttt{acf10},
  \texttt{acf\_first\_zero}, \texttt{pacf2}
& Autocorrelation at lags $1,2,5,10$; first non-positive ACF lag up to $99$
  (scaled by $100$); and lag-$2$ partial autocorrelation. \\

Trend and oscillation
& \texttt{trend\_r2}, \texttt{turning\_rate}, \texttt{crossing\_rate}
& Linear-trend $R^2$, frequency of sign changes in first differences, and
  frequency of crossings of the series mean. \\

Spectrum
& \texttt{spectral\_entropy}, \texttt{seasonal\_peak\_power},
  \texttt{seasonal\_peak\_freq}
& Normalized spectral entropy, power share of the strongest non-DC Fourier
  component, and its normalized frequency. The spectrum uses a Hann window. \\

Distribution shape
& \texttt{skew}, \texttt{kurtosis}, \texttt{iqr\_over\_std},
  \texttt{mad\_over\_std}, \texttt{tail\_frac\_2sig}
& Skewness, excess kurtosis, normalized IQR and MAD, and the fraction of
  observations with $|z|>2$. \\

Smoothness, jumps, and persistence
& \texttt{diff\_std\_ratio}, \texttt{step\_max}, \texttt{runlen\_mean}
& Relative first-difference variability, largest one-step change, and mean
  run length above or below the median. \\
\bottomrule
\end{tabular}}
\end{table}

\begin{table}[h]
\centering\small
\caption{\textbf{Directionality} in feature space: does a view induce a
consistent displacement across contexts? \emph{magnitude} is $\|\bar d\|$,
\emph{consistency} is $\mathrm{mean}_i \cos(d_i,\bar d)$ with a $95\%$
bootstrap CI. Read the two columns together; see the caveat below.}
\label{tab:app:directionality}
\resizebox{\linewidth}{!}{
\begin{tabular}{lrrrrrr}
\toprule
& \multicolumn{3}{c}{\mchronos} & \multicolumn{3}{c}{\mtirex} \\
\cmidrule(lr){2-4}\cmidrule(lr){5-7}
view & magnitude & consistency & $95\%$ CI & magnitude & consistency & $95\%$ CI \\
\midrule
\texttt{diff}          & $6.61$ & $0.773$ & $[0.761,0.785]$ & $13.29$ & $0.781$ & $[0.768,0.793]$ \\
\texttt{smooth\_ma7}   & $2.86$ & $0.768$ & $[0.753,0.781]$ & $7.27$ & $0.752$ & $[0.734,0.770]$ \\
\texttt{jitter\_0.002} & $1.21$ & $0.504$ & $[0.457,0.548]$ & $4.43$ & $0.456$ & $[0.413,0.499]$ \\
\texttt{sdiff}         & $2.19$ & $0.417$ & $[0.391,0.442]$ & $5.43$ & $0.490$ & $[0.463,0.517]$ \\
\texttt{rankgauss}     & $1.23$ & $0.341$ & $[0.310,0.371]$ & $2.22$ & $0.348$ & $[0.319,0.376]$ \\
\texttt{winsor\_0.05}  & $0.54$ & $0.337$ & $[0.312,0.362]$ & $0.93$ & $0.388$ & $[0.364,0.411]$ \\
\texttt{deseason}      & $1.07$ & $0.269$ & $[0.238,0.299]$ & $2.47$ & $0.293$ & $[0.256,0.326]$ \\
\texttt{lookback\_0.5} & $0.35$ & $0.067$ & $[0.050,0.084]$ & $0.73$ & $0.065$ & $[0.049,0.082]$ \\
\texttt{offset\_8}     & $0.04$ & $0.045$ & $[0.032,0.058]$ & $0.05$ & $0.040$ & $[0.029,0.052]$ \\
\texttt{seasontile}    & $0.07$ & $0.045$ & $[0.031,0.059]$ & $0.13$ & $0.048$ & $[0.035,0.063]$ \\
\bottomrule
\end{tabular}}
\end{table}

\textbf{Directionality.} \texttt{diff} and \texttt{smooth} move contexts in
similar directions across examples, with mean cosine $0.75$--$0.78$, compared
with $0.04$--$0.07$ for crops and offsets
(\cref{tab:app:directionality}).

\paragraph{Interpreting the directionality results.}
We treat these results as a mechanism diagnostic rather than direct evidence of
prior transport. The $\sigma=0.002$ jitter control still shows non-trivial
directionality ($\approx 0.50$), because even small perturbations systematically
change noise-sensitive features such as \texttt{diff\_std\_ratio} and
\texttt{turning\_rate}. We therefore also look at displacement magnitude and
mode alignment. Within the same model, \texttt{diff} moves contexts
$3.0$--$5.5\times$ farther than jitter, while jitter is among the smallest
non-crop movements in feature space and is the smallest displacement in the
model embedding space for three of the four models. We also test whether each
view moves toward a consistent rollout mode.

As we have clarified previously, these results should not be read as a literal application of \cref{prop:transport-main}. In particular, \texttt{rankgauss} is fitted to the
context and \texttt{diff} acts across time. Crops and offsets behave differently
as well: they induce only small shape-space
displacements (magnitude 0.04--0.05 for \texttt{offset\_8} and 0.35--0.73 for \texttt{lookback\_0.5}, versus 6.61--13.29 for \texttt{diff} (\cref{tab:app:directionality}). This is expected because they change the observation map rather than the shape
of the series, and this case is covered separately by \cref{rem:orbit}. Overall, we
use the geometry to test whether the views induce structured, model-specific
changes, not to establish causality.

\begin{table}[h]
\centering\small
\caption{\textbf{Non-redundancy of view displacements.}
We compute the cosine similarity between the mean displacement directions
induced by different views. Values near $0$ indicate that two views move
contexts in different directions, while values near $\pm1$ indicate similar or
opposing directions. For each model and representation space, we report the mean and maximum absolute
off-diagonal cosine. For \mchronos\ in feature space, the maximum is the
\texttt{diff}--\texttt{sdiff} pair (0.81).}
\label{tab:app:viewview}
\begin{tabular}{lrrrr}
\toprule
& \multicolumn{2}{c}{feature space} & \multicolumn{2}{c}{embedding space} \\
\cmidrule(lr){2-3}\cmidrule(lr){4-5}
model & mean $|$off-diag$|$ & max & mean $|$off-diag$|$ & max \\
\midrule
\mtirex   & $0.42$ & $0.90$ & $0.54$ & $0.85$ \\
\mchronos & $0.36$ & $0.81$ & $0.55$ & $0.97$ \\
\mrone    & $0.48$ & $0.96$ & $0.44$ & $0.97$ \\
\mtoto    & $0.40$ & $0.84$ & $0.33$ & $1.00$ \\
\bottomrule
\end{tabular}
\end{table}

\textbf{Non-redundancy.} Displacement directions from different views have
mean absolute cosine $0.36$--$0.48$ in feature space, indicating that the
views make distinct changes (\cref{tab:app:viewview}).

The extreme pairs are also consistent with the intended semantics of the views.
For \mchronos, the two differencing families are strongly aligned,
$\cos(\texttt{diff},\texttt{sdiff})=+0.81$, while the intentionally opposing
\texttt{deseason} and \texttt{smooth} views have cosine $-0.62$. In contrast,
\texttt{jitter} and \texttt{lookback}, which do not share a common mechanism,
have only weak alignment, with cosine $+0.12$.

\begin{table}[h]
\centering\small
\caption{\textbf{Destination of the \texttt{diff} displacement.}
For each model, we report the rollout mode most aligned with the mean
\texttt{diff} displacement, together with the cosine between that displacement
and the direction from the real-context centroid to the mode centroid. The
alignment is strong for all four models, but the target mode differs across
models, as also shown in \cref{fig:priorspace}. Mode names are based on the
cluster's most distinctive features, and ``mass'' denotes the fraction of
rollout samples assigned to that mode.}
\label{tab:app:viewmode}
\begin{tabular}{llrr}
\toprule
model & top mode for \texttt{diff} & cosine & mass \\
\midrule
\mtirex   & high \texttt{kurtosis} / high \texttt{skew}                      & $+0.97$ & $0.02$ \\
\mchronos & low \texttt{iqr\_over\_std} / high \texttt{step\_max}            & $+0.97$ & $0.07$ \\
\mrone    & high \texttt{tail\_frac\_2sig} / high \texttt{seasonal\_peak\_freq} & $+0.96$ & $0.05$ \\
\mtoto    & low \texttt{iqr\_over\_std} / low \texttt{mad\_over\_std}        & $+0.92$ & $0.08$ \\
\bottomrule
\end{tabular}
\end{table}

\begin{table}[h!]
\centering\small
\caption{\textbf{Forecast signatures by view family} for \mchronos\ and
\mtirex\ ($15$ configs each). \emph{width} is the log-ratio of predictive
interval width relative to the base, \emph{sign} is the fraction of
configs with the reported sign, \emph{shift} is the scale-normalized
median shift, and \emph{cov} is the change in coverage. Families that materially
change the forecast distribution do so consistently, while the remaining
families have much smaller effects.}
\label{tab:app:signatures}
\begin{tabular}{lrrrrrrrr}
\toprule
& \multicolumn{4}{c}{\mchronos} & \multicolumn{4}{c}{\mtirex} \\
\cmidrule(lr){2-5}\cmidrule(lr){6-9}
family & width & sign & shift & cov & width & sign & shift & cov \\
\midrule
\texttt{diff}       & $+2.249$ & $0.93$ & $0.555$ & $+0.185$ & $+2.103$ & $1.00$ & $0.516$ & $+0.144$ \\
\texttt{sdiff}      & $+0.398$ & $0.87$ & $0.578$ & $+0.034$ & $+0.375$ & $0.93$ & $0.507$ & $+0.015$ \\
\texttt{recursive}  & $-0.461$ & $0.93$ & $0.068$ & $-0.178$ & $-0.548$ & $1.00$ & $0.050$ & $-0.202$ \\
\texttt{fft}        & $-1.328$ & $0.93$ & $0.868$ & $-0.676$ & $-1.541$ & $0.93$ & $0.753$ & $-0.714$ \\
\texttt{rankgauss}  & $+0.004$ & $0.60$ & $0.144$ & $-0.029$ & $-0.002$ & $0.53$ & $0.125$ & $-0.008$ \\
\texttt{jitter}     & $0.000$  & $0.80$ & $0.004$ & $+0.001$ & $0.000$  & $0.80$ & $0.004$ & $0.000$  \\
\texttt{lookback}   & $0.000$  & ---    & $0.036$ & $+0.003$ & $-0.002$ & $0.60$ & $0.045$ & $0.000$  \\
\texttt{offset}     & $0.000$  & ---    & $0.002$ & $0.000$  & $0.000$  & ---    & $0.000$ & $0.000$  \\
\bottomrule
\end{tabular}
\end{table}

\begin{figure}[t]
  \centering

  \begin{subfigure}[t]{0.75\textwidth}
    \centering
    \includegraphics[width=\textwidth]{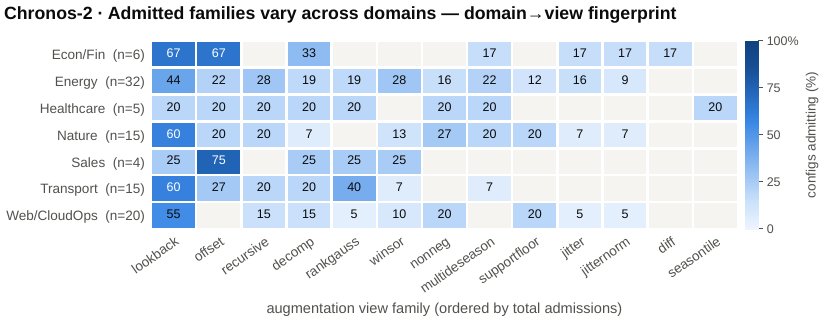}
    \caption{domain $\to$ admitted family}
    \label{fig:fingerprint}
  \end{subfigure}

  \vspace{0.5em}

  \begin{subfigure}[t]{0.75\textwidth}
    \centering
    \includegraphics[width=\textwidth]{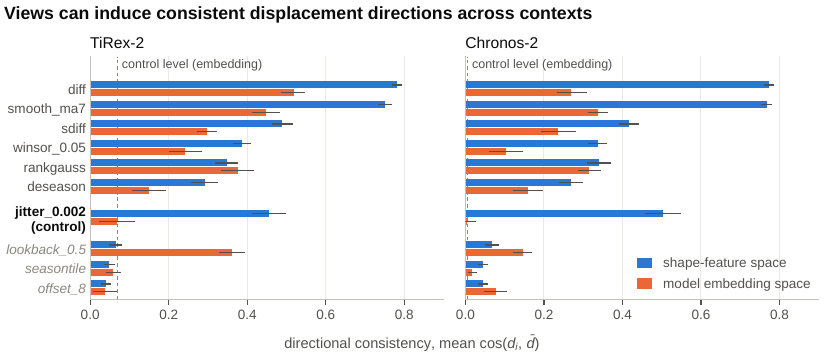}
    \caption{views induce consistent displacements}
    \label{fig:directionality}
  \end{subfigure}

  \vspace{0.5em}

  \begin{subfigure}[t]{0.75\textwidth}
    \centering
    \includegraphics[width=\textwidth]{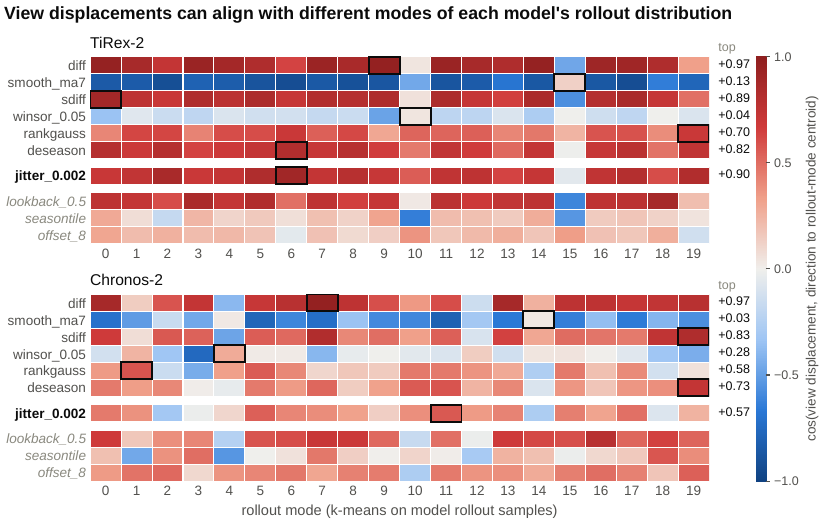}
    \caption{view $\to$ rollout mode, per model}
    \label{fig:viewmode}
  \end{subfigure}

  \caption{\textbf{Summary of the mechanism analyses in \cref{sec:mechanism}.}
\textbf{(a)} Admitted families vary across domains.
\textbf{(b)} Individual views induce consistent displacement directions.
\textbf{(c)} These displacements can align with different modes of each model's
rollout distribution. \Cref{fig:priorspace} visualizes the same effect in a
rollout-based proxy for each model's learned forecasting behavior.}
\end{figure}

\paragraph{Mode movement and poorly aligned views.}
We also measure how strongly each view changes mode assignments, independent of
its displacement direction. For each view, we compute the Jensen--Shannon
divergence between its distribution over rollout modes and the corresponding
identity distribution. On \mchronos, with jitter at $0.019$, we obtain
\texttt{diff} $0.269$, \texttt{sdiff} $0.106$, \texttt{rankgauss} $0.054$,
\texttt{smooth\_ma7} $0.050$, \texttt{deseason} $0.042$, \texttt{winsor}
$0.022$, \texttt{lookback} $0.019$, and $0.001$ for both \texttt{offset} and
\texttt{seasontile}. \mtirex\ shows the same ordering at the top:
\texttt{diff} $0.402$, \texttt{sdiff} $0.157$, and \texttt{smooth\_ma7}
$0.144$, with crop-based views again near the bottom. \texttt{smooth\_ma7} is a useful counterexample. It moves contexts by an
amount comparable to several useful views, but \cref{fig:viewmode} shows little alignment with any rollout mode. Its largest
mode cosine is only $+0.03$ over $20$ modes for \mchronos\ and $+0.13$ for
\mtirex, with the remaining modes negatively aligned. Thus, large movement
alone is not sufficient: \texttt{smooth\_ma7} changes the representation
substantially without moving toward the rollout modes targeted by the useful
views.

\subsection{Forecast signatures (\cref{sec:exp10})}
\label{sec:app:exp10}

We characterize the forecast-level effect of each view family using three
statistics: the log-ratio of predictive interval width, the median shift
normalized by the label scale of each window, and the change in coverage. We
compute these quantities over the test windows where the view is applied, first
per family and config, and then aggregate using the median across windows
and configs.

\texttt{fft} consistently narrows predictive intervals by $-1.33$ to $-1.54$,
while \texttt{lookback}, \texttt{offset}, and \texttt{jitter} have little
effect on interval width (\cref{tab:app:signatures}). Part of
\texttt{diff}'s interval widening comes from the cumulative-sum pull-back,
which sums marginal quantiles across the horizon (\cref{rem:scope}). Thus, the
signature reflects the full view, not just the input transform.

We summarize the signatures in two ways. First, families are clearly distinct
within each model. The between/within-family variance ratios for
(width, shift, coverage) are $[32.0,10.4,71.6]$ for \mchronos,
$[25.9,9.6,76.0]$ for \mtirex, $[34.2,10.0,56.7]$ for \mrone, and
$[35.5,9.9,86.9]$ for \mtoto. Thus, across all four models and all three
statistics, between-family variation is roughly $10$--$87\times$ larger than
within-family variation.

Second, the non-degenerate signatures are stable across models. Comparing
\mchronos\ and \mtirex, the per-family signature cosines range from $+0.95$ to
$+1.00$ for the ten families with nonzero signatures
(\texttt{diff} $+1.00$, \texttt{sdiff} $+1.00$, \texttt{recursive} $+1.00$,
\texttt{fft} $+0.99$, \texttt{jitternorm} $+0.97$). The remaining four
families: \texttt{nonneg}, \texttt{offset}, \texttt{seasontile}, and
\texttt{supportfloor}, have a zero median signature on \mtirex. They leave most windows unchanged, although they alter enough windows to be admitted occasionally
(\cref{tab:app:fingerprint}). Thus, their cosines are undefined rather than small. Since this cosine is computed from
only three statistics, we use it as a consistency check rather than as strong
evidence on its own. We also report the fraction of dropped windows due to
non-finite fp16 values or zero label scale. The worst case is $0.47$ for
\texttt{diff} on \mchronos; the median is $0.0$, and $9/15$ configs
drop no windows.

\clearpage